\documentclass[sn-mathphys-ay]{sn-jnl}% 

\def\tsc#1{\csdef{#1}{\textsc{\lowercase{#1}}\xspace}}
\tsc{WGM}
\tsc{QE}
\usepackage{import} 
\usepackage{siunitx} 
\DeclareSIUnit{\nothing}{\relax}  

\usepackage{xcolor} % MV: to use textcolor commmand
\usepackage{acro}[=v3] %MV: for list of abbreviations

\DeclareAcronym{GT}{
    short = GT,
    long  = ground truth,
    }

\DeclareAcronym{AEC}{
    short = AEC,
    long  = {Architecture, Engineering, and Construction},
    }
    
\DeclareAcronym{KNN}{
    short = KNN,
    long  = K-nearest neighbors,
    }
    
\DeclareAcronym{API}{
    short = API,
    long  = Application Programming Interface,
}

\DeclareAcronym{RTK}{
    short = RTK,
    long  = real-time kinematic,
    }
\DeclareAcronym{iSAM}{
    short = iSAM,
    long  = Incremental Smoothing and Mapping,
    }

\DeclareAcronym{MDC}{
    short = MDC,
    long  = motion distortion correction,
    }

\DeclareAcronym{OBJ}{
    short = OBJ,
    long  = Wavefront .obj file,
    }
    
\DeclareAcronym{STL}{
    short = STL,
    long  = stereolithography,
    }
    
\DeclareAcronym{PNG}{
    short = PNG,
    long  = Portable Network Graphics,
    }
    
\DeclareAcronym{COLLADA}{
    short = COLLADA,
    long  = Collaborative Design Activity,
    }

\DeclareAcronym{DAE}{
    short = DAE,
    long  = Digital Asset Exchange,
    }

\DeclareAcronym{YAML}{
    short = YAML,
    long  = YAML Ain't Markup Language,
    }
    
\DeclareAcronym{PGM}{
    short = PGM,
    long  = Portable Gray Map,
    }
    
\DeclareAcronym{CLI}{
    short = CLI,
    long  = command-line interface,
    }
    
\DeclareAcronym{GeoLab}{
    short = GeoLab,
    long  = Geodätisches Labor,
    }
    
\DeclareAcronym{2D}{
	short   = 2D,
	long    = Zweidimensional,
	}
	
\DeclareAcronym{3D}{
	short   = 3D,
	long    = Dreidimensional,
	}

\DeclareAcronym{BIM}{
	short   = BIM,
	long    = Building Information Modeling,
	}
	
\DeclareAcronym{BIM model}{
	short   = BIM model,
	long    = building information model,
	}
	
\DeclareAcronym{URDF}{
	short   = URDF,
	long    = Universal Robot Description Format,
	}
	
\DeclareAcronym{SOTA}{
	short   = SOTA,
	long    = State-of-the-art,
	}	

\DeclareAcronym{SDF}{
	short   = SDF,
	long    = Simulation Definition Format,
	}
	
\DeclareAcronym{SVG}{
	short   = SVG,
	long    = Scalable Vector Graphics,
	}
	
\DeclareAcronym{BIRS}{
	short   = BIRS,
	long    = Building Information Robotic System,
	}

\DeclareAcronym{Scan-vs-BIM}{
	short   = Scan-vs-BIM,
	long    = comparison between a point cloud and a \ac{BIM} model,
	}

\DeclareAcronym{GIS}{
	short   = GIS,
	long    = Geographic Information System,
	}

\DeclareAcronym{BGU}{
	short   = BGU,
	long    = \DAfaculty{},
	}
    
\DeclareAcronym{EMM}{
	short   = EMM,
	long    = Environment Mapping Module,
	}
	
\DeclareAcronym{SLAM}{
	short   = SLAM,
	long    = Simultaneous Localization and Mapping,
	}
	
\DeclareAcronym{V-SLAM}{
	short   = V-SLAM,
	long    = Visual SLAM,
	}
	
\DeclareAcronym{UGV}{
	short   = UGV,
	long    = Unmanned Ground Vehicle ,
	}	
	
\DeclareAcronym{UAV}{
	short   = UAV,
	long    = Unmanned Aerial Vehicle,
	}	
		
 \DeclareAcronym{UV}{
	short   = UV,
	long    = Unmanned Vehicle,
	}	
	
\DeclareAcronym{UVs}{
	short   = UVs,
	long    = Unmanned Vehicles,
	}
	
\DeclareAcronym{LiDAR}{
	short   = LiDAR,
	long    = Light Detection and Ranging,
	}

\DeclareAcronym{SBAS}{
	short   = SBAS,
	long    = Satellite Based Augmentation Systems,
	}	
	
\DeclareAcronym{IMU}{
	short   = IMU,
	long    = Inertial Measurement Units,
	}

\DeclareAcronym{GNSS}{
	short   = GNSS,
	long    = global navigation satellite system,
	}		

\DeclareAcronym{GPS}{
	short   = GPS,
	long    = Global Positioning System,
	}	
	
\DeclareAcronym{D-GPS}{
	short   = D-GPS,
	long    = Differential \ac{GPS},
	}	
	
\DeclareAcronym{MAP}{
	short   = MAP,
	long    = Maximum A Posteriori,
	}		
	
\DeclareAcronym{EKF}{
	short   = EKF,
	long    = Extended Kalman Filter,
	}	
\DeclareAcronym{UKF}{
	short   = UKF,
	long    = Unscented Kalman Filter,
	}

\DeclareAcronym{BA}{
	short   = BA,
	long    = Bundle-Adjustment,
	}	
		
\DeclareAcronym{DNN}{
	short   = DNN,
	long    = Deep Neural Network,
	}	
	
\DeclareAcronym{GNN}{
	short   = GNN,
	long    = Graph neural networks,
	}	
 
\DeclareAcronym{DL}{
	short   = DL,
	long    = Deep Learning,
	}	
	
\DeclareAcronym{UWB}{
	short   = UWB,
	long    = Ultra Wide Band,
	}	
	
\DeclareAcronym{TLS}{
	short   = TLS,
	long    = Terrestrial Laser Scanner,
	}
	
\DeclareAcronym{MLS}{
	short   = MLS,
	long    = Mobile Laser Scanner,
	}

\DeclareAcronym{SAR}{
	short   = SAR,
	long    = Search and Rescue,
	}	
	
\DeclareAcronym{SfM}{
	short   = SfM,
	long    = Structure from Motion,
	}

\DeclareAcronym{MVS}{
	short   = MVS,
	long    = Multi-View Stereo,
	}

\DeclareAcronym{KPIs}{
	short   = KPIs,
	long    = Key Performance Indicators,
	}

\DeclareAcronym{MVE}{
	short   = MVE,
	long    = Multiview Environment,
	}	
	
\DeclareAcronym{RTPS}{
	short   = RTPS,
	long    = Real Time Positioning System,
	}
\DeclareAcronym{NIR}{
	short   = NIR,
	long    = Near-infrared,
	}

 \DeclareAcronym{RGB}{
	short   = RGB,
	long    = {red, green, and blue},
	}

\DeclareAcronym{P2P}{
	short   = P2P,
	long    = Point-to-Point,
	}
	
\DeclareAcronym{ICP}{
	short   = ICP,
	long    = Iterative Closest Point,
	}

\DeclareAcronym{GICP}{
	short   = GICP,
	long    = Generalized \ac{ICP},
	}	
	
\DeclareAcronym{AMCL}{
	short   = AMCL,
	long    = Adaptive Monte Carlo Localization,
	}	
	
\DeclareAcronym{GMCL}{
	short   = GMCL,
	long    = General Monte Carlo Localization,
	}	
 
\DeclareAcronym{LIO}{
	short   = LIO,
	long    = LiDAR Inertial Odometry,
	}	
\DeclareAcronym{DLIO}{
	short   = DLIO,
	long    = Direct LiDAR Inertial Odometry,
	}	
 
\DeclareAcronym{SER}{
	short   = SER,
	long    = Similar Energy Region,
	}	
		
\DeclareAcronym{PF}{
	short   = PF,
	long    = Particle Filter,
	}	
	
\DeclareAcronym{RANSAC}{
	short   = RANSAC,
	long    = Random Sample Consensus,
	}	
	
\DeclareAcronym{ROS}{
	short   = ROS,
	long    = Robot Operating System,
	}	

\DeclareAcronym{DoF}{
	short   = DoF,
	long    = Degrees of Freedom,
	}
	
\DeclareAcronym{MAV}{
	short   = MAV,
	long    = Micro Aerial Vehicle,
	}	
	
\DeclareAcronym{VP}{
	short   = VP,
	long    = Vanishing Points,
	}	
	
\DeclareAcronym{VL}{
	short   = VL,
	long    = Vanishing Lines,
	}	
	
\DeclareAcronym{VR}{
	short   = VR,
	long    = Virtual Reality,
	}	
	
\DeclareAcronym{AR}{
	short   = AR,
	long    = Augmented Reality,
	}	
	
\DeclareAcronym{MR}{
	short   = MR,
	long    = Mixed Reality,
	}	
	
\DeclareAcronym{LoD}{
	short   = LoD,
	long    = Level of Detail,
	}	
 
\DeclareAcronym{FoV}{
	short   = FoV,
	long    = Field of View,
	}	
 
\DeclareAcronym{IFC}{
	short   = IFC,
	long    = Industry Foundation Classes,
	}	
	
\DeclareAcronym{CPS}{
	short   = CPS,
	long    = Cyber-Physical Systems,
	}		
	
\DeclareAcronym{LOAM}{
	short   = LOAM,
	long    = LiDAR Odometry and Mapping,
	}	
	
\DeclareAcronym{A-LOAM}{
  short = A-LOAM,
  long = Advanced implementation of LOAM
  }
  
\DeclareAcronym{F-LOAM}{
  short = F-LOAM,
  long = Fast LiDAR Odometry And Mapping
  }

\DeclareAcronym{IFR}{
  short = IFR,
  long = International Federation of Robotics
  }

\DeclareAcronym{RGB-D}{
  short = RGB-D,
  long = Red-Green-Blue-Depth
  }
   
\DeclareAcronym{OGM}{
  short = OGM,
  long = Occupancy Grid Map
  }
  
\DeclareAcronym{KLD}{
  short = KLD,
  long = Kullback-Leibler distance
  }
    
 \DeclareAcronym{MCL}{
  short = MCL,
  long = Monte Carlo Localization
  }
    
 \DeclareAcronym{GBL}{
  short = GBL,
  long = Graph-based Localization
  }     
  
\DeclareAcronym{RViz}{
  short = RViz,
  long = ROS visualization
  }

\DeclareAcronym{APE}{
  short = APE,
  long = Absolute Pose Error
  }  

\DeclareAcronym{RE}{
  short = RE,
  long = Rotational Error 
  } 
  
  \DeclareAcronym{RMSE}{
  short = RMSE,
  long =  Root Mean Square Error 
  } 
  
  \DeclareAcronym{PGBM}{
  short = PGBM,
  long =  Pose Graph-based Maps 
  } 

 \DeclareAcronym{CAD}{
  short = CAD,
  long =  Computer-aided Design
  } 

 \DeclareAcronym{Ogm2Pgbm}{
  short = Ogm2Pgbm,
  long =  \ac{OGM} to Pose Graph-based map
  } 

 \DeclareAcronym{Scan-Map deviations}{
  short = Scan-Map deviations,
  long =  discrepancies between the reference map and the current state of the real-world
  } 
  
\DeclareAcronym{SCD}{
  short = SCD,
  long =  Scan Context Descriptor
  } 
  
\DeclareAcronym{SC}{
  short = SC,
  long =  Scan Context
}

\DeclareAcronym{ISC}{
  short = ISC,
  long =  Indoor Scan Context
} 

\DeclareAcronym{ISCD}{
  short = ISCD,
  long =  Indoor Scan Context Descriptor
  } 
  
\DeclareAcronym{PD}{
  short = PD,
  long =  Positive Differences
}

\DeclareAcronym{ND}{
  short = ND,
  long =  Negative Differences
  } 

\DeclareAcronym{VG}{
  short = VG,
  long =  Voxel Grid
  } 
  
\DeclareAcronym{SD}{
  short = SD,
  long =  Session Data
  } 
  
\DeclareAcronym{MSS}{
  short = MSS,
  long =  Multi-Session SLAM
  } 
  
\DeclareAcronym{PP}{
  short = PP,
  long =  Path Planner
  } 
  
\DeclareAcronym{BlenSor}{
  short = BlenSor,
  long =  Blender Sensor Simulation Toolbox 
  } 

\DeclareAcronym{TSP}{
  short = TSP,
  long =  Travelling Salesperson Problem  
  } 
  
\DeclareAcronym{DT}{
  short = DT,
  long =  Digital Twin  
  }

\usepackage{amssymb}
\usepackage{amsthm}
\usepackage{amsmath}
 
\usepackage{lineno}

\usepackage{multirow}	
\usepackage{booktabs}  

\usepackage{amsmath}
\usepackage{geometry}
\usepackage{array}
\usepackage{longtable}
\usepackage{fix-cm}

\usepackage{graphicx}
\usepackage{subfig}
\usepackage{xurl}

\usepackage{titlesec}
\usepackage{hyperref}

\titleclass{\subsubsubsection}{straight}[\subsection]
\newcounter{subsubsubsection}[subsubsection]
\renewcommand\thesubsubsubsection{\thesubsubsection.\arabic{subsubsubsection}}
\titleformat{\subsubsubsection}
  {\normalfont\normalsize\bfseries}{\thesubsubsubsection}{1em}{}
\titlespacing*{\subsubsubsection}
{0pt}{3.25ex plus 1ex minus .2ex}{1.5ex plus .2ex}

\titleclass{\subsubsubsubsection}{straight}[\subsubsubsection]
\newcounter{subsubsubsubsection}[subsubsubsection]
\renewcommand\thesubsubsubsubsection{\thesubsubsubsection.\arabic{subsubsubsubsection}}
\titleformat{\subsubsubsubsection}
  {\normalfont\normalsize\bfseries}{\thesubsubsubsubsection}{1em}{}
\titlespacing*{\subsubsubsubsection}
{0pt}{3.25ex plus 1ex minus .2ex}{1.5ex plus .2ex}

\makeatletter
\renewcommand\paragraph{\@startsection{paragraph}{5}{\z@}%
  {3.25ex \@plus1ex \@minus.2ex}%
  {-1em}%
  {\normalfont\normalsize\bfseries}}
\renewcommand\subparagraph{\@startsection{subparagraph}{6}{\parindent}%
  {3.25ex \@plus1ex \@minus .2ex}%
  {-1em}%
  {\normalfont\normalsize\bfseries}}
\def\toclevel@subsubsubsection{4}
\def\toclevel@subsubsubsubsection{5}
\def\toclevel@paragraph{6}
\def\toclevel@subparagraph{7}
\def\l@subsubsubsection{\@dottedtocline{4}{7em}{4em}}
\def\l@subsubsubsubsection{\@dottedtocline{5}{10em}{5em}}
\def\l@paragraph{\@dottedtocline{6}{14em}{6em}}
\def\l@subparagraph{\@dottedtocline{7}{17em}{7em}}
\makeatother

\usepackage{algorithm}%
\usepackage{algorithmicx}%

\usepackage{orcidlink}
\usepackage{graphicx}

\begin{document} %--------------------------------------------------

\title[SLAM2REF: Advancing Long-Term Mapping with 3D LiDAR and Reference Map Integration for Precise 6-DoF Trajectory Estimation and Map Extension]{SLAM2REF: Advancing Long-Term Mapping with 3D LiDAR and Reference Map Integration for Precise 6-DoF Trajectory Estimation and Map Extension}

% Authors
% \author*[1]{\fnm{blinded.} \sur{-}}\email{-} 
% \orcid{0000-0002-0606-4035}

\author*[1]{\fnm{Miguel A.} \sur{Vega-Torres}}\email{miguel.vega@tum.de} 
% \orcid{0000-0002-0606-4035}

\author[1]{\fnm{Alexander} \sur{Braun}} %\email{alex.braun@tum.de}
% \orcid{0000-0003-1513-5111}

\author[1]{\fnm{Andr\'{e}} \sur{Borrmann}}
%\email{andre.borrmann@tum.de}
% \orcid{0000-0003-2088-7254}

\affil*[1]{\orgdiv{Chair of Computational Modeling and Simulation}, \orgname{Technical University of Munich}, \orgaddress{\street{Arcisstraße 21}, \city{Munich}, \postcode{80333}, \state{Bavaria}, \country{Germany}}}

%% Abstract

\abstract{This paper presents a pioneering solution to the task of integrating mobile 3D LiDAR and inertial measurement unit (IMU) data with existing building information models or point clouds, which is crucial for achieving precise long-term localization and mapping in indoor, GPS-denied environments.
Our proposed framework, SLAM2REF, introduces a novel approach for automatic alignment and map extension utilizing reference 3D maps. 
The methodology is supported by a sophisticated multi-session anchoring technique, which integrates novel descriptors and registration methodologies.
Real-world experiments reveal the framework's remarkable robustness and accuracy, surpassing current state-of-the-art methods. 
Our open-source framework's significance lies in its contribution to resilient map data management, enhancing processes across diverse sectors such as construction site monitoring, emergency response, disaster management, and others, where fast-updated digital 3D maps contribute to better decision-making and productivity.  
Moreover, it offers advancements in localization and mapping research. Link to the repository: \url{https://github.com/MigVega/SLAM2REF}, Data: \url{https://doi.org/10.14459/2024mp1743877} . 
}

% Keywords

\maketitle 

% Main text %---------------------------------------------------------

% \linenumbers  % Remove for2columns

% 1: Introduction
\section{Introduction}
\label{sec:Introduction}

%   (1/5) Background or Context: 
%   (2/5) Significance or Importance (Motivation):
%   (3/5) Research Question or Problem Statement:
%   (4/5) Scope (assumptions) and Limitations: 
%   (5/5) Overview of the Paper:

%    (1/5) Background or Context: 
% why SLAM alone is not the solution:
Nowadays, mobile mapping systems incorporated into mobile robots or handheld devices equipped with sensors and applying state-of-the-art \ac{SLAM} algorithms allow the quick creation of updated 3D maps.
However, these maps are in their local coordinates systems and, therefore, separated from any prior information. 
Additionally, they might contain potential drift issues, rendering them unsuitable for comparative analysis or change detection.

Several real-world applications require the capacity to align, compare, and manage 3D data received at various intervals that may be separated by lengthy intervals of time. 
This process is referred to as long-term map management.

%     (2/5) Significance or Importance: Why is this problem significant:

Long-term map management is crucial since the real world constantly evolves and changes. 
This applies to humans who want to utilize the map to comprehend the current situation and its evolution and to autonomous robots for effective and fast navigation.

%%%% 2.1) About digital twin.

Moreover, achieving accurate alignment and effective management of extensive datasets represent significant challenges in enabling the creation of \acl{DT}s (DTs) for cities and buildings \citep{bormann2024,Mylonas:2021:digitalTwin}.  
As explained by \citep{Botin-Sanabria:2022:digitalTwin}, in complex implementations, automatic alignment of 3D data becomes imperative to achieve \ac{DT}s with maturity levels of 3 or higher. 
Such levels necessitate the augmentation of models with a continuous flow of real-world information.

%%%% 2.2) Mobile Mapping (NavVis) ~ State of Practice!

An automatic map alignment and change detection pipeline also contribute to the seamless integration of mapping devices into existing workflows in the industry. 
A recent survey revealed that the compatibility of mapping devices with existing tools is, after the budget, the second most crucial barrier surrounding the usage of mobile mapping devices \citep{NavVis2022}. 

%%%% 2.3) Concrete use cases (end users)
For example, an up-to-date 3D digital map can help construction site managers promptly distinguish as-planned and as-built differences, thus reducing the probability of long-schedule delays and high-cost overruns. 
Similarly, an updated 3D representation of the site can also help first responders during an emergency to improve situational awareness and enable decision-making to save lives effectively and safely \citep{alliez2020real, Guo2021MappingAndLocalization}. 

Furthermore, if a robot can align the measurements of an onboard sensor with a reference map (i.e., the robot can localize itself within the map), the semantically enriched \ac{BIM model} or the reference map can serve as a valuable source of information for various autonomous robotic activities. 
These activities include but are not limited to path planning \citep{Dugstad:2022:SAS-Indoor-PP}, object inspection \citep{Kim.2022}, and maintenance and repair operations \citep{Kim.2021}.

\ac{GPS} can be a viable option for outdoor localization and rough alignment. However, for indoor environments, GPS is often impractical because it requires a direct line of sight to at least four satellites—three to determine the 3D position and one for time correction. To address this, various Indoor Positioning System (IPS) alternatives use radio signals, such as Wi-Fi or Bluetooth, as well as AprilTags \citep{LOPEZDETERUEL2017_wifi, KAYHANI2022_AprilTags, Kayhani2023_AprilTags, Koide2022_tags}. The downside of these systems is that they require additional strategically placed sensors or landmarks, which can increase the cost and effort of implementing such a positioning system. 
Nevertheless, although not always accessible, 3D prior maps of buildings are increasingly becoming standard in modern construction. These maps, often in the form of \ac{BIM model}s or point clouds, document the state of the building during and after construction or in the design phases.

%%%% 2.4) About GT poses for SLAM research

Besides being useful for autonomous robotic tasks, aligning sensor data with an accurate reference map allows the retrieval of the sensor's precise 6 \ac{DoF} \ac{GT} poses in the entire trajectory.

These \ac{GT} poses serve multiple functions. 
They enable precise identification of the capture locations of point clouds and images necessary for generating an accurate, updated 3D map.
Additionally, they facilitate the assessment of the efficacy of SLAM, odometry, and localization algorithms. This capability is particularly crucial for advancing research and development in this field.

Historically, obtaining \ac{GT} poses has necessitated costly equipment like \ac{RTK}-corrected \ac{GNSS} for outdoor environments or laser trackers and motion capture systems for indoor settings \citep{liu2021simultaneous}. 
However, the expensive costs associated with these methods pose a substantial barrier for individual researchers. 
Additionally, acquiring dense GT poses for extended trajectories, especially in indoor scenarios, has been found to be challenging \citep{hiltiChallenge2022}.
 
Recent studies, such as by the ConSLAM \citep{trzeciak2023conslam,trzeciak2023conslamExtension} and Newer College \citep{Ramezani_2020, zhang2022multicamera} datasets, have leveraged \ac{TLS} point clouds—providing millimeter-precise 3D scans of the environment—to be used as reference \ac{GT} map and overcome these limitations. 
Through semi-automatic techniques, researchers have effectively aligned mobile \ac{LiDAR} measurements with TLS point clouds.
This advancement represents a significant step forward in \ac{SLAM} research towards automatic, accurate \ac{GT} pose acquisition methods suitable for both large indoor and outdoor scenarios.

%%------------------------------------------
%     (3/5) Research Question or Problem Statement/ Research Gap 

%%      3.1) A brief description of the problem that this research handles:
To enable long-term map management and the automatic retrieval of precise 6-\ac{DoF} poses for mobile \ac{LiDAR}-based localization and mapping research, this study proposes \textbf{SLAM2REF}, an open-source\footnote{ Link to the repository: \url{https://github.com/MigVega/SLAM2REF}} framework that uses a \ac{BIM model} or a pre-existing point cloud as a reference map to allow an automatic alignment and correction of a map created with \ac{SLAM} or odometry systems.

Herein, we adopt the term \textit{reference map} to encompass a spectrum of environmental representations, such as designated \ac{BIM model}s, point clouds, or meshes.

As will be discussed in Section \ref{chap:related_work}, several researchers have investigated using a reference map for robot localization \citep{vega:2022:2DLidarLocalization}.
However, only a few aim to create an accurate, updated 3D map that is aligned and corrected with the information in the reference map. 

Furthermore, most research methods demand a reasonably good estimate of the robot's initial position, which must also be within the reference map. 
In addition, nearly no approach takes \ac{Scan-Map deviations} into account. 

%     (4/5) Scope (assumptions) and Limitations: 
\ac{Scan-Map deviations} can be classified into three categories: Firstly, the deviations coming from the presence of clutter or furniture absent in the reference map; Secondly, deviations due to the presence of dynamic (i.e., moving) elements in the environment while scanning; and thirdly, the presence of alterations on the permanent elements of the building, such as walls and columns.
In this research, we focus on addressing the first two categories of deviations.
Nonetheless, minor discrepancies in some permanent elements of the environment, such as holes or slight shifts in single columns or walls, would not hinder the successful implementation of our framework.

In general, while we allow \ac{Scan-Map deviations}, we presume that the reference map remains a reliable map suited for localization, i.e., the BIM model or reference point cloud has enough features that comply geometrically with the current state of the environment.

%     (5/5) Overview of the Paper: Proposed framework and reading guide
To address the previously described research gaps, we present SLAM2REF, a novel framework that integrates 3D LiDAR data and \ac{IMU} measurements with a reference map to achieve precise pose estimation, enabling also map extension and long-term map management.

The three key components and functionalities of our framework are the following:

\begin{itemize}
\item An automatic method that enables the creation of accurate \ac{OGM}s and 3D session data from large-scale building information models (BIM models) or point clouds. 

\item A pipeline that leverages fast place recognition and multi-session anchoring to allow the alignment and correction of drifted session acquired with \ac{SLAM} or LiDAR-inertial odometry systems. Provided that the reference map is accurate enough, the framework allows the retrieval of the 6 \ac{DoF} poses of the entire trajectory, also enabling map extension, and surpassing state-of-the-art methods such as the one introduced by \citep{trzeciak2023conslamExtension}.

\item A module that allows the analysis of the acquired aligned data, providing not only positive but also negative difference detection for an updated 3D map visualization.
\end{itemize}

We demonstrate the effectiveness of SLAM2REF through extensive experiments in various large-scale indoor \ac{GPS}-denied real-world scenarios, showcasing its ability to achieve centimeter-level accuracy in trajectory estimation and robust map alignment over extended periods.
Additionally, we demonstrate that the method enables the robust automatic alignment of the data with a reference \ac{BIM model}, which does not contain clutter, furniture, or dynamic elements as the real-world data.

This is achieved through innovative feature descriptors based on the widely used Scan Context descriptor \citep{kim2021scan} and a novel YawGICP registration algorithm built based on the Open3D \ac{GICP} method. Additionally, we incorporate motion distortion correction of individual scans by integrating \ac{IMU} measurements to create continuous-time trajectories inspired by the Direct LiDAR Inertial Odometry system \citep{chen2023directDLIO}. These elements are holistically integrated into a multi-session anchoring framework that enables the registration of drifted SLAM session data with a reference map. 

While our framework draws significant inspiration from LT-SLAM \cite{kim2022ltMapper}, our method is able to retrieve ground truth poses when an accurate reference map is available. Furthermore, our method incorporates motion distortion correction and is well-suited for indoor scenarios. It also can utilize any 3D map, such as point clouds or BIM models, as a reference, thus not being limited to the registration of session data pairs.

% The remainder of this paper is structured as follows.
The following is the structure of the remainder of this paper.

%% 2. Theoretical background
Section \ref{chap:theory} introduces the factor graph problem formulation of \ac{SLAM} as well as of multi-session anchoring to align different sessions in a unified coordinate system.

%% 3. Related work
Section \ref{chap:related_work} covers work on map-based \ac{LiDAR} localization and mapping. 

%% 4. Method
% Several algorithms used for 2D-LiDAR localization and 
Section \ref{chap:methodology} introduces our modular SLAM2REF framework, divided into three main steps: \textbf{Step 1.} Generation of \ac{SD} from a reference map, \textbf{Step 2.} Introduces the reference map-based multi-session anchoring method, which allows the alignment and correction of new session data with the reference map and \textbf{Step 3.} Change detection and meshing of new or removed elements in the environment.

%% 5 and 6. Experiment and results
Section \ref{chap:experiments} explains the experimental parameters and implementation details, followed by the results and analysis in section \ref{chap:results}. 

%% 7 and 8. Discussion and limitations
Sections \ref{chap:disccusion} and \ref{chap:limitations}, present the discussion and limitations related to the proposed pipeline and results.

%% 9 and 10. Conclusions and future research

Finally, sections \ref{chap:conclusions} and \ref{chap:futurework} summarize what we have accomplished and bring our work to a close by discussing possible future research directions.

% 2: Theory
\section{Theoretical background}
\label{chap:theory}

Before presenting the current state-of-the-art methodologies, an introduction to the theoretical concepts behind localization and mapping algorithms, as well as the multi-session anchoring process employed in this research, is presented. 
For better understanding, a table with all mathematical variables and the corresponding description can be found in the appendix \ref{appendix}.

% %-------------------------------------------------
% % Problem definition
% %-------------------------------------------------

In multi-session anchoring, similar to \ac{SLAM} or a tracking scenario, the objective is to optimize the posterior probability of the poses in a trajectory based on collected measurements.
In other words, we aim to find the poses for which the provided measurements have the highest probability.

However, in multi-session anchoring, we also aim to find the best alignment between sessions. 
Each session consists of successive sensor data collected from a specific location at varying time intervals.

%-------------------------------------------------
% Factor graph -> Why? 
%-------------------------------------------------
These types of problems can be formulated as a \ac{MAP} estimate that maximizes the posterior density $p(X|z)$ of the states $X$ given the measurements $Z$.
Instead of using Bayes Net, the problem can be considered as a factor graph factorization in which each factor is proportional to a conditional probability density.

While Bayesian nets provide a practical modeling framework, factor graphs facilitate rapid inference. 
Like Bayesian networks, factor graphs enable the representation of a joint density as a product of factors \citep{dellaert2017factor}.

In robotics, various challenges, including pose estimation, planning, and optimal control, often involve solving optimization problems. 
These problems typically center around maximizing or minimizing objectives composed of numerous local factors or terms specific to small subsets of variables. 
Factor graphs allow the encapsulation of this local structure, with factors representing functions related to subsets of variables \citep{dellaert2021factor}.

%-------------------------------------------------
% Factor graph formal definition 
%-------------------------------------------------
A factor graph $F=(\mathcal{U}, \mathcal{V}, \mathcal{E})$ comprises nodes connected by edges $e_{i j}\in{\mathcal{E}}$.
The nodes can be of two types: factors $\phi_{i}\in \mathcal{U}$ and variables $x_{i}\in{\mathcal{V}}.$  
The factor graph represents the factorization of a global function, where each factor is a function of the variables in its adjacency set. 
Given that $\textstyle X_{i}$ is the group of variables $x_{i}$ connected to a factor $\phi_{i}$, a factor graph specifies the factorization of a global function $\phi(X)$ as
$$
\phi(X)=\prod_{i}\phi_{i}(X_{i}). 
$$
Stated differently, each factor $\phi_{i}$ relies solely on the adjacent variables $\textstyle X_{i}$ and is connected to other factors via the edges $e_{i j}$.

%-------------------------------------------------
% Factor graph for SLAM
%-------------------------------------------------
An elegant representation of a SLAM problem is called \textit{pose SLAM}, which eliminates the need to directly include landmarks in the optimization process. 
The focus of pose SLAM is to predict the robot's trajectory based on constraints from odometry and loop closures between the different poses in a trajectory \citep{juric2021Comparison}. 
These odometry constraints, describing the relative poses, can be derived from various sources (e.g., camera or wheel encoders); in this case, we use \ac{IMU} and \ac{LiDAR} measurements, as it will be described later in \ref{substep:2_1}.

%----------------------------------------------------------------------------------
% Factor graph as non-linear least square problem 
%-----------------------------------------------------------------------------------
In general, \ac{MAP} inference involves maximizing the product of all factor graph potentials for any arbitrary factor graph \citep{dellaert2017factor}.
$$
X^{\operatorname{MAP}}=\underset{x}{\operatorname{argmax}}\ \prod_{i}\phi_{i}(X_{i}).  \label{eq:map_product}
$$
Assuming that all factors can be modeled by a measurement function $h_{i}$, with normally distributed priors and factors from measurements $z_{i}$ with zero-mean Gaussian noise models $\Sigma_{i}$, then we have the following conditional density $p(z_{i}|x_{i},l_i)$ on the measurement $z_{i}$.  

$$
p(z_i|x_i,l_i)={\mathcal{N}}(z_{i};h_{i}(x_i,l_i),\Sigma_{i})={\frac{1}{\sqrt{|2\pi \Sigma_{i}|}}}\exp\left\{-{\frac{1}{2}}\left|\left|h_{i}(x_i,l_i)-z_{i}\right|\right|_{\Sigma_{i}}^{2}\right\}. 
$$
Thus, we face factors that are proportional to:
$$
\phi_{i}(X_{i})\propto\exp\left\{-{\frac{1}{2}}\left\|h_{i}(X_{i})-z_{i}\right\|_{\Sigma_{i}}^{2}\right\}, \label{eq:gaussian_proportion}
$$
Taking the negative log of Eq.~(\ref{eq:map_product}) and dropping the factor 1/2 allows us to instead minimize a sum of non-linear least squares:

\begin{equation}
\begin{aligned} 
X^{\operatorname{MAP}}
& =\underset{x}{\operatorname{argmin}}-\log \ \prod_{i}\phi_{i}(X_{i}). \\
& =\underset{x}{\operatorname{argmin}}\sum_{i}\left\|h_{i}(X_{i})-z_{i}\right\|_{\Sigma_{i}}^{2}. \label{eq:map_sum}
\end{aligned} 
\end{equation}
%MV: Because The log of a product is the sum of the logs, and the log of exp(X) is simply X.

%-------------------------------------------
% Multi-session anchoring
% Encounters -> measurement equation
%-------------------------------------------
In the context of multi-session anchoring, inter-session, or between sessions, loop closure detections, also called \textit{encounters} $\mathbf{c}$  (which are also poses in the special Euclidean group $\mathrm{SE}(3)$), can be added to the non-linear least squares formulation in Eq.~(\ref{eq:map_sum}) with the following Gaussian measurement equation:

$$
\mathbf{c}=h\left(\mathbf{x}_{R}, \mathbf{x}_{Q}\right)+ \eta,
$$
% MV: given the measurements in the poses X_R  and X_q, we can estimate an encounter (pose in SE_(3)) in the global frame %% see Chapter 6 of Probabilistic robotics
where $h{\big(} .{\big)}$ is a relative measurement prediction function, and $\eta$ is a normally distributed zero-mean measurement noise with covariance $\Sigma_c$. 
Furthermore, $\mathbf{x}_{R}$ and  $\mathbf{x}_{Q}$ are the sensor poses in the two sessions $\mathcal{S_R}$ and  $\mathcal{S_{Q}}$, respectively.
This yields the following conditional density $p( \mathbf{c}|\mathbf{x}_{R},\mathbf{x}_{Q})$ on the measurement $\mathbf{c}$  
$$
\begin{aligned}
p(\mathbf{c}|\mathbf{x}_{R}, \mathbf{x}_{Q})
&={\frac{1}{\sqrt{|2\pi \Sigma_c|}}}\exp\left\{-{\frac{1}{2}}\left|\left|h(\mathbf{x}_{R},\mathbf{x}_{Q})-\mathbf{c}\right|\right|_{\Sigma_c}^{2}\right\}. 
\end{aligned}
$$

Similarly, an odometry model $f{\big(} . {\big)}$, which usually incorporates a scan-matching process, among other techniques, produces constraints $\mathbf{u}_i^s$ between consecutive poses: $\mathbf{x}_{i}$ and $\mathbf{x}_{i+1}$.

Unifying the encounter measurement model $h{\big(}. {\big)}$ together with the odometry model $f{\big(}. {\big)}$ in Eq.~(\ref{eq:map_sum}), we obtain the following equation (omitting intra-session loop closures for simplicity).

\begin{equation}
\begin{aligned}
X^{\operatorname{MAP}}&= \underset{x}{\operatorname{argmin}} \left\{\sum _{\mathcal{S}} 
\left( \left\|\mathbf{p}_s-\mathbf{x}_{s,0}\right\|_{\Sigma_P}^2
+\sum_{i \in M_s}\left\|f_i\left(\mathbf{x}_{s,i}, \mathbf{u}_{s,i}\right)-\mathbf{x}_{s,i+1}\right\|_{\Sigma_O}^2\right) \right. \\
&+\left.\sum_{j \in N_e }\left\|h_j\left(\mathbf{x}_{R, j}, \mathbf{x}_{Q, j}\right)-\mathbf{c}_{j}\right\|_{\Sigma_c}^2\right\} 
\end{aligned}
\label{eq:least_sqrt}
\end{equation}

Where $\mathcal{S}\in \{\mathcal{S}_Q, \mathcal{S}_{R}\}$, $M_s$ is the number of poses in the session $\mathcal{S}$, and $N_e$ is the number of encounters between sessions.

Here, we directly incorporate the initial pose of each session as a prior factor $\mathbf{p}_s$. This fixes the initial pose to the origin, effectively eliminating that gauge of freedom, i.e., assigning a local reference coordinate system to each session.
 
As in a multi-robot mapping problem, having two sessions or more requires a strategy to handle the fact that the sessions can have different initial poses and, therefore, other initialization prior \citep{Lajoie:2024:Swarm_SLAM}.

%-------------------------------------------------
%  Anchor nodes!
%-------------------------------------------------
We employ anchor nodes to address this problem and facilitate the integration of inter-session constraints.

%% Anchors are:

The anchor $\Delta_Q$ is a $\mathrm{SE}(3)$ pose for the session $\mathcal{S}_Q$ that determines how the entire trajectory is positioned concerning a global coordinate frame. 

Essentially, we maintain the individual pose graphs of each session in their respective local frames and bind them with anchor factors to the global frame.
For each session, an anchor node is added to the pose graph problem as the first pose of the session; this pose can be selected arbitrarily (usually set to the origin).  

During the initial encounter, no modifications are made to the pose graphs of the respective sessions; only the anchor nodes change, bringing both graphs to a global coordinate system where they can be compared.
In subsequent encounters, information can propagate between the two pose graphs, similar to the scenario of loop closures in a single session. 
The incorporation of anchor nodes makes efficient updates and quick optimization feasible.

%-----------------------------------------
% Anchors are helpful for:
%-----------------------------------------
As described by \citet{kim2010multiple}, the anchor nodes allow us to estimate the offset between sessions.
Moreover, they provide faster convergence to least-squares solvers and allow each session to optimize their poses before considering global constraints, such as from inter-session loop closures \citep{ozog2016long}.

This feature is advantageous for \textit{long-term mapping} since it enables the production of the first consistent map of the environment when the data is gathered.
Whenever a map containing a new session is constructed in a posterior period, and at least one encounter is detected, the anchor nodes allow the computation of the transformation that aligns this recent session with the previously acquired session. 
Subsequent inter-session loop closure detections will allow correction and improvement of both sessions. 

Now that we conclude the introduction to the theory behind the selected method to align two or multiple sessions, in the following session, the latest \ac{SOTA} methods to achieve this alignment with a reference map, with particular emphasis in \ac{BIM model}s will be summarized.

% 3: Related work
\section{Related research}
\label{chap:related_work}

This section will provide an overview of the state-of-the-art approaches that allow this alignment by using prior building information, such as \ac{BIM model}, floor plans or point clouds, and methods supporting mapping.

\subsection{Map-based 2D \ac{LiDAR} localization and mapping}

\citeauthor{Follini.2020}\citeyear{Follini.2020} show how the standard \ac{AMCL} technique may be utilized to obtain the transformation matrix between the robot reference system and an extracted 2D map from the \ac{BIM model}. 
They also state that the \ac{AMCL} algorithm could overcome small objects that are not present in the \ac{BIM model} due to the probability distribution of its beam model.

The same technique was applied by \cite{Prieto.2020}, \cite{Kim.2021}, \cite{Karimi.21.04.2021}, and \cite{Kim.2022} to localize a wheeled robot in a 2D \ac{OGM} produced from a \ac{BIM model}. The primary distinction between these strategies is how they extract the \ac{OGM} from the \ac{BIM model}. 

An \ac{OGM} discretizes the environment into 2D square cells with a predetermined resolution; the value in each cell reflects the likelihood that an obstacle occupies the cell. Thus, an \ac{OGM}  allows distinguishing whether a space is free, occupied, or undiscovered. 

\cite{Prieto.2020} make use of the geometry of the spaces in the \ac{IFC} file as well as the location and size of each opening, in contrast to \cite{Follini.2020}, who use the vertices of elements that intersected a horizontal plane and the Open CASCADE viewer to create an \ac{OGM} in \textit{pgm} format.

\cite{Karimi.2020} created Building Information Robotic System (\acs{BIRS}), an ontology that allows the generation and transfer of topological, semantic, and metric maps from a \ac{BIM model} to \ac{ROS}.  
An optimal path planner was included in the tool in \citep{Karimi.21.04.2021}, incorporating crucial elements for the evaluation of the construction. 
However, this method still does not incorporate Mechanical, Electrical, and Plumbing (MEP) equipment.

A technique to transform an \ac{IFC} file into a \ac{ROS}-compliant \ac{SDF} world file appropriate for robot job planning was implemented by \cite{Kim.2021}. They evaluated their strategy for an automatic painting of interior walls. The prototype includes a converter that generates a \ac{ROS}-compliant world file from \ac{IFC} file and subprocesses that perform localization, navigation, and motion planning.

Later, a method to turn an \ac{IFC} model into an \ac{URDF} building environment was proposed by \cite{Kim.2022} in order to add dynamic objects and for the purpose of door inspection. From this point, a robot may directly access lifecycle information from the \ac{BIM model} for job planning and execution. Once they have the \ac{URDF} model, they use PgmMap \citep{Yang.2018} to extract an \ac{OGM} from it.

For 2D-\ac{LiDAR} localization, \citet{Hendrikx.2021} propose a method that uses a robot-specific world model representation taken directly from an \ac{IFC} file rather than from an \ac{OGM}.
In their factor graph-based localization strategy, the system receives information about the lines, corners, and circles in the immediate environment of the robot and builds data linkages between those items and the laser readings. 
They updated and assessed their approach for global localization in \citep{hendrikx2022local}, producing superior results when compared to \ac{AMCL}. 

\cite{Boniardi.2017} uses an architectural floor plan based on \ac{CAD} rather than a \ac{BIM model}. They use a \ac{GICP} implementation for scan matching together with a pose graph \ac{SLAM} system in their localization and mapping system. 
They transform a \ac{CAD} floor plan into a 2D binary image and use it for robot localization in a wear-house-like scenario. 

Later, they suggested an improved pipeline that outperformed \ac{MCL} in the pose tracking problem for long-term localization and mapping in dynamic situations \cite{Boniardi.2019}.

In one of our previous contributions \citep{vega:2022:2DLidarLocalization}, we proposed a method to create an \ac{OGM} from a multistory \ac{IFC} Model. 
Furthermore, we showed that the commonly used \ac{AMCL} is not as resistant to change and dynamic environments as compared to \ac{GBL} methods, such as Cartographer \citep{Hess.2016b} and SLAM Toolbox \citep{Macenski.2021}. 
Based on these findings, we also offered an open-source approach that transforms \ac{OGM} to \ac{PGBM} for reliable tracking of robot poses. 
This method was released to ease the transition of the localization algorithms from the classical \ac{PF} to more robust \ac{GBL} methodologies, similar to what happened with the development of the SLAM algorithms.

%%%%%%%%%%%%%%%%%%%%%%%%%%%%%%%%%%%%%%
\subsection{Map-based 3D \ac{LiDAR} localization and mapping}

Other approaches investigated 3D \ac{LiDAR} localization using 3D reference maps.

\cite{gawel2019fully} presented a very accurate robotic building construction system. They use ray tracking with three \textit{laser distance sensors}, a 3D \ac{CAD} model, and a robust state estimator that merges \ac{IMU}, 3D \ac{LiDAR}, and wheel encoders to locate the end-effector with subcentimeter precision. 
They did this by taking several orthogonal range measurements while the robot was static.

In the technique proposed by \cite{ErcanJenny.2020} and \cite{Blum.2020}, the 3D \ac{LiDAR} scan is aligned with the \ac{BIM model} using the \ac{ICP} algorithm. 

While \cite{ErcanJenny.2020} limits the alignment to a few carefully chosen reference-mesh faces to overcome ambivalence, \cite{Blum.2020} uses picture information to separate the foreground and background in the point cloud and uses only the latter for registration. 
The pipeline was then extended to provide a self-improvement semantic perception technique that can better handle environmental clutter and increase accuracy \citep{Blum.2021}.

To take advantage of the high performance of Google Cartographer \citep{Hess.2016b} for localization, \cite{Moura2021} suggest a method to create \textit{.pbstream} maps from \ac{BIM model}s.
Although this approach is quite practical, since they only employ Cartographer in localization mode, their method does not create a map of the environment if the robot is not localized and inside the boundaries of the reference map.

\cite{Oelsch2021} propose Reference-LOAM (R-LOAM), a technique that uses a combined optimization that includes point and mesh characteristics for 6 \ac{DoF} \ac{UAV} localization. Later, in \citep{Oelsch.2022}, they improved their approach using pose-graph optimization to decrease drift even when the reference object is not visible.

A semantic \ac{ICP} approach was presented by \cite{YIN_2023_104641}. This method uses the 3D geometry and semantic data of a \ac{BIM model} to achieve a reliable 3D \ac{LiDAR} localization method. Their system suggests a BIM-to-Map conversion, turning the 3D model into a point cloud that is semantically enhanced. 
Their research demonstrates that a 3D \ac{LiDAR}-only localization can be accomplished using an \ac{BIM model} in uncluttered environments.

Another exciting strategy, suggested by \cite{shaheer2022robot}, relies on geometric and topological information in the form of walls and rooms rather than object semantics for localization. 
They build Situational Graphs (S-Graphs) using these data, which are subsequently used for precise pose tracking. 
Later, they improved their technique by allowing the acquisition of a map before localization, as well as the posterior matching and merging with an A-graph (extracted from \ac{BIM model}s). 
The combined map's ultimate designation was an informed Situational Graph (iS-Graph) \citep{shaheer2023graphbased}.

Direct \ac{LiDAR} localization (DLL) is a fast localization method introduced by \cite{caballero2021dll}. 
They use a registration method based on non-linear optimization of the distance between the points and a reference point cloud. 
Their method does not require feature extraction to achieve an accurate and fast registration.
By correcting the anticipated pose using odometry, the technique can follow the robot's pose with subdecimeter precision in real-time. 
Their technique performed better compared to \ac{AMCL} 3D \citep{amcl3d:2017}.

%%%%%%%%%%%%%%%%%%%%%%%%%%%%%%%%%%%%%%%%%%%5
%% Research gap 
Numerous methods have been developed that use reference 2D and 3D maps for \ac{LiDAR} localization and mapping. 
Most of them have concentrated on real-time localization without enabling pose-graph-based optimization approaches to provide a more accurate estimation of the calculated poses.

Additionally, practically every method requires the scanning to begin in a known initial pose that must be inside the boundaries of the reference map.

This requirement means that for several methods, there is no chance of localization or the generation of an aligned map if the robot starts from a location where the reference map is not visible or from where there are large \ac{Scan-Map deviations}, like in a cluttered environment.

Furthermore, rather than retrieving a posterior accurate, updated, and extended map of the environment and detecting environmental changes, most researchers focused only on improving the accuracy of the pose-tracking process.

In this paper, we provide a strategy that addresses these problems and show that it is feasible to create an aligned, optimized map that is near the ground truth and discover changes in the environment.

% 4: Methodology
\section{Methodology}
\label{chap:methodology}

\begin{figure}[!htb]
    \centering
    \includegraphics[width=\textwidth]{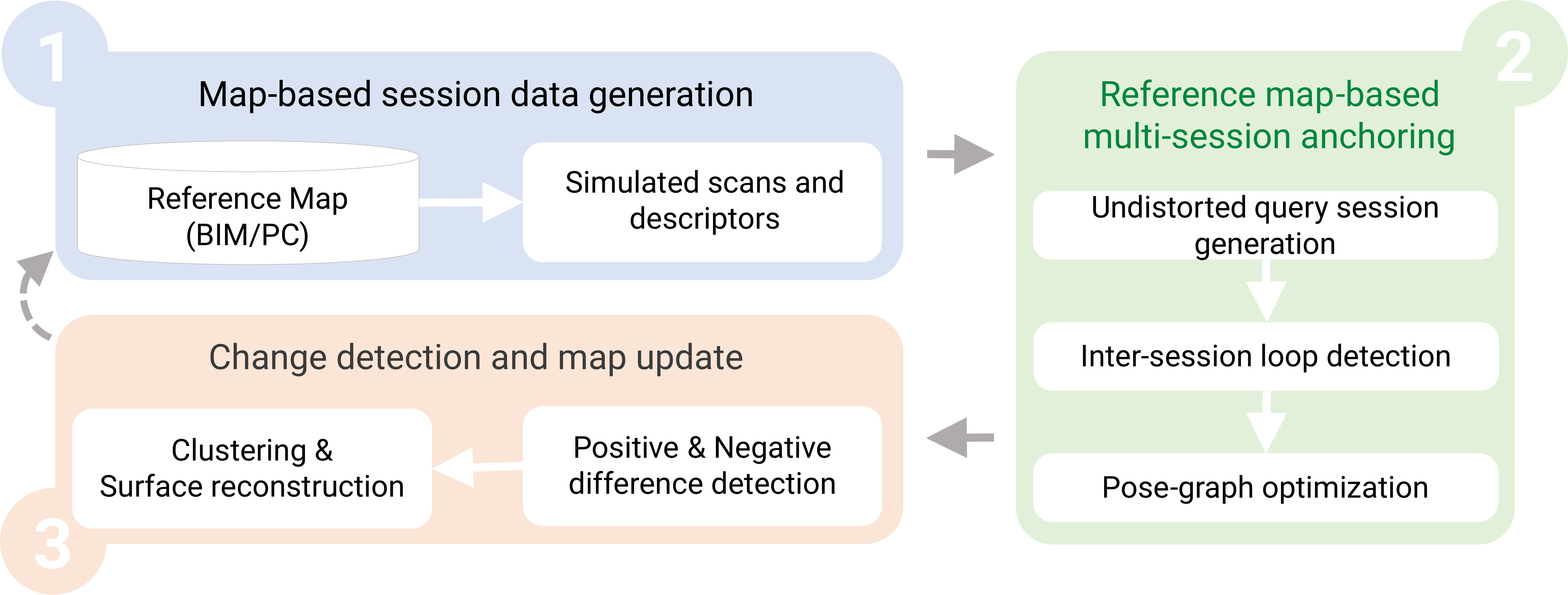}
    \caption{Overview of \textbf{SLAM2REF}. The pipeline consists of three steps: map-based session data generation, Reference map-based multi-session anchoring, and Change detection and map update.}
    \label{fig:method}
\end{figure}

Our approach can be broken down into three key components, as shown in Figure  \ref{fig:method}. 
In \textbf{Step 1}, synthetic reference \ac{SD} is generated automatically from large-scale 3D reference \ac{BIM model}s or point clouds.

Then, in \textbf{Step 2}, a real-world undistorted LiDAR \ac{SD} acquired using a state-of-the-art \ac{LIO} algorithm is aligned and corrected using the reference 3D map.

Finally, in \textbf{Step 3}, the aligned map is further automatically analyzed, allowing the creation of an updated 3D map, which considers the detection of positive and negative environmental changes.

% Step 1. Map-based session data generation
\subsection{Map-based session data generation (Map to Session Data)}
\label{step:1}

In this step, our objective is to encapsulate the geometry of the reference 3D map—whether it is a \ac{BIM model} or a point cloud—into individual LiDAR scans with their corresponding feature descriptors. 
These descriptors serve to encode the visible geometry from the origin of the scan within the reference map, enabling us to rapidly find the correct alignment of real-world session data with a reference map.

In real-world data acquisition, \acl{SD} (SD) refers to consecutive sensor data acquired from a particular place at different periods \citep{cramariuc2022maplab}. 
Nonetheless, since we aim to convert a reference map to synthetic \ac{SD}, these data can be considered a set of LiDAR scans (with known carefully selected positions) and their corresponding descriptors.

Formally, a session $\mathcal{S}$ is defined as follows:

\begin{equation}\label{eq:1}
\mathcal{S}:=\left(\mathcal{G},\left\{\left(\mathcal{P}_i, d_i\right)\right\}_{i=1, \ldots, n}\right)
\end{equation}

Here, $\mathcal{G}$ is a pose-graph map that contains the coordinates of the pose nodes, odometry edges, and optionally recognized intra-session loop edges with uncertainty matrices. 
These matrices represent how certain the positions of these edges are.  
This map can be saved in a text file, usually in \textit{.g2o} format.

The $\left(\mathcal{P}_i, d_i\right)$ are the pairs of 3D \ac{LiDAR} scans with their corresponding global descriptors of the $i^{th}$ keyframe and $n$ is the total number of equidistantly sampled keyframes.

% Overview
Generating synthetic \ac{SD} (simulated scans and descriptors) from a reference map can be subdivided into three substeps.
% 1
First, an \ac{OGM} is extracted from the reference map. This extraction is achieved in an automated manner, taking as input only the \ac{IFC} model or the reference point cloud and the floor level (z coordinate value) from where the \ac{OGM} should be generated.
% 2
In a second substep, the \ac{OGM} is used to find the poses in which the \ac{LiDAR} scans will be simulated. 
% 3
In a third and final substep, \ac{LiDAR} scans are rapidly simulated in the positions calculated in the previous step, and the corresponding descriptors are calculated.

These substeps have been optimized so that it is possible to efficiently simulate data from large-scale 3D BIM models and point clouds.
The following subsections provide a more detailed explanation of each substep.

% % % % % % % % % % % % % % % % % % % % % % % % % % % % % % % % % % % % % % % % % % % % % v
% Substep 1
\subsubsection{OGM from reference map}
\label{step:1_1}
% Why an OGM?
Initially, and for convenience, the 3D geometry of the reference map is reduced into a 2D \ac{OGM}. This dimensional reduction has been demonstrated to be very computationally efficient, allowing the implementation of the pipeline in complex, large-scale models.

Moreover, a 2D \ac{OGM} (with known scale and origin) allows the direct usage of the map with the \ac{ROS} navigation stack for autonomous navigation \citep{Macenski_2023}. Besides path planning, cost maps, and navigational algorithms, the \ac{ROS} navigation stack includes several state-of-the-art features, such as the regulated pure pursuit algorithm to adjust the robot's speed depending on the path with a particular focus on safety in constrained and partially observable spaces \citep{macenski2023regulated}.

The method for creating \ac{OGM} varies based on the input data. Here, we outline how to do it for BIM models and point clouds. 

\subsubsubsection{OGM from IFC model (BIM2OGM)}

The proposed automated generation of \ac{OGM}s from \ac{BIM model}s builds upon prior work described in \citep{vega:2022:2DLidarLocalization}. 
However, the key distinction lies in the enhanced automation of the pipeline.

% Tools 
For this purpose, we leverage the IfcConvert \citep{IfcConvert} tool and employ image-processing techniques akin to previous related works. IfcConvert, a command-line interface application within the open source IfcOpenShell project \citep{krijnen2015ifcopenshell}, facilitates the versatile conversion of a 3D \ac{BIM model} from the \textit{.ifc} file format to various other formats such as 3D meshes (\textit{.obj}, \textit{.dae}) or 2D layers (\textit{.svg}). Detailed documentation for the IfcConvert functionality is available \citep{Ifcconvert_documentation}. 

%% Overview of the process 
The input 3D \ac{IFC} model is first converted to \ac{SVG} format and then processed with the OpenCV library to output different layers as \ac{PNG} files. 
These layers will then be merged to produce the final \ac{PGM}. 
 
To ensure compatibility with the \ac{ROS} navigation stack and facilitate accurate scan simulations, the 2D \ac{PGM} map must adhere to specific guidelines. It should represent unknown (external) regions in gray, navigable space (floor) in white, and potential collision-causing objects (e.g., walls and columns) in black.

IfcConvert is used to convert the 3D \ac{IFC} model into 2D \ac{SVG} files with the desired elements intersecting a plane at the chosen height. 
Furthermore, the resolution and size of the output \ac{SVG} image are modified to only include the elements of interest.

To generate the \ac{OGM}, we leveraged the semantics of the \ac{BIM model}, focusing on extracting permanent elements like walls, ceilings, columns, and floors. 
This process excludes non-permanent features and objects invisible to \ac{LiDAR} sensors, such as spaces, doors, windows, and curtain walls.

Filtering just permanent structural information about the building enables finding reliable correspondences between the geometry from the BIM model and real-world 3D \ac{LiDAR} data. 
In this letter, we assume that the permanent structures in the BIM model are reliable features for localization and scan-matching. 
In the presence of open doors and windows, their exact placement in the space is unknown (open, closed, or semi-open) and is not provided in the BIM model; therefore, those elements should not be considered while creating the 2D \ac{OGM} or any source of information used for alignment or localization.

A critical consideration in the conversion of \ac{SVG} files to \ac{PNG} format is the choice of units utilized within the original \ac{SVG} file. By default, IfcConvert assigns millimeters as the unit of measurement for the \ac{SVG} files. 
However, these millimeters do not undergo a direct one-to-one transformation to pixels during the conversion to \ac{PNG}. 
Consequently, it becomes imperative to eliminate explicit unit specifications within the \ac{SVG} file to ensure consistent scaling and preservation of the established coordinate origin during the conversion to \ac{PNG}.

Additionally, it is critical to consider the effect of displacement while creating sections at different heights. 
While the scale will be maintained, the values of the coordinates of the geometry (saved in \textit{paths}) in the \ac{SVG} file will be adjusted according to the elements that intersect that specific height. 
To counteract this effect and have all the \ac{PNG} images in the same coordinate system, the images are shifted according to the $x$ and $y$ values saved in the \textit{data matrix} of the \ac{SVG} generated with IfcConvert.

Automating the creation of the \ac{OGM} involves producing the following two sections:
\begin{enumerate}
    \item In the indoor layer, the floor area is designated as white. 
    This section is generated at the z-coordinate corresponding to the upper surface of the slab of interest, i.e. at the floor level where the alignment should happen. Subsequently, the resultant gray-scale \ac{PNG} image from this \ac{SVG} is converted to binary. Then, its inverted version represents the indoor layer, in which the floor is represented as white pixels and the rest as black.
   
    \item In the collision layer, we extract permanent elements like walls and columns while excluding non-permanent structures such as doors and windows. 
    The creation of this layer occurs slightly above (1 m) the z-coordinate of the previous layers.
    It is crucial to note that the coordinate system of this image deviates from the preceding layer due to its creation at a different height. Therefore, it is imperative to compensate for this offset, as previously explained, before converting it into \ac{PNG} format.
    
\end{enumerate}

Subsequently, the indoor layer is placed over a gray image of the same size, allowing to distinguish outdoor (unknown) and indoor areas. 

\begin{figure}[!htb]
    \centering
    \includegraphics[width=0.9\textwidth]{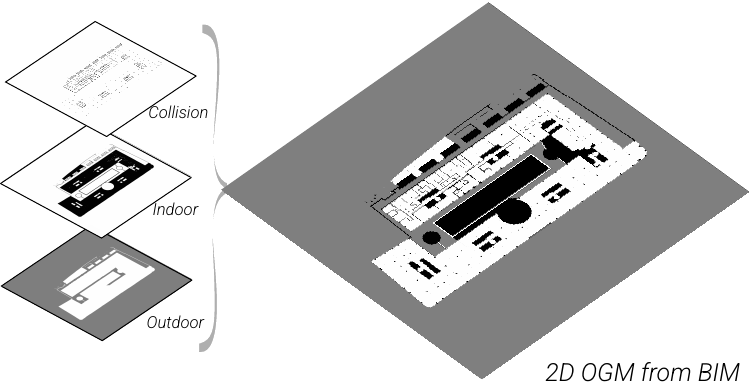}
    \vspace*{5mm} 
    \caption{Generated \ac{OGM} from the \ac{BIM model}. On the left, the different layers, and on the right, the merged final OGM.}
    \label{fig:OGM}
\end{figure}

Finally, the pixels in the black color of the collision layer are transferred to the indoor layer.
Given this, the final \ac{OGM} is created and saved in the rasterized \ac{ROS} standard \ac{PGM} format. Figure \ref{fig:OGM} illustrates the layers and the final 2D map.

Additionally, a corresponding \ac{YAML} configuration file is generated, containing crucial details such as the origin and resolution of the 2D map, extracted from the \textit{data-matrix} of the initial \ac{SVG} file.

Besides being an essential step in our pipeline, accurately creating a 2D \ac{OGM} holds significant potential for \ac{SOTA} localization algorithms, facilitating rapid and collision-free autonomous navigation. 
This has been exemplified by \cite{vega:2022:2DLidarLocalization} and corroborated by numerous other studies (refer to Section \ref{chap:related_work}).

\subsubsubsection{OGM from a point cloud}

The steps involved in creating a \ac{OGM} from a point cloud are as follows:
First, a 2D grid is created to the length and width of the point cloud and scaled given a grid resolution.
Each cell within this grid is initially assigned a gray color.

Then, and as discussed in \citep{Vega:2022:ObjectDetection}, the points are projected onto the XY plane, considering the resolution of the grid and its origin (the minimum XY coordinate of the point cloud). If points within a cell are found to be near the floor level (within a range of ±0.5 m), the cell is colored white, signifying navigable space.

On the other hand, cells are colored black if points are detected at a height 1 m above the floor level, assuming that this region predominantly consists of walls, columns, and other permanent elements.

% % % % % % % % % % % % % % % % % % % % % % % % % 
% Substep 2
\subsubsection{Locations for data simulation}
% OGM -> skeleton -> MW-poses -> downsampling -> KNN

Once a correct \ac{OGM} is generated from the reference map, this is utilized to find proper locations where LiDAR scans will be simulated. 
These locations should be equally separated coordinates ordered by proximity, aiming to closely replicate real-world data acquisition with full coverage of the map.
To this aim, we first extract the skeleton of the image, which gives a smooth path similar to the one a person would follow during acquisition with a mobile LiDAR or scanning device. Then, points are sampled over this path in a uniform manner.

Similarly, as proposed in \citep{vega:2022:2DLidarLocalization}, the process extracts a skeleton from the \ac{OGM}. This skeleton is derived using the approach outlined by \cite{lee1994building}, producing a smooth trajectory over the free space that interconnects all rooms and open areas within the \ac{OGM}.

In a previous version of our pipeline \citep{vega:2023:BIM_SLAM}, we used a Wavefront Coverage \ac{PP} \citep{Zelinsky.1993} over this skeleton to find the waypoints in which the 3D LiDAR will be simulated. 
However, the Wavefront Coverage \ac{PP} approach is inherently intricate, making it unfeasible to be applied over large \ac{OGM}s without consuming large amounts of computational resources.

Therefore, to handle large-scale reference maps, we propose the following method instead, which tries to sample uniformly key points over the path created with the skeleton approach:

(1) The scan locations are initially extracted using image processing techniques. This involves generating masks with equally spaced vertical and horizontal lines, isolating only the white pixels intersecting these masks and the previously generated skeleton. The idea behind this is that only isolated pixels will remain rather than elongated lines present in the skeleton.

(2) Subsequently, the corresponding center points of the remaining pixels are extracted using a contour detection algorithm. To ensure a minimum distance between points, the spatial distribution of these coordinates is downsampled.

(3) Finally, the coordinates are sequentially ordered using the nearest neighbor algorithm.

\begin{figure}[!htb]
    \centering
    \includegraphics[width=0.9\textwidth]{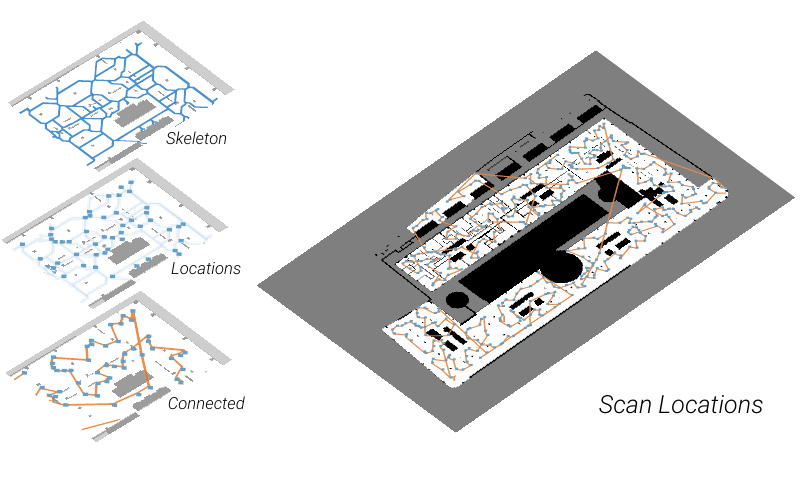}
     % for2columns width=0.48
    \caption{Calculated locations for scan simulation. On the left are the main steps, and on the right are all the calculated positions in the entire OGM.}
    \label{fig:s1_2_ScanLocations}
\end{figure}

Figure \ref{fig:s1_2_ScanLocations} shows the calculated scan locations for an OGM of a large building.

% % % % % % % % % % % % % % % % % % % % % % % % % 
% Substep 3
\subsubsection{LiDAR data simulation} 
\label{step:1_3}

In our previous work \citep{vega:2023:BIM_SLAM}, we utilized the identified waypoints to set navigational goals for a robot operating autonomously within the \ac{ROS} navigation stack, simulated in the Gazebo physics engine \citep{Gazebo_koenig2004design}. Then, a sequence of simulated 3D \ac{LiDAR} scans was produced with Gazebo and saved in rosbag files.
Here, we present an enhanced approach that eliminates the need for \ac{ROS} or Gazebo; by such means, we avoid the creation of large rosbag files containing redundant information. 

Instead, we propose leveraging \ac{BlenSor}, a versatile software designed for simulating various range scanners \citep{blensor2011, BlenSor_PhdThesis}.  
With the \ac{BlenSor} \ac{API}, we can automatically load the coordinates for simulating LiDAR scans (calculated in the previous step), streamlining the simulation process.

The process of simulating LiDAR data can be subdivided into three main steps: 

\begin{enumerate}
    \item The reference map is converted to an \ac{STL} mesh. 
In the case of a \ac{BIM} model, this involves conversion to \ac{OBJ} format after filtering only permanent structures using IfcConvert, similar to the process employed in creating the 2D \ac{OGM}. However, instead of generating an \ac{SVG} file, our method creates an \ac{OBJ} file containing the 3D geometry of the model described explicitly.
To ensure precise 3D conversion, our approach selectively \textit{includes} required permanent elements (e.g., walls, columns, floors, and slabs) rather than \textit{excluding} entities. Our experiments revealed that the \textit{exclusion} command does not consistently produce satisfactory results for this 3D conversion.
Subsequently, the generated \ac{OBJ} file is converted to \ac{STL} format for seamless integration of the geometry into \ac{BlenSor}.

When dealing with a point cloud as the reference map, the ball pivoting method has consistently demonstrated reliability in reconstructing mesh surfaces from 3D point clouds. 
Before applying this method, the process involves estimating the normals of the point cloud and calculating an optimal radius based on the average nearest neighbor distance, facilitating accurate and efficient surface reconstruction.

\item Later, the coordinates determined in the preceding steps, where the data will be simulated, are transformed from pixels (in 2D) to meters (in 3D). 
This conversion utilizes the scale and origin information specified in the \ac{YAML} file of the corresponding \ac{OGM}. 

\item Subsequently, the simulated \ac{LiDAR} properties are adjusted to align with those employed in real-world scanning. Then, a sub-process initiates the parallel simulation of $360^\circ$ \ac{LiDAR} scans at these coordinates using \ac{BlenSor}.

\end{enumerate}

Finally, and after the simulation, \ac{SC} descriptors are created for each simulated scan. More information about these descriptors will be provided in the following section \ref{substep:2_1} (Step 2.1). 

\begin{figure}[!htb]
    \centering
    \includegraphics[width=0.9\textwidth]{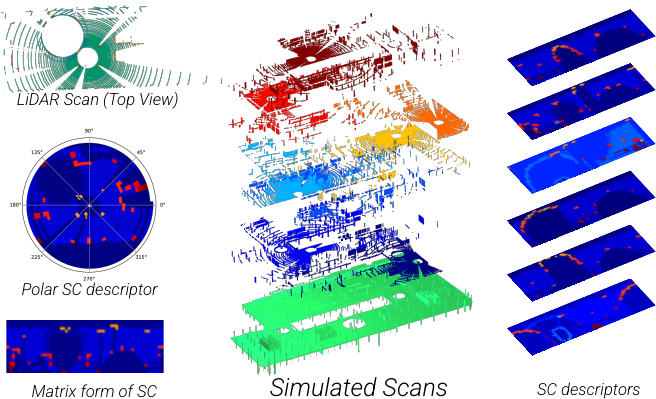}
     % for2columns width=0.48
    \vspace*{5mm} 
    \caption{Synthetic session data from the reference map. On the left, from top to bottom: Top view of one LiDAR scan, its corresponding polar scan context (SC) descriptor, and the descriptor in the matrix form. In the middle, a set of simulated scans and the STL mesh from the BIM model are used. Right, corresponding SC descriptors for the simulated scans.}
    \label{fig:s1_3}
\end{figure}

Following the steps above, the geometry of the reference map or the permanent objects in the \ac{BIM model} is now established as a reference session, denoted as $\mathcal{S_R}$, and is illustrated in Figure \ref{fig:s1_3}. 

In the subsequent step, this synthetic \acl{SD}, encompassing descriptors and simulated scans, will be leveraged for fast place recognition and data alignment.
However, before this process, it is necessary to generate session data from real-world datasets.

% Step 2. Reference map-based multi-session anchoring
\subsection{Reference map-based multi-session anchoring}
\label{step:2}

% Three substeps:
% 2.0. Overview  (here).
% 2.1. Creation of the real-world query session.
% 2.2. Data association or place recognition -> inter-session ISC loop detection for data alignment.
% 2.3. Pose graph optimization, KNN, FinalICP for fine alignment and pose categorization.

To derive a globally consistent map aligned with the reference map from real-world sequential \ac{LiDAR} data, the following three substeps are executed:
(1) Creation of the real-world motion-undistorted query session $\mathcal{S_Q}$, which is similar to the synthetic reference session $\mathcal{S_R}$ (created as explained in the previous section); however, from real-world data.
(2) Place recognition for inter-session loop detection between $\mathcal{S_Q}$ and $\mathcal{S_R}$.
(3) Pose graph optimization with multi-session anchoring and pose refinement with KNN loops and a final ICP registration.
These substeps are described in detail in the following subsections.

\begin{figure}[!htb]
    \centering
    \includegraphics[width=\textwidth]{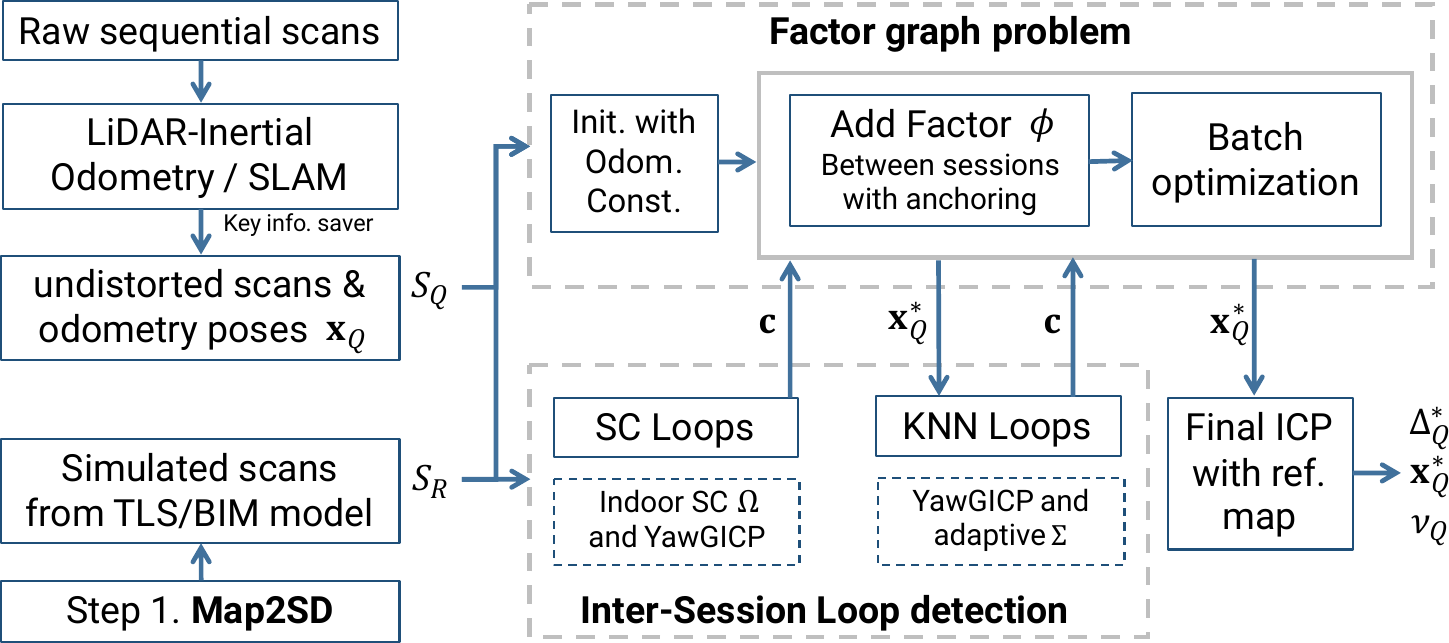}
    \vspace*{1mm} 
    \caption{Comprehensive flowchart illustrating the multi-session anchoring process within \textbf{SLAM2REF}. This process includes the generation of session data from the reference map $\mathcal{S_R}$, creation of the real-world query session $\mathcal{S_Q}$, inter-session loop detection using Indoor Scan Context and YawGICP, and pose refinement with KNN loops and \textit{final ICP}. The outcome includes the anchor node $\Delta_Q^*$, optimized 6-DoF poses $\textbf{x}_Q^*$, and a confidence level list $\mathbf{\nu}_Q$ for each pose in the query session.}
    \label{fig:pgo_flowchart}
\end{figure}

%%%%%%%%%%%%%%%%%%%%%%%%%
%-  overview image -> flow chart
Figure \ref{fig:pgo_flowchart} illustrates a flowchart outlining the complex multi-session anchoring process in the \textbf{SLAM2REF} framework.

% Comprehensive flowchart illustrating the multi-session anchoring process within \textbf{SLAM2REF}. 

Following the generation of session data (SD) from the reference map $\mathcal{S_R}$ (Step 1 presented in Section \ref{step:1}) and the construction of the real-world query session $\mathcal{S_Q}$ (Section \ref{substep:2_1}), the alignment procedure can be initiated. This involves an inter-session loop detection phase employing Indoor Scan Context and YawGICP  (Section \ref{substep:2_2}), which identifies encounters $\mathbf{c}$ denoting correspondences between the sessions. These encounters, along with initial odometry constraints, are integrated into a factor graph problem. Subsequent to optimization, pose refinement is carried out using KNN loops (Section \ref{substep:2_3}) and a \textit{final ICP} process. The resulting information comprehends the following elements attributed to the query session: the anchor node $\Delta_Q^*$, which facilitates the global alignment to the reference map, the optimized 6-DoF poses of each scan $\textbf{x}_Q^*$, and a confidence level list $\mathbf{\nu}_Q$ providing the reliability of each pose after scan registration.

%%%%%%%%%%%%%%%%%%%%%%%%%

%%%%%%%%%%%%%%%%%%%%%%%%%%%%%%%%
%% Step 2.1 Creation of the real-world query session
\subsubsection{Creation of the real-world query session}
\label{substep:2_1}

% 1. MDC equations (from DLIO)
% 2. Key info saver.
% 3. Scan Context and Indoor SC explanation

The correct generation of a query session $\mathcal{S_Q}$ from real-world data involves three primary substeps, elaborated upon as follows. 

%  (1/3) MCD: Motion distortion Correction: About issues with LiDAR distortion
\subsubsubsection{Motion distortion correction}
Point clouds acquired from mobile spinning \ac{LiDAR} sensors often experience motion distortion because the rotating laser array collects points in various instances during a sweep, leading to inaccuracies. 
Therefore, one of the main issues using \ac{LiDAR}-only algorithms is the difficulty in correcting motion-distorted \ac{LiDAR} scans in the presence of fast motion. 
 
In some \ac{SOTA} \ac{LiDAR}-only \ac{SLAM} algorithms, the authors have assumed constant velocity models to overcome this issue, as done in KISS-ICP \citep{vizzo2023kiss}. 
Although this assumption can hold for data acquired with \ac{LiDAR} placed over autonomous cars and simplistic motion patterns,  as in the KITTI raw dataset \citep{kitti_raw_Geiger2013IJRR}, the constant-velocity model cannot capture subtle movements and generally does not hold for data acquired with handheld devices or Unmanned Vehicles (UVs) in indoor or outdoor scenarios \citep{zheng2023traj}.

Therefore, we take advantage of the \ac{MDC} of one \ac{SOTA} \ac{LIO} system to generate undistorted scans before alignment with the reference map.

In particular, we leverage the \ac{MDC} implementation in \ac{DLIO} \citep{chen2023directDLIO},  which, inspired by \cite{forster2016manifold}, generates continuous-time trajectories. 
Their approach considers a motion model characterized by constant jerk and angular acceleration compensated with \ac{IMU} measurements. 
This enables fast and parallelizable point-wise motion correction.

Once the scans are deskewed with the information from the \ac{IMU}, keyframe scans can be extracted with timestamps and odometry calculated poses.
This process is explained in the subsequent section.

%  (2/3) Key info saver:
\subsubsubsection{Key information saver}
The goal here is to save equally spaced undistorted scans (i.e., after a specific variation of time, translation, or rotation) with respective odometry estimated poses from a sequence of data that was previously recorded in a ROS \textit{bagfile} during acquisition with a mobile mapping system device.

To extract keyframes and construct the real-world query session $\mathcal{S_Q}$, the methodology proposed by \cite{kim2022ScLidar} presents a viable approach. 
The authors implemented loop closure mechanisms and keyframe information-saving capabilities as an extension in several \ac{SOTA} algorithms.

In general, the approach can vary depending on the available data.
When dealing with \ac{LiDAR}-only data, SC-A-LOAM \citep{kim2022ScLidar}, an enhanced version of A-LOAM \citep{Zhang.07122014} is a valid technique; however, it assumes constant velocity for \ac{MDC}.
For an additional calibrated 9-axis \ac{IMU}, the corresponding enhanced version of LIO-SAM \citep{lio_sam_shan2020lio} can be used.

If we deal with 9-axis or only 6-axis \ac{IMU} measurements, which are typical for the internal \ac{IMU}s of LiDAR and camera sensors, our open source keyframe information saver\footnote{\label{keyInfoSaver}\url{https://github.com/MigVega/Key-Info-Saver-SLAM}} together with almost any \ac{LIO} pipeline can be used (e.g., FAST-LIO2 \citep{FAST-LIO2_Xu.2022}, FASTER-LIO \citep{fasterLio2022} or iG-LIO \citep{iglio2024}). 
Something essential to consider is that the \ac{LIO} pipeline should publish (i.e., make available) the ROS topic with the undistorted scan in the local coordinate system.
This last characteristic is not standard and depends on the used \ac{MDC} strategy. 

Since \ac{DLIO} showed the best \ac{MDC} results in our experiments, we implemented and made open-source the corresponding enhanced version that transforms the deskew scan to the correct local pose after undistortion\footnotemark[\value{footnote}].

After saving the keyframe scans along with odometry information (i.e., time-stamped approximate 6-DoF poses), the final step to generate the query session involves feature descriptor extraction to encode the geometric information of the scans. 
This process will facilitate efficient comparison with reference session descriptors later.

%  (3/3) Scan Context and Indoor SC explanation
\subsubsubsection{Indoor Scan Context descriptor}
For place recognition, we introduce the new \ac{ISCD}. This variant diverges from the original Scan Context descriptor by focusing exclusively on indoor scans, as opposed to outdoor scans typically encountered in autonomous car environments, for which SC was originally conceived.
With \ac{ISCD}, our objective expands beyond merely eliminating ceiling points, which are notably common in indoor scans, especially in acquisitions with significant variations in pitch and roll angles, as usually encountered in handheld systems.
Moreover, we aim to selectively filter permanent vertical building elements perpendicular to the XY-plane, characterized by visible vertical surfaces of considerable length.

Inspired by \cite{kim2018scan, kim2021scan}, and by the formal definitions in \citep{wang2020intensity,li2021ssc}, the creation of \ac{ISCD} is as follows:
Azimuthal and radial bins split the 3D scan from the top view following an equally spaced arrangement (for reference, see an example on the left side of Figure \ref{fig:s1_3}). 

In the Cartesian coordinate system, we defined a LiDAR scan with $n$ points as $\mathcal{P}=\left\{\mathbf{p}_1, \mathbf{p}_2, \cdots, \mathbf{p}_n\right\}$ with each point $\mathbf{p}_{\mathbf{k}}=\left[x_k, y_k, z_k\right]$. 
Each point $p_k$ can be converted into a polar coordinate system, as follows:
$$
\begin{aligned}
\mathbf{p}_{\mathbf{k}} & =\left[r_k, \theta_k, z_k\right], \\
\ r_k & =\sqrt{x_k^2+y_k^2}, \\
\theta_k & =\arctan \frac{y_k}{x_k}.
\end{aligned}
$$
The point cloud is then segmented into $N_s$ sectors and $N_r$ rings by equally diving polar coordinates in azimuthal and radial directions. Each block is represented by:
$$
\begin{aligned}
B_{i j}= & \left\{ \mathbf{p}_{\mathbf{k}} \in \mathcal{P} \mid \frac{(i-1) \cdot R_{\max }}{N_r} \leq r_k<\frac{i \cdot R_{\max }}{N_r},\right. \\
& \left.\frac{(j-1) \cdot 2 \pi}{N_s}-\pi \leq \theta_k<\frac{j \cdot 2 \pi}{N_s}-\pi\right\},
\end{aligned}
$$
where $i \in\left[1, N_s\right], j \in\left[1, N_r\right]$, and  $R_{\max }$ is the maximum radius considered to create the descriptor. 
In contrast with the original \ac{SCD}, instead of taking only the $z$ value of the highest point in the bin $b_{i j}$, in \ac{ISCD}, we only assign a value equal to 1 if there are a minimum of $ISC_{\text{min}}$ points in the bin, and 0 otherwise. 
Formally:
$$
b_{ij} = \begin{cases} 
      1 & \text{if } \text{count}(\mathbf{p}_{\mathbf{k}} \in B_{ij}) \geq ISC_{\text{min}} \\
      0 & \text{otherwise}
\end{cases}
$$

The final \ac{ISCD} $\Omega \in \mathbb{R}^{N_r \times N_s}$, can be generated by:
$$
\Omega(i, j)= b_{i j} .
$$
The global signature $\Omega$ is a $2 \mathrm{D}$ matrix that efficiently encodes the geometry of mainly permanent elements (e.g., walls and columns) visible from the position of the sensor. 

\begin{figure}[!htb]
    \centering
    \includegraphics[width=\textwidth]{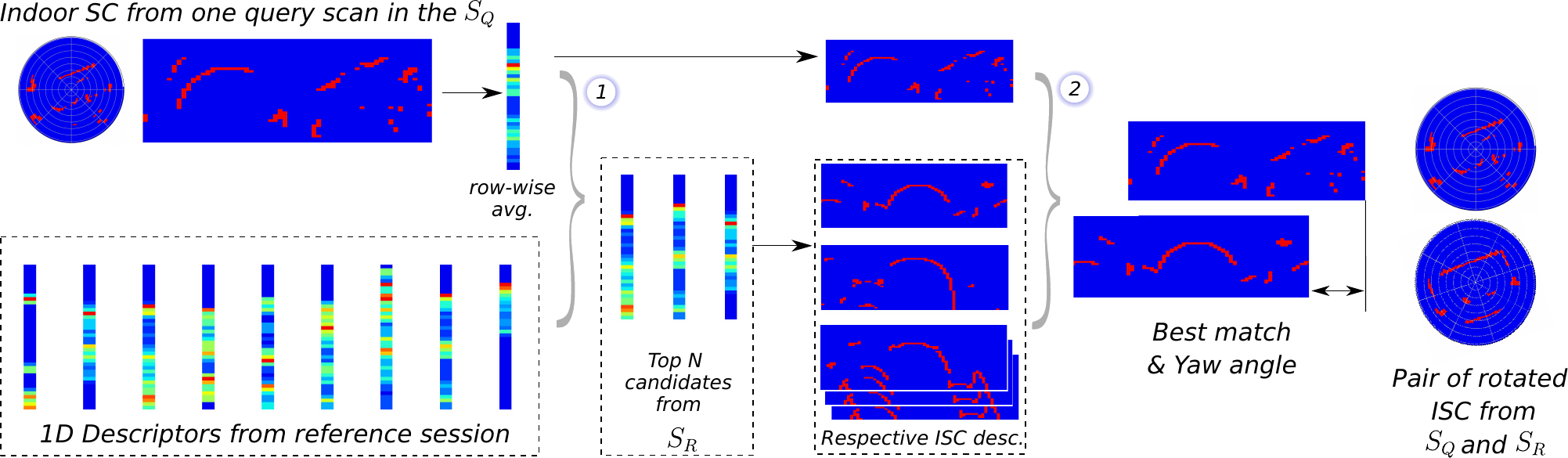}
    \vspace*{1mm} 
    \caption{Indoor Scan Context loop detection: The query session's scan is converted into 1D rotational invariant descriptors. These descriptors are quickly compared with those from the reference session to select the top $N_c$ candidates (see number 1). In the second phase, the 2D descriptors of these candidates are compared using cosine similarity while systematically varying the column position to identify the best match and optimal yaw angle alignment.}
    \label{fig:s2_2}
\end{figure}

Note that if $B_{i j} \in \varnothing$, $\Omega(i, j) = b_{i j}=0$, i.e., if in the bin there are no scan data because the bin is free or occluded, the bin will have a value of zero and will be visible as a blue color in the image representation of the descriptor (as shown in \ref{fig:s1_3} and \ref{fig:s2_2}).

In the following section, these descriptors are exploited to rapidly determine the rough alignment between the query and reference sessions.

%%%%%%%%%%%%%%%%%%%%%%%%%%%%%%%%
%% Step 2.2 Place recognition for inter-session loop detection
% About how SC loops are found + Yaw GICP
\subsubsection{Place recognition for inter-session loop detection} % or Place recognition between sessions
\label{substep:2_2}

% Content:
% (1/2). ISC Loop detection: Best match finding:
% - 1D Rotational invariant key comparison
% - 2D Column-wise cosine distance
% (2/2). YawGICP for verification -> parallel computation

%  (1/2). ISC Loop detection:
Having $\mathcal{S_Q}$ (real-world query session) and $\mathcal{S_{R}}$ (session from the reference map), we aim to align these two sessions. 
With this aim, we look for correspondences comparing the previously generated \ac{ISCD}s between the sessions to find inter-session loop closures.
This task is also known as \textit{place recognition}, in which one aims to identify or determine the specific location or place of sensor measurements (in our case, single LiDAR scans) within a given map.
% inter- means "between" (as in happening between two things)}

In order to facilitate quick comparison, the 2D descriptor is condensed into a one-dimensional vector. This vector is generated by calculating the average of the rows in the 2D descriptor. This average ensures rotation invariance, meaning that if a scan is in a location that is approximately the same but with a different yaw angle, the resulting 1D descriptor will remain unchanged.

The comparison between the query scan (from $\mathcal{S_Q}$) and the scans from the  $\mathcal{S_{R}}$ is facilitated by employing a \ac{KNN} search in a KD-Tree and using the L2-norm metric. 
 
Subsequently, the corresponding 2D descriptors of the $N_c$ closest 1D descriptors are compared using the \textit{column-wise cosine distance}.

This column-wise cosine distance is calculated to identify the similarity between two \ac{ISCD}s $\Omega^q$ and $\Omega^r$.
Let $\mathbf{v}_i^q$ and $\mathbf{v}_i^r$ be the $i^{t h}$ column of $\Omega^q$ and $\Omega^r$; the score can be found by:
$$
\varphi_i\left(\Omega^q, \Omega^r\right)=\frac{1}{N_s} \sum_{i=0}^{N_s-1}\left(\frac{\mathbf{v}_i^q \cdot \mathbf{v}_i^r}{\left\|\mathbf{v}_i^q\right\| \cdot\left\|\mathbf{v}_i^r\right\|}\right) .
$$
A comparison conducted column by column is beneficial for handling dynamic entities or slight differences between the reference map and the query session (e.g., new furniture or clutter) since although some columns of the 2D descriptor may show variations, the remaining columns will exhibit similarities. 
However, relying solely on this comparison overlooks the possibility of revisiting the exact location from a different perspective. 
To tackle this limitation and ensure rotational invariance in the matching process, the method computes distances using a range of column-shifted scan contexts. 
Then, it identifies the shift that yields the minimum distance.
This procedure resembles the coarse alignment of two sets of points, focusing mainly on aligning the yaw angle.
By implementing this approach, the optimal number of column shifts (i.e., optimal yaw angle) for alignment and the corresponding minimum distance can be determined.

Formally, if we compare $\Omega_k^q$ and $\Omega^r$ where $\Omega_k^q$ is $\Omega^q$ shifted by $k^{t h}$ column. 
The final score is calculated as follows:

$$
\Phi_i\left(\Omega^q, \Omega^r\right) = \underset{k}{\operatorname{argmin}}\ \varphi_i\left(\Omega_k^q, \Omega^r\right).
$$
The matched pairs are subsequently refined through a filtering process employing an empirical threshold, denoted as $\epsilon$, applied to the calculated minimum distance metric, $\Phi_i$.

%  (2/2). YawGICP
After detection of \ac{ISC} loop closures, a 6D relative constraint is established between two keyframes if there is a successful alignment between a sub-map from the reference session, denoted as $\mathcal{P}_{R, i}$ (which comprises the three closest scans to the one that matched the scan in the query session), and the single undistorted scan from the query session, denoted as $\mathcal{P}_{Q, j}$. 

The correctness of the alignment between these two keyframes is essential for the subsequent steps in the pipeline, as it dictates the effectiveness of the initial global registration between sessions.

To achieve this alignment in a robust way, we introduce YawGICP, an improved variant of the \ac{GICP} algorithm. 
YawGICP primarily focuses on translational changes and yaw angle adjustments, thereby mitigating significant pitch or roll rotations commonly induced by conventional GICP alignment procedures. 
This precaution prevents instances where standard GICP may accidentally rotate the source point cloud by 90 degrees (in pitch or roll), leading to erroneous associations between wall, ceiling, or floor points. 

The YawGICP is initialized with the yaw angle calculated in the previous step.

%% TODO: provide algorithm of  YawGICP (see KISS-ICP and IG-ICP papers).

\begin{figure}[!htb]
    \centering
     \includegraphics[width=0.5\textwidth]{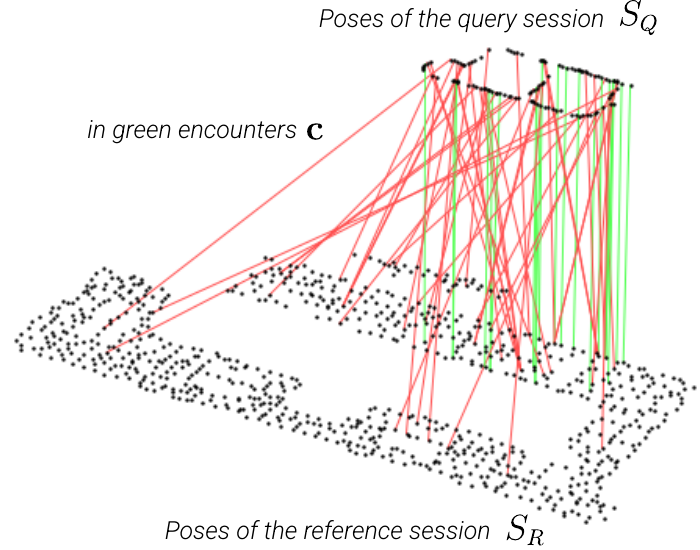}
    \caption{
    Detection of Indoor Scan Context loop closure between sessions. On the top are the poses from the query session, and on the bottom are the poses from the reference session, in this case, created from a BIM model.
    Correct correspondences are represented by green lines, while erroneous ones as red. After the YawGICP step, the erroneous correspondences are effectively discarded.
    }
    \label{fig:LC}
\end{figure}

Consistent with prior work \citep{vega:2023:BIM_SLAM,kim2022ltMapper}, only \ac{ISC} loops exhibiting a satisfactory fitness score, indicating a high percentage of inliers, are considered.
These loops are then incorporated into the factor graph problem with low covariance $\Sigma_c$, serving as factors between sessions with anchoring. 
Further elaboration on the factor graph problem will be provided in the subsequent section (\ref{substep:2_3}).  
Figure \ref{fig:LC} illustrates the detected ISC loop closures, which are then classified into correct and incorrect using YawGICP.

% \nointerlineskip 

%%%%%%%%%%%%%%%%%%%%%%%%%%%%%%%%
%% Step 2.3 Pose graph optimization and data alignment
% Factor graph problem + KNN Loops + Final ICP
\subsubsection{Pose graph optimization and data alignment}
\label{substep:2_3}

% Content:
% (1/3). Factor graph initialization with Odom. and ISC loop constraints
% (2/3). KNN Loops With adaptive covariance
% (3/3). Final ICP 

%----------------------------------
% (1/3). Factor graph initialization with Odom. and ISC loop constraints
%-------------------------------------------
% Intro: requirements &  multi-session anchoring
%-------------------------------------------

In this substep, the initial odometry constraints derived from the preserved session data (referenced in substep \ref{substep:2_1}) and previously identified inter-session \ac{ISC} loop closures (introduced in substep \ref{substep:2_2}) are leveraged to achieve the data alignment. 

The objective is to first roughly align the entire query session with the reference session from the reference map. 
Consequently, even if some scans within the query session's keyframes do not have any correspondence with the reference session, they are still aligned to the most cohesive pose based on the identified correspondences (SC loops) with adjacent scans and the provided odometry constraints.

Formally, in this contribution, the alignment between the sessions is done using \textit{multi-session anchoring}.
This method was originally introduced by \cite{kim2010multiple} and was further developed by \cite{McDonald2013_1144}, \cite{ozog2016long}, \cite{kim2022ltMapper}.
One of the main motivations behind these projects is to solve the so-called \textit{multi-robot mapping} problem. 
In this context, and as explained in Section \ref{chap:theory}, maps generated by different robots commonly have distinct reference coordinate systems, which require the merging of these maps to form a globally consistent map with a unified global coordinate system.

%-------------------------------------------------
% Problem definition ->  BIM-SLAM
%-------------------------------------------------

We formally define our problem as follows: Given two sessions, $\mathcal{S_Q}$ and $\mathcal{S_{R}}$, each provided with odometry constraints, and in the case of $\mathcal{S_Q}$, potentially equipped with intra-session loop closure constraints identified by a SLAM algorithm with a key information saver (as explained in Section \ref{substep:2_1}), our objective is to determine the optimal poses for the nodes in $\mathcal{S_Q}$. 
These poses should effectively align the measurements within $\mathcal{S_Q}$ with those of $\mathcal{S_{R}}$, considering the existence of inter-session loop closure constraints between the two sessions. 

%\textbf{the prefix intra- means "within" (as in happening within a single thing), while the prefix inter- means "between" (as in happening between two things)}.

As explained in \ref{chap:theory}, multi-session anchoring can be formulated as a factor graph \ac{MAP} optimization problem. 

%-----------------------------------------------------
% Encounter in the MAP formulation
%-----------------------------------------------------
To properly consider the encounter measurements ($\mathbf{c}$) in the \ac{MAP} formulation in Eq.~(\ref{eq:least_sqrt}), we need to redefine the relative measurement model $h{\big(} .{\big)}$  in the global frame with the help of the anchor nodes. 

This adjustment is needed, considering that the encounter is a global assessment between two trajectories. 
However, the pose variables for each trajectory are defined in the session's local coordinate frame.
With the anchor nodes, the poses of the respective sessions are transformed into a global frame, where a comparison with the measurement becomes possible. 

The measurement model $h{\big(} .{\big)}$ is modified to  $h^{\prime}{\big(} .{\big)}$, to incorporate the anchor nodes, and therefore, the respective term in Eq.~(\ref{eq:least_sqrt})  is changed to:
$$
\sum_{j \in N_e }\left\|h_j^{\prime}\left(\mathbf{x}_{R, j}, \mathbf{x}_{Q,j}, \Delta_{Q}, \Delta_{R}\right)-\mathbf{c}_j\right\|_{\Sigma_c}^2
$$
The difference $\mathbf{c}$  in the global frame between a pose $\mathbf{x}_{R}$ and a pose  $\mathbf{x}_{Q}$ is estimated by $\mathbf{c}=\left(\Delta_{R} \oplus \mathbf{x}_{R}\right) \ominus\left(\Delta_Q \oplus \mathbf{x}_{Q}\right)$, where $\oplus$ and $\ominus$ are the $\mathrm{SE}(3)$  pose composition operators \citep{smith1990estimating,jose_tutorial2021}. 

The operation \(  \Delta_{Q} \oplus \mathbf{x}_{Q}\) represents concatenating the transformation of $\mathbf{x}_{Q}$ (the second pose) to the reference system already transformed by the anchor node $\Delta_{Q}$. In $\mathrm{SE}(3)$, the operator $\oplus$ is equivalent to matrix multiplication \citep{jose_tutorial2021}. 

Hence, the subsequent \textit{factor between sessions with anchoring} will properly integrate the encounters in the pose graph optimization. It achieves this by initially transforming the poses of each session into the global frame using the anchor nodes.
\begin{equation}
\begin{aligned} 
& \phi\left(\mathbf{x}_{R, i}, \mathbf{x}_{Q, j}, \Delta_{R}, \Delta_Q\right)\\
& \propto \exp \left(-\frac{1}{2}\left\|\left(\left(\Delta_{R} \oplus \mathbf{x}_{R, i}\right) \ominus\left(\Delta_Q \oplus \mathbf{x}_{Q, j}\right)\right)-\mathbf{c}\right\|_{\Sigma_c}^2\right) \label{eq:anchore_factor}
\end{aligned}
\end{equation}

%-------------------------------------------------
% Factor graph initialization
%-------------------------------------------------

While initializing the factor graph, the odometry constraints from both sessions and the constraints after ISC loop detection are added to the optimization problem, the first as \textit{between factors} and the latter as \textit{factors between sessions with anchoring}.

Considering that in our scenario, our objective is to use the coordinate system of $\mathcal{S_{R}}$ as the global system for alignment, the anchor node $\Delta_{R}$ of the reference session should be assigned an insignificantly small covariance ($\Sigma_P$).
Conversely, for the anchor node $\Delta_Q$ of the query session, a significant covariance is assigned ($\Sigma_L$).

Moreover, the odometry poses are also added to the factor graph. 
However, since $\mathcal{S_{R}}$ comes from the reference map, its poses $\mathbf{x}_{R}$ are treated as fixed and should not be altered by the optimization.
To avoid changes to these poses, they are added to the factor graph optimization problem as \textit{prior factors} with very low covariance ($\Sigma_P$) in its noise model.

Following batch optimization, the intermediate optimized values of the anchor node $\Delta_Q^*$ and the poses $\textbf{x}_Q^*$ are obtained. However, these poses are expressed in the local coordinate system of $\mathcal{S_{Q}}$. To convert them from this local coordinate system (denoted as ${ }^Q \mathcal{G}_Q^*$) to the global coordinate system ${ }^W \mathcal{G}_Q^*$ of the reference map, we apply the following transformation to each pose $\mathbf{x}$ in the graph:
\[ { }^W \mathbf{x}_Q^* = \Delta_Q^* \oplus  { }^Q \mathbf{x}_Q^* , \]
where $W$ is the global coordinate system, or in our case, the coordinate system of the reference session.

%%%%%%%%%%%%%%%%%%%%%%%%%
% (2/3). KNN Loops With adaptive covariance
After the previous step, the query session roughly aligns with the reference session. 
To further refine the poses of the query session, we introduce a rapid \ac{KNN} loop detection method with adaptive covariance. 
Initially, submaps are generated by selecting \ac{KNN} scans from the scan to be aligned within the query session, along with the k-nearest scans from the reference session. 
Subsequently, the YawGICP algorithm (see \ref{substep:2_2}) is employed to register these two submaps, and the quality of registration is assessed based on a predefined fitness threshold, classifying the alignment as either good, acceptable, or unacceptable. 

Upon acceptance of the alignment, the constraints are added to the optimization problem as factors between sessions with anchoring with adaptive covariance. 
This adaptive covariance strategy assigns a very low covariance in the noise model to constraints originating from well-registered keyframe submaps, while constraints from just acceptable registrations receive a higher covariance. 
This approach allows the pose graph optimization to appropriately weigh the influence of these constraints in calculating optimized poses.

 % (3/3). Final ICP 
 
After conducting batch optimization one more time with incorporated odometry, ISC, and KNN constraints in the factor graph problem, the resulting poses undergo further refinement through a \textit{final ICP} registration.
Unlike previous steps that relied on registration with simulated scans from the reference map, this stage utilizes a one-centimeter-dense point cloud obtained from the reference map as the registration target. 
In case the reference map is a BIM model, this point cloud is created by sampling uniformly points over a mesh of permanent elements in the building (i.e. without doors and windows similarly as done in Step 1, section \ref{step:1_1})

Due to the high density of the target point cloud, \ac{GICP} fails to offer any significant advantage over \ac{P2P}-ICP \citep{p2p_ICP:1992}. 
In fact, in specific scenarios, \ac{GICP} yields inferior results.
Therefore, we have opted to use \ac{P2P}-ICP, which not only produces competitive results but also operates considerably faster.

To speed up computations and avoid the time-intensive KNN search associated with registrations involving a large target point cloud, scans within the query session are allocated into proximity-based groups. Subsequently, for each group, a target point cloud is created, dynamically cropping the reference map into spheres. 
The individual source scans within each group are then registered concurrently, leveraging parallel computing techniques.

The registration results are evaluated using three metrics. One metric is the \ac{RMSE}, and the other two correspond to fitness scores calculated at two distinct maximum \ac{P2P} distances: $F_1$ and $F_2$. 
The fitness score is the percentage of source inliers, considering a maximum \ac{P2P} distance threshold to classify points as inliers after registration. 

These metrics are computed explicitly for points located within 30 cm from the target point cloud after registration. This approach ensures the exclusion of points outside the reference map or those influenced by significant environmental changes, such as the addition of new walls or large pieces of furniture.

Depending on the metric values, the resulting aligned scans are categorized into four classes: Perfect, Good, Bad, and Outside the Map. The result is saved on a list, denoted as $\nu_Q$.

The resulting poses will be used in the subsequent step to create the final aligned map and compare it accordingly with the reference map.

% Step 3. Change detection
\subsection{Change detection and map update} % Consider renaming to "Data fusion" or Map merging
\label{step:3}

%% Content - Scan-vs-BIM 
% 1) Positive diffs.: Reconstruction of the final map and Positive difference detection (meshing -> not only voxels)   
% 2) Negative diffs. : Free space (with OctoMap) and Negative difference detection 
% 3) Map (PC/BIM) Update -> Voids and new elements (obj2ifc) % for future paper

Following the completion of the prior steps, the two sessions have been precisely aligned, and they now share a unified coordinate system. 
Subsequently, a comprehensive 3D map of the most up-to-date environmental state can be generated by placing the keyframes $\mathcal{P}_{Q, i}$ from the query session $\mathcal{S_{Q}}$ in the estimated poses ${ }^W \mathbf{x}_{Q, i}^*$, which are now in the global coordinate system.

If desired and to ensure the integrity and fidelity of the final map representation, it is recommended to exclusively incorporate scans classified as "perfectly" or "good" aligned within $\nu_Q$ during the map construction process. 

However, it is essential to note that although the remaining poses may not meet the strict alignment criteria with the reference map, they have already undergone significant optimization through odometry and loop closure constraints. 
Consequently, they can be utilized to generate the final map and even extend the reference map if the scan extends beyond its boundaries.

Since both maps are now aligned, a comparison of the two 3D maps becomes feasible.
The comparison process involves categorizing the elements in the map into three distinct types: Positive differences (PDs) denote instances where new objects have been introduced compared to the reference map; negative differences (NDs) signify the removal of objects previously documented in the reference map; and unaltered elements (UEs) denote features that remain constant across both maps.

This categorization is facilitated with the OctoMap library \citep{octomap}.
OctoMap, a widely-used library in robotics and 3D mapping, operates by dynamically updating voxel occupancy status within its octree structure as new point clouds are integrated. 
The analysis of measurement densities in OctoMap enables us to distinguish between occupied and free space, facilitating reliable 3D mapping. 

Additionally, we also leverage the probabilistic capabilities of OctoMap during measurement accumulation to facilitate the automatic removal of dynamic elements from the final point cloud. This removal is done based on occupancy patterns across multiple scans.
The resulting map is the one used to detect PDs and UEs in the preceding step. %could be improved with equations (as Olaf did).
Moreover, OctoMap calculates free space by identifying regions where the sensor fails to detect objects; this free space will be leveraged for NDs detection later.

% 1/2) Positive diffs.: Reconstruction of the final map and Positive difference detection (meshing -> not only voxels)   

To detect PDs and UEs, a P2P distance threshold is used between a point cloud from the reference map (also used in the previous final ICP step) and the newly created map with OctoMap, similar to what was presented in \citep{vega:2023:BIM_SLAM}.
A signed distance computation allows the distinction of points that are near and far from the reference map.
Near points allow for the confirmation of UEs, whereas distant points are regarded as PDs. 

The point cloud of identified PDs is passed through an outlier removal process. 
Subsequently, the point cloud undergoes a segmentation process through the density-based clustering technique (DBSCAN). This step is based on a neighbor-distance threshold and a minimum number of points per cluster.

Lastly, for each \ac{PD} cluster, a mesh is created using cubes from a \ac{VG} of the point cloud. 

% Compared with other surface reconstruction methods (like Poisson or Alfa), a voxelized representation of the new elements is considered more convenient since interpolation between points is avoided. 

Voxels, in contrast to other surface reconstruction approaches, capture the actual geometry of objects present in the scene. This leads to improved visualization of the new elements in conjunction with the reference map, providing a better understanding of the scene.

% 2/2) Negative diffs. : Free space (with OctoMap) and Negative difference detection 

The process of detecting NDs involves conducting a visibility analysis using individual scans from the query session ($\mathcal{P}_{Q, i}$). 
As mentioned before, the OctoMap library facilitates this analysis by calculating the free space, i.e. areas where the LiDAR did not detect any objects from its origin point. 
Similarly, as with the PDs, this free space is used together with a P2P distance threshold against a point cloud sampled from the reference map to identify the NDs.

The regions at the intersection between the reference map and the free space are the NDs. 
These are then passed through the outlier removal and clustering process, removing isolated points and small clusters.

\begin{figure}[!htb]
    \centering
	\subfloat[]{\label{fig:s3_a}{
	\includegraphics[height=2.5cm]{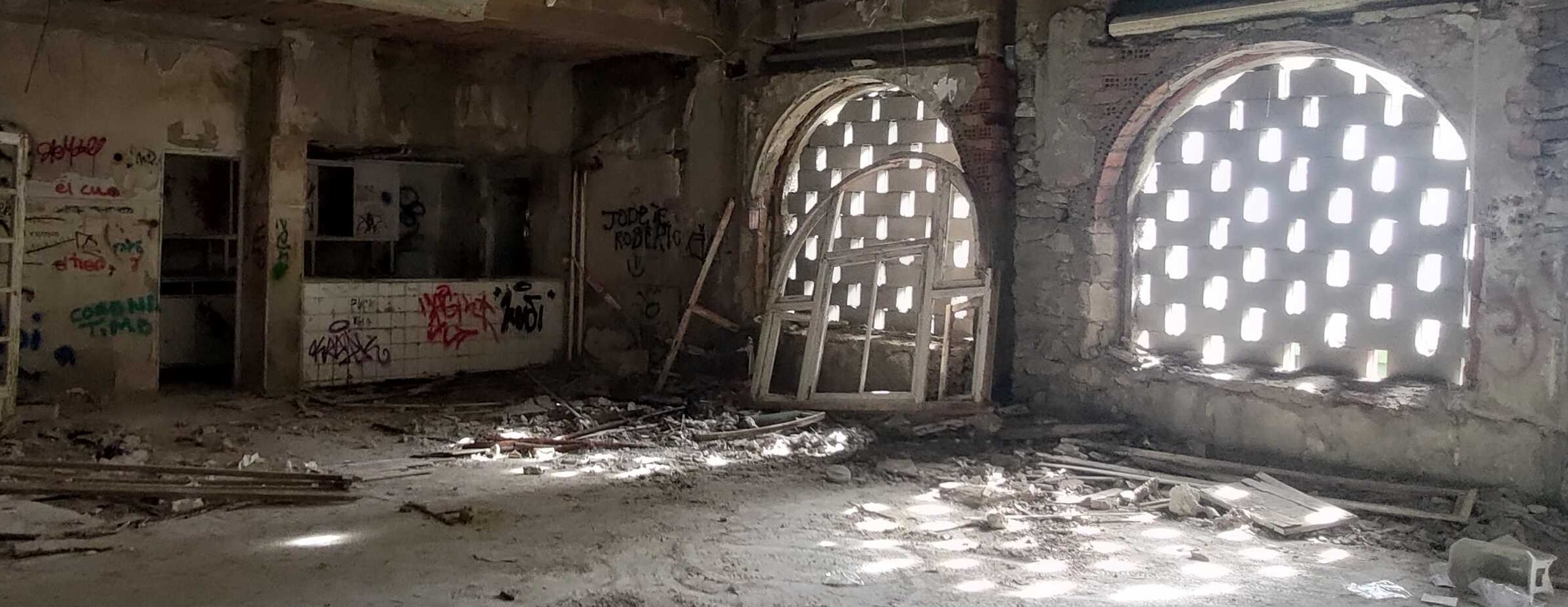}}}\hfill
 	\subfloat[]{\label{fig:s3_b}{
	\includegraphics[height=2.5cm]{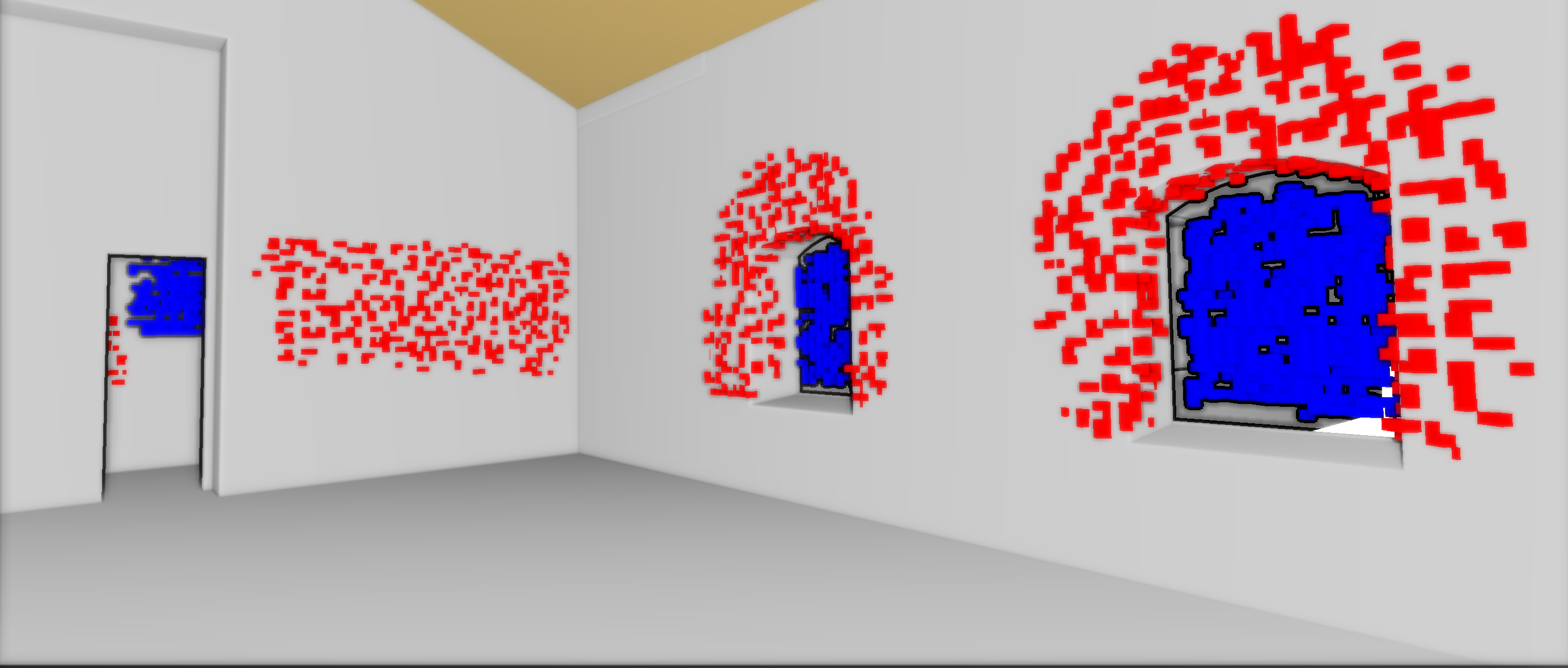}}}
    \caption{Positive and negative differences between the point cloud and the reference \ac{BIM model} are illustrated as follows: (a) A picture of the real-world scene. (b) Visualization of the detected changes in the form of voxelized clustered meshes with positive differences depicted in blue and negative differences in red. 
    Particularly, it is visible that the windows in the model are smaller compared to the real-world windows.}
    \label{fig:final_result}
\end{figure}

The final voxels are transformed into meshes and are colored blue for PDs and red for NDs.  An exemplary result is depicted in \ref{fig:final_result}.

\nointerlineskip

% 5: Experiments 
\section{Experiments}
\label{chap:experiments}

In this section, we present the data used to evaluate the efficacy of the proposed strategies. Comprehensive implementation details, such as the values of the essential parameters, are meticulously outlined to ensure a thorough understanding of our approach.

\subsection{ConSLAM dataset}

To ensure reproducibility and benchmarking, we evaluated our approach by applying it to the recently released open-access ConSLAM dataset \citep{trzeciak2023conslam,trzeciak2023conslamExtension}.

% brief description 
The ConSLAM dataset consists of four sequences of a construction site captured with a handheld system. It incorporates synchronized timestamped LiDAR scans, 9-axis \ac{IMU} measurements, and \ac{RGB} and \ac{NIR} camera images.

% about the BIM
Given the \ac{TLS} point cloud of sequence number two, we elaborate a half-centimeter-accurate \ac{BIM model}. 

We used the OA-LICalib library \citep{lv2020targetless,lv2022} to retrieve the extrinsic calibration parameters (rotation and translation) between the LiDAR and the \ac{IMU} sensors.

%%%%%%%%%%%%%%%%%%%%%%%%%%%%%%%%%%%%%%%%%%%%%%%%%%
\subsection{Implementation details}

While \textit{Step 1} and \textit{3} were implemented in Python, \textit{Step 2} was written in C++.

%% Step 1.3. LiDAR simulation

\subsubsection{\textbf{Step 1}: Reference session generation}

In \textit{Step 1}, to generate the reference session data ($\mathcal{S_R}$),  the vertical \ac{FoV} of the simulated LiDAR scans can be customized according to preferences. 
To achieve alignment with a TLS point cloud as a reference map, the simulated LiDAR scans encompass a range from -45 degrees to 45 degrees in the vertical \ac{FoV}. 
However, in our experiments, while aligning the data with a BIM model, we observed improved ISC loop detection when no ceiling points were present in the simulated scans. 
Consequently, the scans are adjusted to cover only from 0 to -25 degrees in the vertical direction.
In Blensor, during the LiDAR simulation process, the noise was set to a mean of zero with a standard deviation of 0.03 m, an angular resolution of 0.1728 degrees, and a maximum distance of 15 m.

%%% Step 2 

\subsubsection{Step 2: Query session generation, alignment, and correction}

%% Step 2.1.1.   MDC
\subsubsubsection{Step 2.1: Query session creation}
% \subsubsection{Motion Distortion Correction}
In \textit{Step 2}, to generate the query session from the real-word data ($\mathcal{S_Q}$), for the \ac{MDC} step, we opted for using \ac{DLIO}, because, in contrast to FAST-LIO2 \citep{FAST-LIO2_Xu.2022},  it does not require heavy downsampling of the point cloud for deskewing and registration. 
Hence, clean, undistorted scans with \ac{DLIO} allow dense map reconstruction.
As suggested by \citep{zhang2022multicamera}, we reproduced the data in the \textit{bagfiles} at a low rate (half of the original speed) to avoid errors during the distortion process.
%% Step 2.1.2.   key information saver
Regarding the key information saver, while it is possible to await a minimum variation on translation or rotation between consecutive scans, we opted to save scans given either a list of timestamps or after a specific interval of time has passed. 
This feature is convenient since we want to compare our results with existing ConSLAM GT poses. Therefore, we are mainly interested in specific frames with known timestamps.
%% Step 2.1.3. ISCD
For the creation of \ac{ISCD} we opted for $N_s=60$, $N_r=20$ (as suggested in \citep{kim2021scan}), $ISC_{\text{min}} = 40$, and a maximum radius of \SI{10}{\meter}.

%%%%%%%%%%%

\subsubsubsection{Step 2.2: Inter-session loop detection with ISC}
% Step 2.2.1. ISC Loops
Nanoflann \citep{blanco2014nanoflann} is used to create a KD-tree of 1D rotational invariant descriptors.
A total of 100 ($N_c$) top candidates were chosen to evaluate in 2D after the 1D descriptor comparison; it is worth mentioning that the retrieval of correct correspondences is very sensitive to this value. 
A cosine similarity threshold $\epsilon=0.3$ is used to filter out pairs of 2D descriptors that passed with the minimum distance among the possible column shifts $k$. Only column shifts of 10\% of the total number of columns (i.e., 36 deg) are considered for the alignment. 
% Step 2.2.2. YawGICP
All YawGICP registrations in the ISC and KNN loops are done using parallel computations with OpenMP.
Unlike conventional ICP implementations, when employing YawGICP, it is imperative to express the target point cloud (i.e., from the reference map) in the local coordinate system of the source scan (i.e., the point cloud to be aligned). 
Otherwise, the process will yield undesirable results. 
This shift is critical because the resulting transformation matrix is relative to the origin of the source scan, with the aim of rotating the point cloud from its local origin rather than the origin of the global coordinate system.

\subsubsubsection{Step 2.2: KNN loops, pose-graph optimization and final ICP}

% Step 2.3.1. KNN Loops
The \acl{KNN} used to create the submaps for KNN loop detection in the second step of optimization is 5.
To ensure correct alignment with the BIM model as the reference map, we opted to omit the KNN loop detection process. This decision was made because this process tends to induce erroneous correspondences.
% Step 2.3.2. PGO
Meanwhile, in Step 2.3.2 (Section \ref{substep:2_3}), the pose-graph optimization is done with GTSAM using iSAM2; the following are the values of the variances of the different noise models: $\Sigma_L = \pi^2$ (significant noise for query session's anchor node); $\Sigma_P= 1 \times 10^{-102}
$ (prior noise for reference map poses and initial poses); $\Sigma_O = 1 \times 10^{-4}$ (for odometry constraints); $\Sigma_c = 0.5$ (robust noise for encounters, i.e., loop closure constraints). 
% Step 2.3.3. Final ICP
The parallel creation of spheres for target point cloud registration and the P2P ICP of the single source scans is done using OpenMP in C++.
In our case, we use the following two maximum distances to calculate the fitness scores: $F_1 = 1$ cm and $F_2 = 3$ cm. 

%%%%%%%%%%%
\subsubsection{Step 3: Change detection and map update}

% Step 3. Change detection and map update
In \textit{Step 3}, the process is performed with Trimesh, OctoMap, and Open3D. 

We use a P2P distance threshold of 0.3 m to calculate the positive and negative differences.

OGM2PGBM \citep{vega:2022:2DLidarLocalization}, Scan Context \citep{kim2018scan}, and LT-SLAM \citep{kim2022ltMapper} were projects that we used as a reference and that are freely available online.

%%%%%%%%%%%
% Evaluation
For the evaluation of the results, presented in the following section, the trajectories were compared against the ground truth using evo \citep{grupp2017evo} in TUM format \citep{Sturm2012ABF} and using the Umeyama alignment \citep{umeyama}.

% 6: Results and analysis
\section{Results and analysis}
\label{chap:results}

In this section, we provide the results of our pipeline with respect to the alignment with an accurate TLS point cloud and with a BIM model as a reference map.

Table \ref{tab:GTresults} shows the \ac{APE} summary of the different methods in each sequence of the ConSLAM dataset after alignment with the corresponding TLS point clouds. 
In the table, the performance of DLIO (with Umeyama alignment) is compared against the results of our framework after improving the DLIO trajectory with Indoor Scan Context (ISC) loop detection and after KNN loop detection and optimization. 
The results after the \textit{final ICP} step correspond to the current ground truth (used to evaluate the methods); therefore, they are not numerical values for this step. 
Moreover, the results are compared against the original ConSLAM ground truth poses provided by the authors together with the dataset. 

\begin{figure}[!htb]
	\centering
	\subfloat[]{\label{fig:s4_translation}{
	\includegraphics[width=0.48\textwidth]{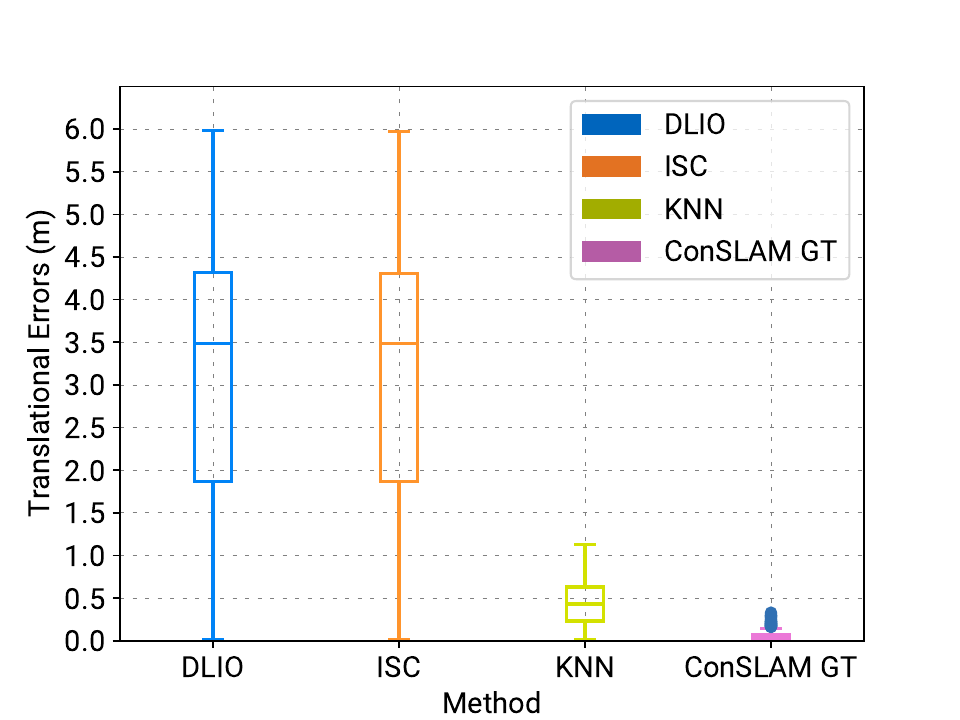}}}\hfill
	\subfloat[]{\label{fig:s4_rotation}{
	\includegraphics[width=0.48\textwidth]{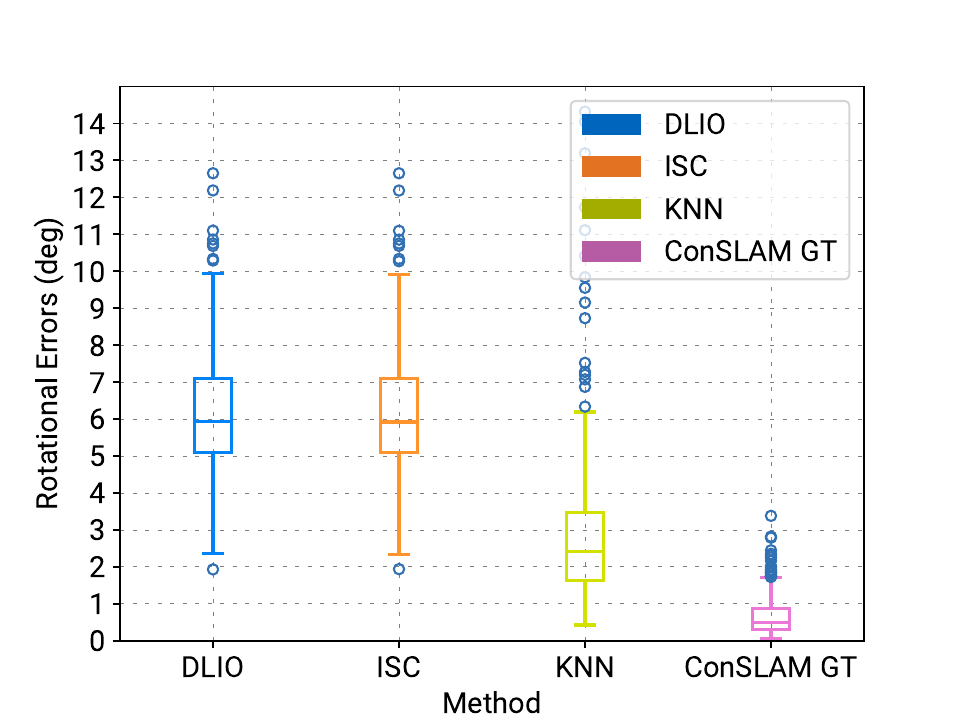}}}
	%\vspace*{3pt}
    \caption{Translational (a) and rotational (b) errors for sequence 4 after alignment with the respective TLS point cloud.} \label{fig:seq4_results}
\end{figure}

\begin{figure}[!htb]
    \centering
    \includegraphics[width=\textwidth]{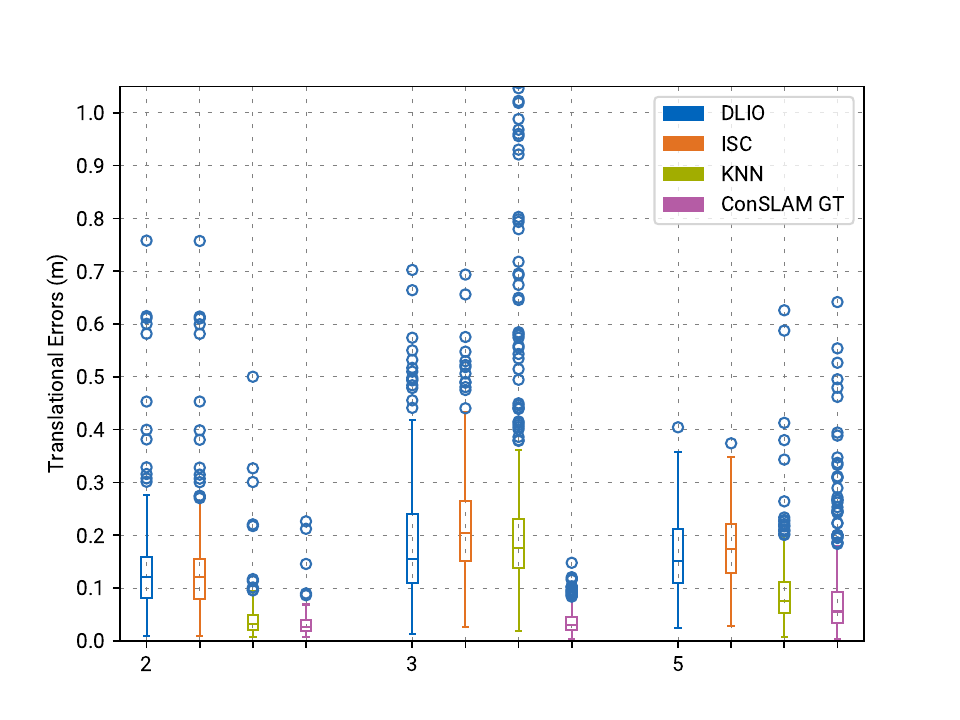}
    \caption{Translational errors for sequences 2, 3, and 5 after alignment with the respective TLS point clouds. }
    \label{fig:s235_translation}
\end{figure}

\begin{figure}[!htb]
    \centering
    \includegraphics[width=\textwidth]{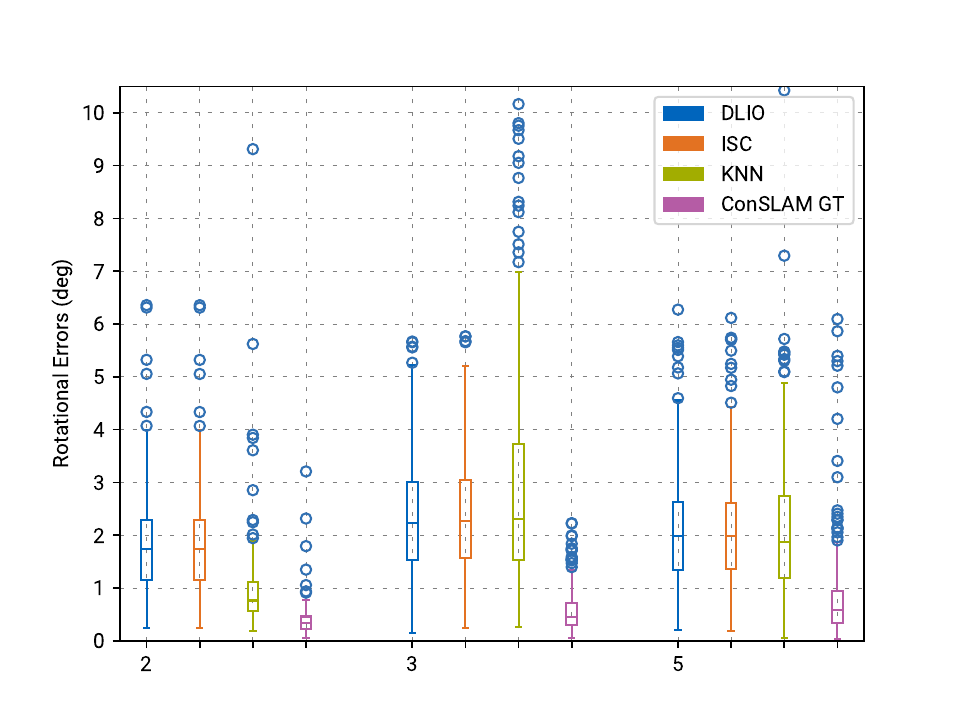}
    \caption{Rotational errors for sequences 2, 3, and 5 after alignment with the respective TLS point clouds.}
    \label{fig:s235_rotation}
\end{figure}

\begin{figure}[!htb]
    \centering
    \includegraphics[width=\textwidth]{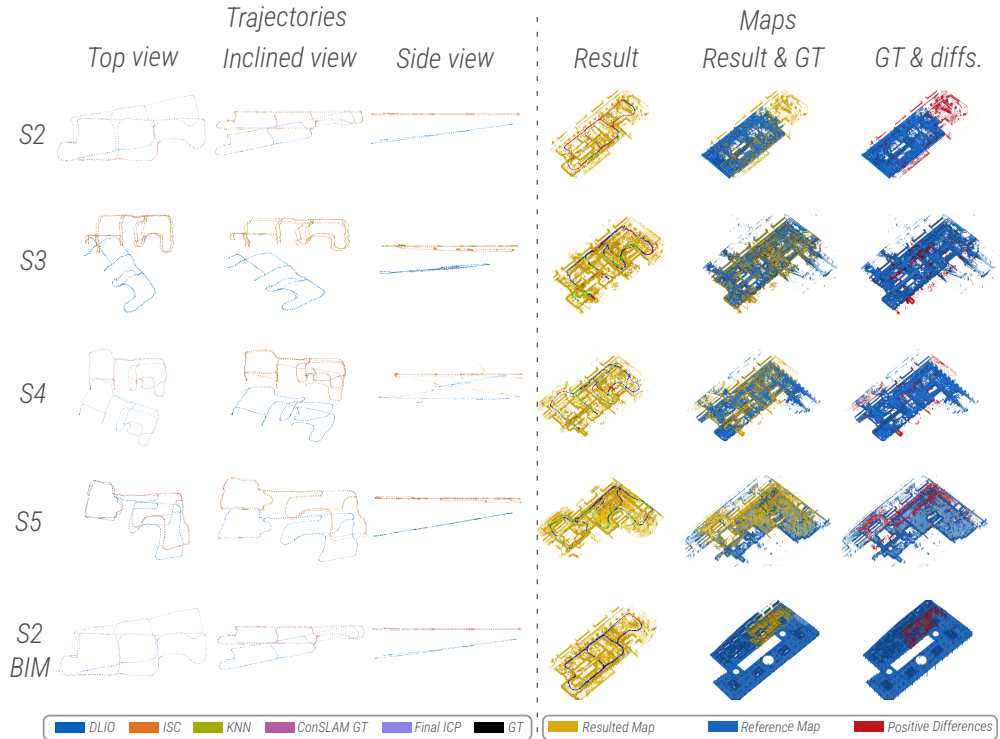}
    \caption{Trajectories and maps for sequences 2, 3, 4, and 5 after alignment with the respective TLS point clouds, and for sequence 2 after alignment with the BIM model. The trajectories of the first three columns correspond to the results of the different methods/steps, which have the same label colors as in Fig. \ref{fig:s235_translation}. Additionally, the ground truth trajectory is shown in black. The trajectory in the fourth column displays points in different colors to indicate registration results: perfect (green), good (blue), bad (red), or outside of the map (black). In the fourth and fifth columns, the resulting map is shown in yellow, and the reference target map is shown in blue. In the last column, the differences (new elements in the resulting map) are depicted in red.}
    \label{fig:results_visual}
\end{figure}

Furthermore, figures \ref{fig:seq4_results}, \ref{fig:s235_translation}, \ref{fig:s235_rotation} illustrate the distribution of errors (translational and rotational) for each method in the various sequences. Figure \ref{fig:results_visual}, provides a visual representation of the resulted trajectories and 3D maps.

\begin{table*}[!htb]
\begin{tabular}{@{}l|rr|rr|rr|rr|rr@{}}
\toprule
% Method     & \multicolumn{2}{c|}{Seq. 2} & \multicolumn{2}{c|}{Seq. 3} & \multicolumn{2}{c|}{Seq. 4} & \multicolumn{2}{c|}{Seq. 5} & \multicolumn{2}{c}{Average} \\ \midrule
Method     & \multicolumn{2}{c|}{S2 (225 m)} & \multicolumn{2}{c|}{S3 (340 m)} & \multicolumn{2}{c|}{S4 (275 m)} & \multicolumn{2}{c|}{S5 (320 m)} & \multicolumn{2}{c}{Average} \\ \midrule
DLIO       & 20.2          & 2.2         & 21.4          & 2.6         & 359.0         & 6.4         & 17.4          & 2.3         & 104.5            & 3.4           \\
SC         & 20.1          & 2.2         & 24.3          & 2.6         & 358.6         & 6.4         & 18.7          & 2.3         & 105.4            & 3.4           \\
KNN        & 9.0           & 1.4         & 34.6          & 4.3         & 53.4          & 3.5         & 11.2          & 2.5         & 27.1             & 2.9           \\
ConSLAM & 5.2           & 0.7         & 4.2           & 0.7         & 9.3           & 0.9         & 12.1          & 1.1         & 7.7              & 0.8           \\ \bottomrule
\end{tabular}
\vspace*{1mm}

\caption{Quantitative comparative results for each ConSLAM sequence (S2, S3, S4 and S5). Translational and angular APE RMSE in centimeters and degrees, respectively. Additionally, the length of each sequence is given in meters. SC refers to the results of DLIO after Scan Context loop detection and optimization, similarly KNN refers to the results after KNN loops. ConSLAM refers to the ground truth poses provided together with the dataset.} \label{tab:GTresults}%
\end{table*}

\nointerlineskip
\vspace{5cm}

\nointerlineskip

\nointerlineskip

Notably, the errors exhibit an evident reduction across almost all sequences while the pipeline evolves.

%% (1/3) ISC loops
The ISC loops primarily allow the critical first rough alignment between the query and reference sessions. 
Since only a few ISC loops are retained due to rigorous threshold criteria, the outcomes after ISC loop detection and optimization exhibit minimal alteration in trajectory accuracy when compared to the initial results derived from DLIO.

%% (2/3) KNN loops
On the other hand, the subsequent KNN loops exhibit a more pronounced impact on the results after ISC loops. 

While the average rotational error, as depicted in Figure \ref{fig:s235_rotation}, experiences a significant decrease in sequences 2 and 4, it exhibits apparent stability or even an increase in sequences 3 and 5.

%% (3/3) ConSLAM GT
Regarding the \ac{GT} poses provided with the ConSLAM dataset, although the \ac{RMSE} for \ac{APE} remains below 8 cm and 1 degree for translation and angular errors, respectively (as shown in the last column of Table \ref{tab:GTresults}), the maximum errors escalate to 20 or even 60 cm in sequences 2 and 5 (see Figure \ref{fig:s235_translation}). 
While these significant discrepancies are in relatively small sections of the trajectories, it is also essential to recognize that for a LiDAR-based SLAM dataset, ground truth poses should ideally exhibit accuracy levels of at least one centimeter across the entire trajectory. 
This level of accuracy is now achievable in a highly automated manner with the proposed SLAM2REF framework.

%\input{content/Tables/04_ComputationalTime}

\begin{figure}[!htb]
	\centering
	\subfloat[]{\label{fig:s2_BIM_translation}{
	\includegraphics[width=0.48\textwidth]{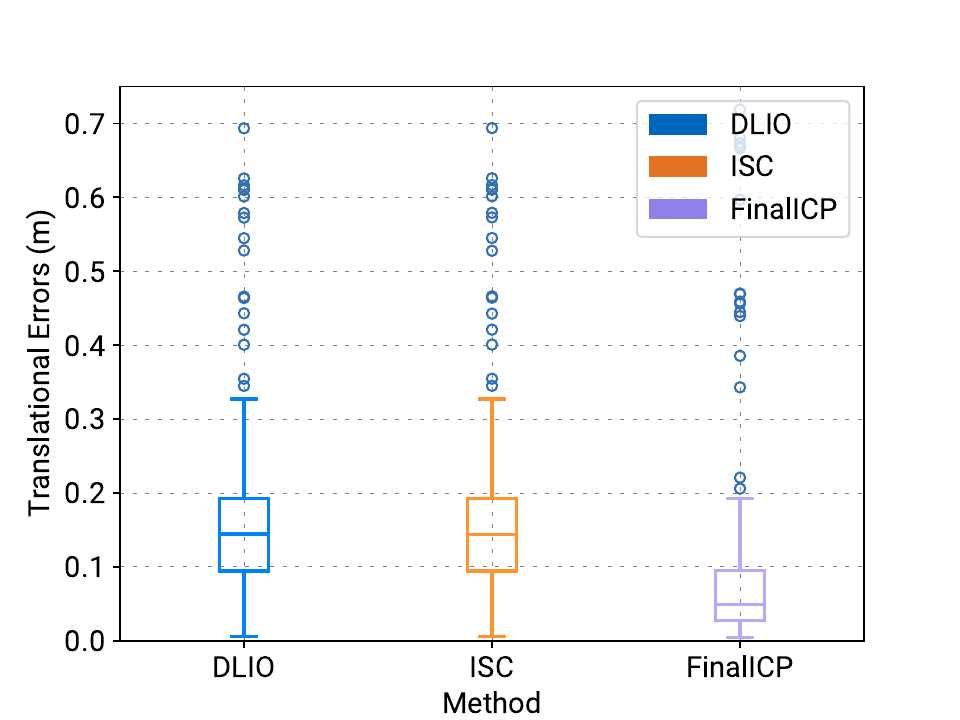}}}\hfill
	\subfloat[]{\label{fig:s2_BIM_rotation}{
	\includegraphics[width=0.48\textwidth]{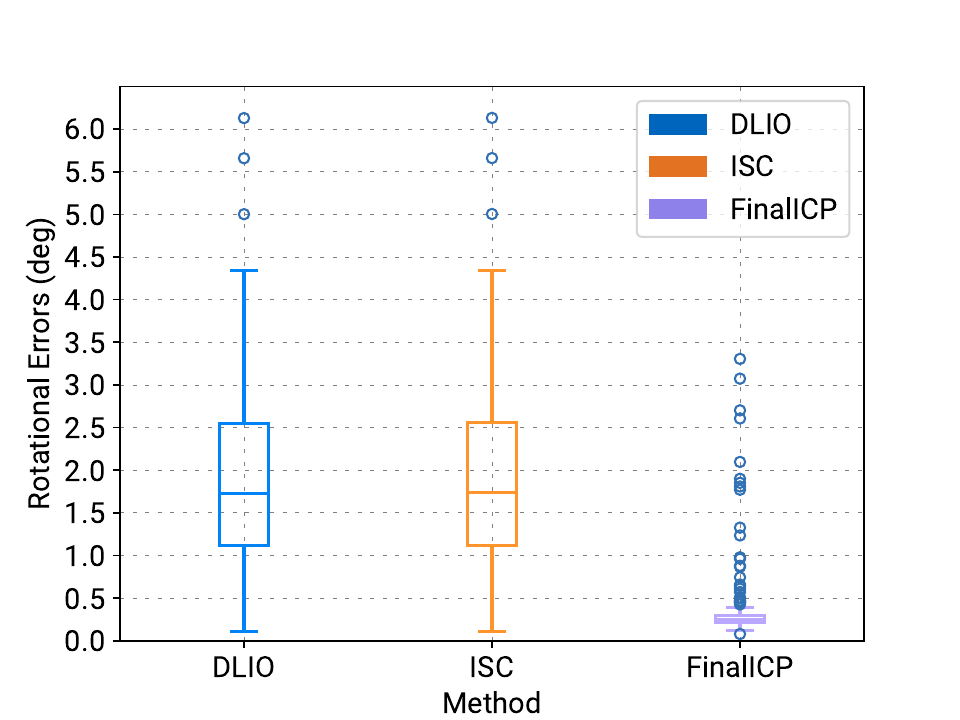}}}
	%\vspace*{3pt}
    \caption{Translational (a) and rotational (b) errors for the sequence 2 after alignment with BIM model.} \label{fig:seq2_BIM_results}
\end{figure}

%%%%%%%%%%%%%%%%%%%%%%%%%%%%%%%%%%%%
% Alignment with BIM
Additionally, we demonstrate that it is possible to align and correct a 3D map using a BIM model as a reference map,  despite significant deviations between the current map and the reference BIM model  (Scan-Map deviations of the types 1 and 2 as stated in the introduction, see Section \ref{sec:Introduction}). 
This significant level of deviation is particularly evident in the context of the ConSLAM construction site. % TODO Maybe add an image with the differences -> could be done with a TLS PC. 
Figure \ref{fig:seq2_BIM_results} and \ref{fig:results_visual} depicts the results after alignment with BIM. Here, the error values after the final ICP step are visible since they do not coincide with the ground truth poses anymore.

Similar to the alignment process with the TLS point cloud presented previously, the error does not decrease after ISC loops; however, it notably reduces after the final ICP step. 
The translational RMSE of the APE decreases to 14.8 cm, while the rotational RMSE is 0.56 degrees.

\nointerlineskip
\vspace{5cm}

\nointerlineskip

\begin{figure}[!htb]
	\centering
	\subfloat[]{\label{fig:s2_TLS_diff}{
	\includegraphics[width=0.48\textwidth]{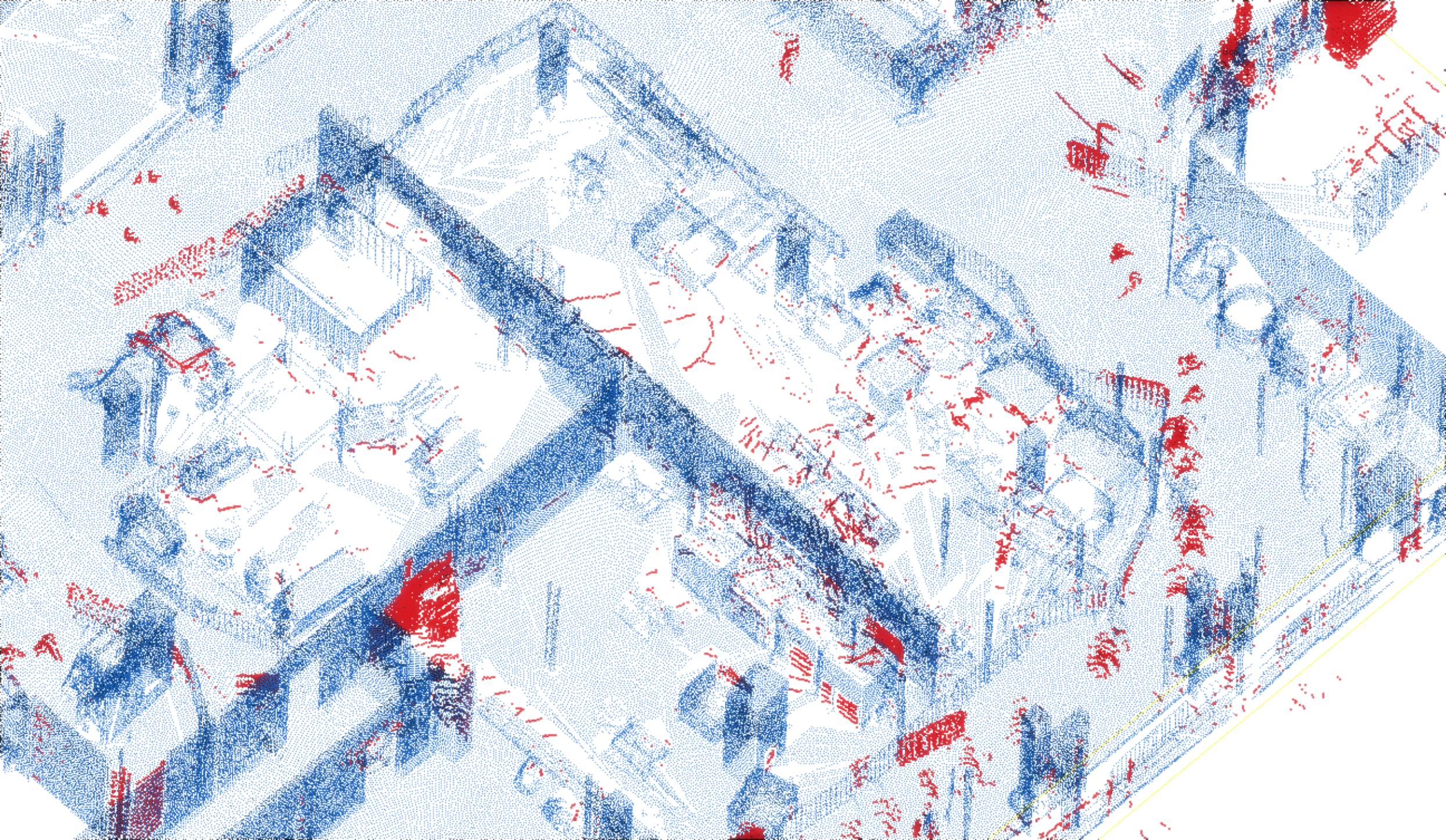}}}\hfill
	\subfloat[]{\label{fig:s2_BIM_diff}{
	\includegraphics[width=0.48\textwidth]{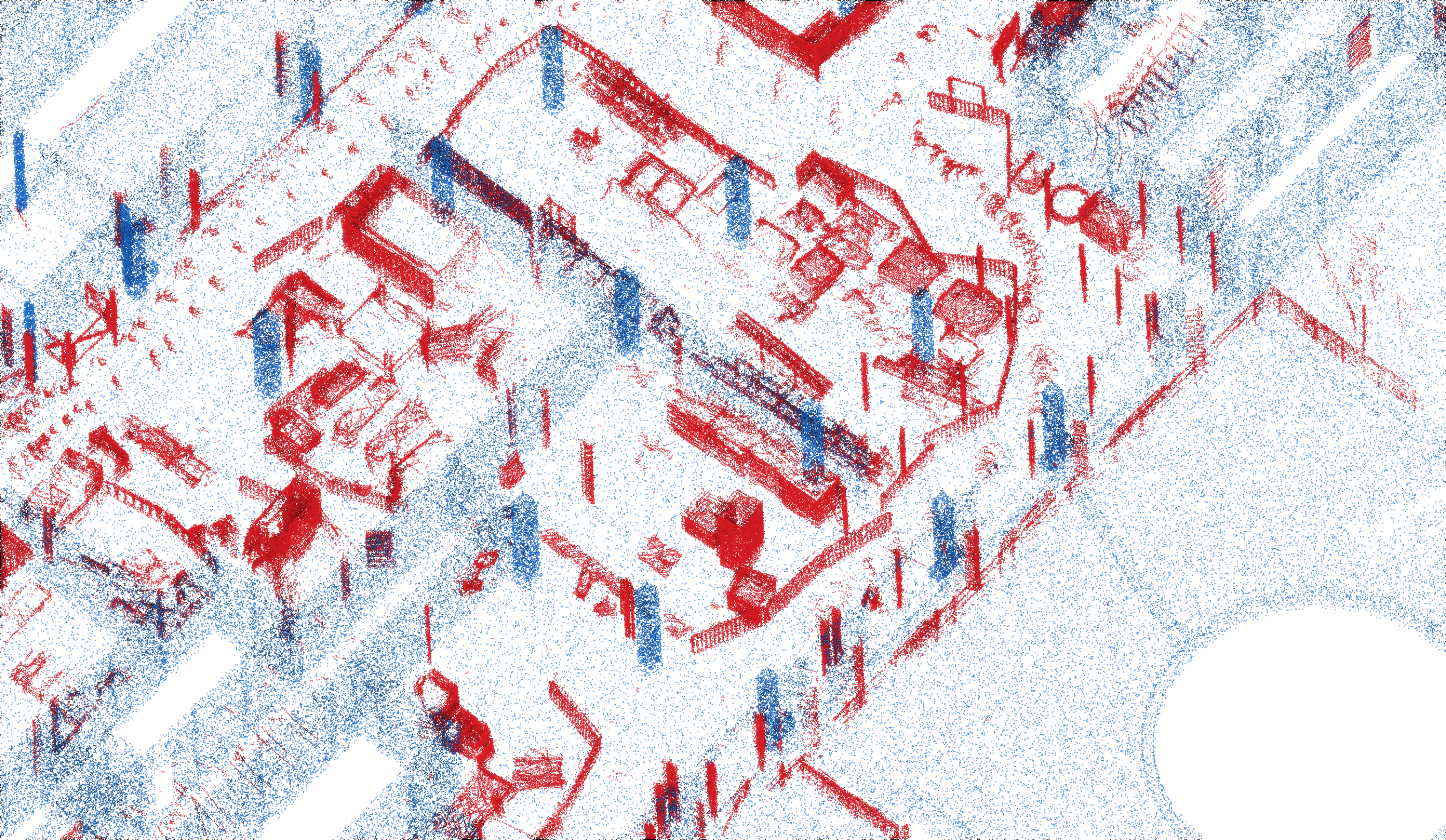}}}
    \caption{Change detection after map alignment. Both images correspond to the results of sequence 2: in (a), the sequence was compared against the respective TLS point cloud, and in (b) against the BIM model. The positive differences, i.e., new elements in the resulting map, are depicted in red.}
\end{figure}

Regarding the results of \textit{Step 3}, the identified positive changes are highlighted in red within the final column of Figure \ref{fig:results_visual}. Furthermore, Figures \ref{fig:s2_TLS_diff} and \ref{fig:s2_BIM_diff} provide detailed visualizations of the discrepancies observed in sequence 2 following alignment with both the TLS point cloud and the BIM model, respectively. While the disparities with the TLS point cloud are relatively minor, involving slight shifts in the positions of certain fences and construction resources, the distinctions when compared to the BIM model (Figure \ref{fig:s2_BIM_diff}) are notably substantial. This serves to exhibit the robustness of our alignment methodology in effectively accommodating considerable levels of Scan-BIM deviations.

\nointerlineskip

% 7: Discussion
\section{Discussion}
\label{chap:disccusion}
% What do the results mean?" and "How do they contribute to our understanding of the topic
This section contains a more detailed interpretation of the results reported previously. 
Furthermore, we look into the motivation for our methodology and how it contributes to progress in this field of research, explaining the enhancements of our approach compared to prior works and outlining directions for future studies.

%%%%%%%%%%%%%%%%%%%%%%%%%%%%%%%%%%%%%%%%%%%%%
% (1/4). Interpretation of Results
The apparently contradictory pattern of the rotational errors in sequences 2 and 5 can be attributed to the Umeyama alignment process \citep{umeyama}.
In certain regions, the actual trajectory after ISC loop detection (without Umeyama alignment) deviates approximately 1.5 meters from the ground truth in the Z and X directions, leading to erroneous identification of KNN loops. 
Nonetheless, these erroneous loops are effectively identified and filtered out during the final ICP step.

%% ConSLAM GT possible mistakes
One potential source of error for the ConSLAM GT poses lies in the \ac{MDC} step.
Contrary to common practice, the authors extracted the scans directly from the recorded \textit{bagfiles}, omitting the deskewing process \cite{trzeciak2023conslamExtension}. 
Avoiding the undistortion process can mislead any registration method, particularly affecting the accuracy of the calculated poses in sections where the trajectory was recorded under rapid motion.

%% Why not KNN for BIM % todo add ref to plot
The reason why KNN loops tend to yield incorrect correspondences during alignment with the SD from a BIM model can be attributed to Scan-Map deviations, as well as the absence of ceiling points in simulated scans from the BIM model. 
These facts complicate the registration of small sub-maps from the real world with sub-maps from a BIM model, particularly given that elements on the construction site have corners and features sometimes misinterpreted by the YawGICP registration process as permanent elements. 
Nonetheless, the final ICP method overcomes this challenge by utilizing a dense point cloud from the BIM model and relying solely on point-to-point (P2P) correspondences, thus avoiding estimating tangent planes for the alignment.

%% Step 1.3:  Why simulated scans instead of cropped spheres?
An alternative to simulating LiDAR scans (as done in section \ref{step:1_3}) could involve cropping a point cloud from the reference map within spheres as performed for the final ICP step (section \ref{substep:2_3}). 
However, simulating scans offers a critical advantage: it enables the incorporation of only the geometry of elements visible from the scan's origin.
This visibility filter is crucial for ensuring the robustness of descriptor-based alignment in the ISC loop detection step, as only the information of single scans is compared here.
Furthermore, when registering real-world scans with simulated ones, the process not only demonstrates quickness but also mitigates potential interference from double surfaces, such as from walls, as only the visible surfaces from the sensor origin are considered.

% 2/4. Comparison with Prior Work
In comparison to our prior research \citep{vega:2022:2DLidarLocalization} and other localization algorithms, SLAM2REF presents notable advantages. 
Since it enables the creation of a map and subsequent alignment with a reference map, unlike typical localization methods, our framework does not require the sensor to initiate mapping from within the map itself. 
Instead, it allows the sensor to start from any location, ensuring that the resulting map aligns with some overlapped regions of the reference map. 
Therefore, SLAM2REF also supports the extension of the reference map. 
This means that even if sensor measurements expand beyond the map boundaries, they are still aligned with the existing map in the most coherent manner.

Additionally, our pipeline does not necessitate any manual intervention to align the first keyframe, a process typically required by methods utilized to generate the ground truth poses in some of the latest datasets, such as in \citep{trzeciak2023conslam,trzeciak2023conslamExtension, Ramezani_2020, zhang2022multicamera}.
Moreover, due to the initialization of our pipeline with SLAM or odometry-calculated poses and the optimized parallel registrations, our pipeline also enables the rapid retrieval of GT poses utilizing dense, accurate reference maps.

When contrasting with BIM-SLAM \citep{vega:2023:BIM_SLAM}, SLAM2REF showcases several distinct advantages: Firstly, it is compatible with large-scale reference maps, encompassing not only large BIM models but also dense high-quality point clouds. 
Secondly, it effectively considers motion distortion in LiDAR data and mitigates it by leveraging IMU measurements. 
Thirdly, it achieves significantly improved accuracy in 6-DoF pose retrieval through the final ICP step and a TLS point cloud as a reference map. 
Lastly, our pipeline enables the alignment in the presence of Scan-Map deviation, such as with a BIM model, leveraging the proposed enhanced version of the Scan Context descriptor tailored for indoor environments.

Additionally, our pipeline operates independently of \ac{ROS} or Gazebo (used previously for scan simulation).

Another remarkable characteristic of our method is its adaptability, as it is not restricted to Manhattan-world environments with enclosed rooms, as the method proposed by \cite{shaheer2023graphbased}.

% % % % % % % % % % % % % % % % % % %
% 4/4 Limitations and Future Directions:
\section{Achievements and limitations}
\label{chap:limitations}
As mentioned in the previous section, SLAM2REF enables the automatic alignment and correction of data with a reference map. 
This alignment can achieve high accuracy and map extension, even with Scan-Map deviations and in environments that do not follow the Manhattan-world assumption.
Although meticulous attention has been given to its design, the proposed framework inherently has limitations that influence its effectiveness in aligning the new data with a reference map. 
This section elaborates on specific areas where the proposed methodology may encounter challenges.

One of the most critical limitations is the sensitivity to the initial poses calculated with the SLAM or LIO system.
In scenarios where these poses exhibit significant displacement or drift from their correct positions, our pipeline may not be able to rectify them. 
One way to possibly overcome this issue might be to implement a second ISC loop detection with a radius-based search, following the first ISC loop closure detection.
The first ISC loop will provide an initial coarse alignment of the trajectory, probably with only a few correspondences. 
Subsequently, the second ISC loop closure process will serve to correct parts of the trajectory that manifest substantial drift. 
However, it is worth noting that this approach may introduce wrong correspondences, particularly in cases with symmetric environments.

\begin{figure}[!htb]
	\centering
	\subfloat[]{\label{fig:l1_a}{
	\includegraphics[height=2.5cm]{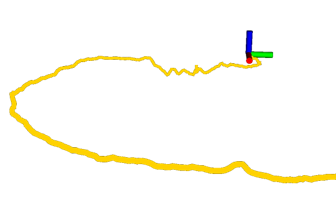}}}\hfill
 	\subfloat[]{\label{fig:l1_b}{
	\includegraphics[width=0.32\textwidth]{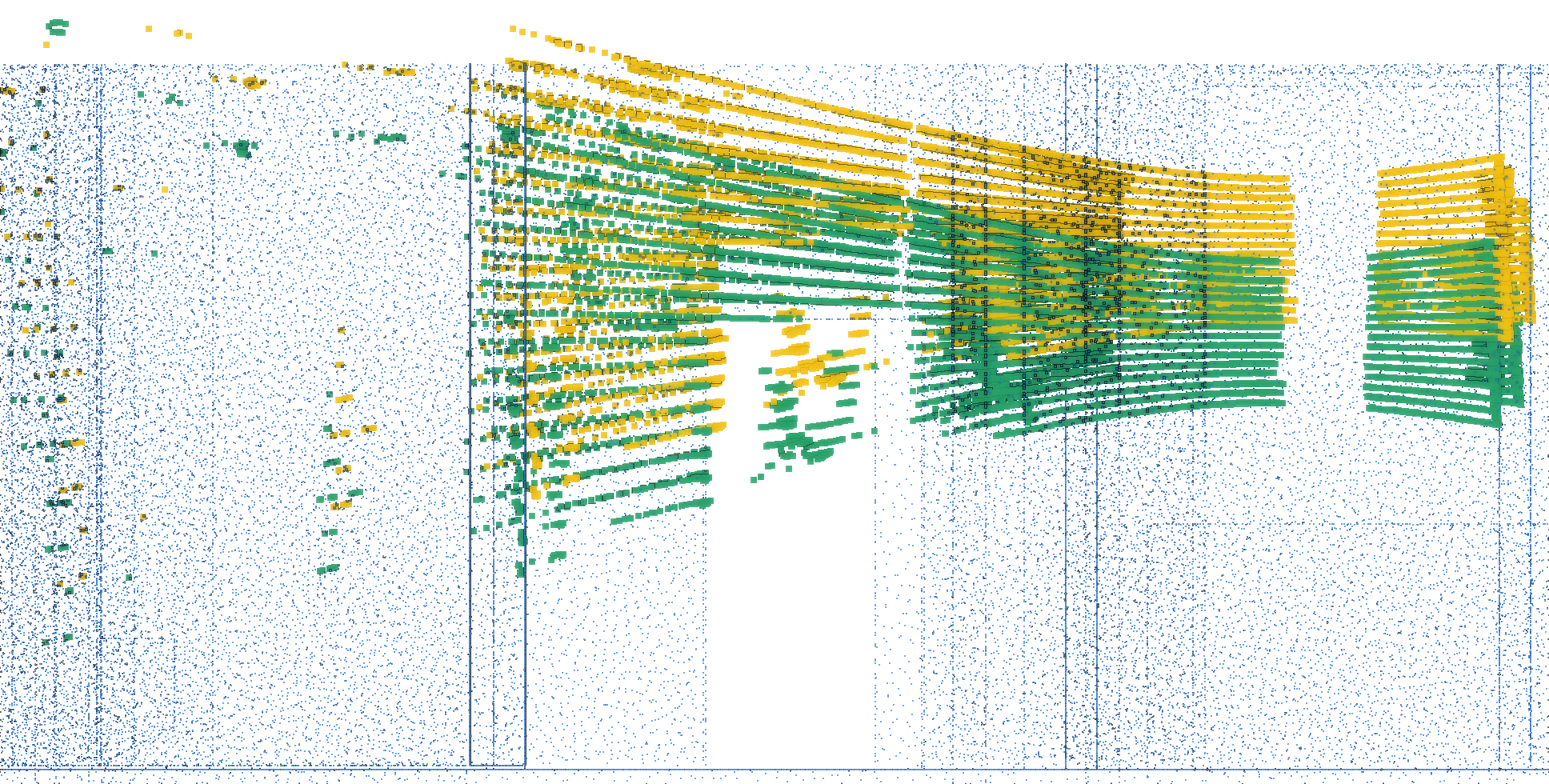}}}\hfill
	\subfloat[]{\label{fig:l1_c}{
	\includegraphics[width=0.32\textwidth]{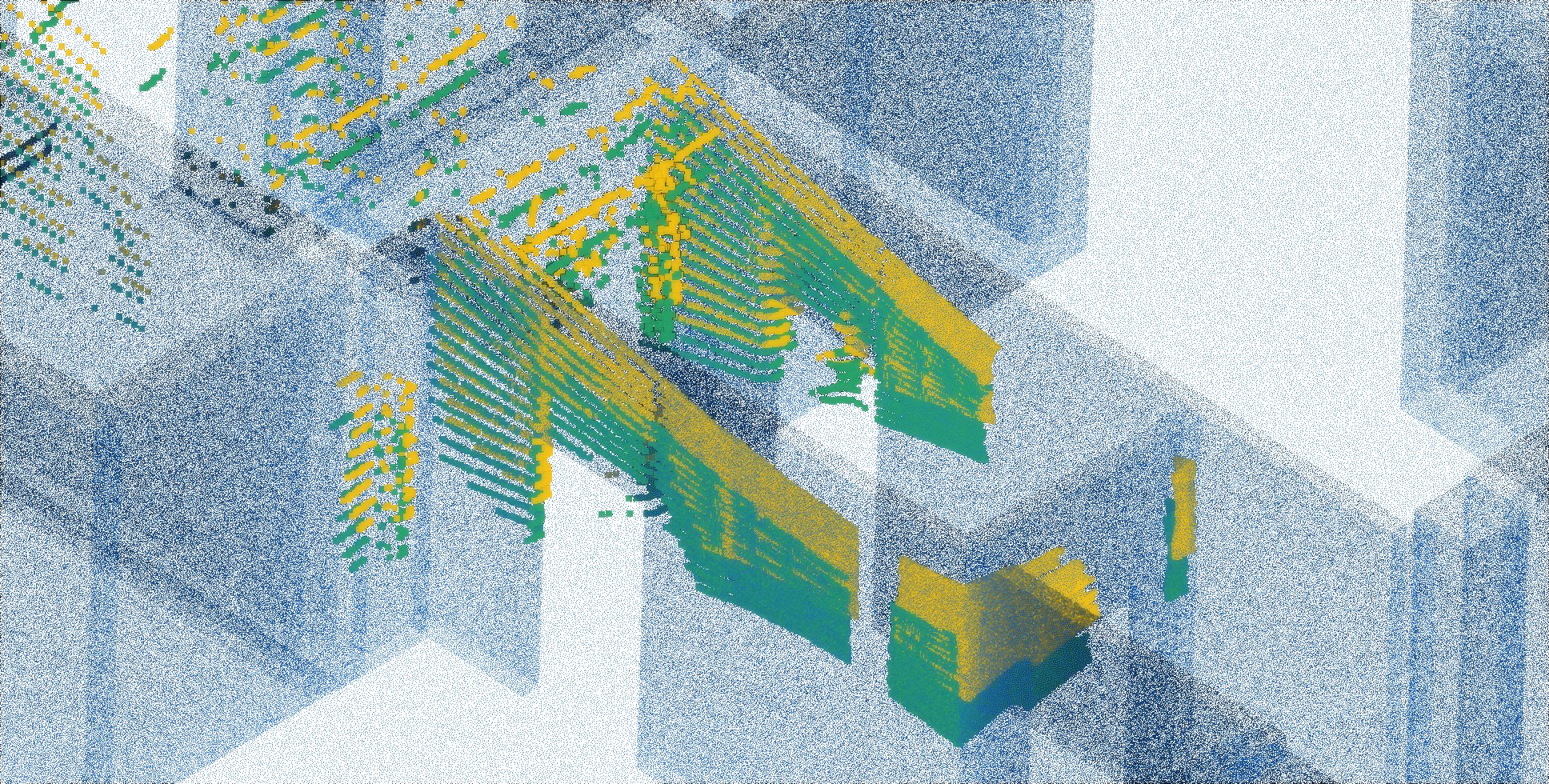}}}
    \caption{Limitation regarding the inaccuracy of initial poses: If the initially calculated poses drift in the Z direction, as depicted in (a), the final scan might not be automatically registered in the correct poses if they are located in narrow corridors. This occurs because specific scans, like the yellow one in (b) and (c), lack the ceiling or floor points necessary to determine the correct sensor height. (b) and (c) are the side and perspective views of the same scene. The green scan represents the manually correctly registered scan, while the blue depicts the reference map, in this case, a point cloud sampled from the BIM model.} \label{fig:limitation1}
\end{figure}

In particular, as depicted in Figure \ref{fig:limitation1}, when encountering a Z-drift within a narrow corridor, the limited information provided by individual scans regarding horizontal elements (such as floors or ceilings) can sometimes make automatic height retrieval very challenging.
One possible approach to mitigate this issue involves the utilization of the \textit{free space}, which can be calculated using OctoMap (as outlined in Section \ref{step:3}). 
Assuming most transition elements, such as doors and windows, are open during scanning, they could serve as reference elements for height retrieval by utilizing their frames as a feature for registration.

\nointerlineskip

% 4.2. Inaccuracies of the reference map (BIM)

Another limitation is related to the accuracy of the reference map. 
For instance, in scenarios characterized by substantial \ac{Scan-Map deviations}—such as significant discrepancies in the positions of permanent walls and columns, minimal overlap between the scan and the reference map, or symmetrical environments—our method may encounter challenges in achieving accurate alignment.

Even if the alignment is possible, the final ICP step might not yield correct results in the regions where the reference map is erroneous and the alignment process passes the specified fitness thresholds. 
This and the previous one are the main reasons why the results of the alignment with the BIM model (shown in Figure \ref{fig:seq2_BIM_results}) are not perfect.

Furthermore, the computational time can also be considered as a constraint. Given the requirements of high accuracy, our method does not achieve real-time performance, particularly during the final ICP step. This step involves registering individual scans with a dense point cloud, a process that can extend to several dozen minutes.

%  Limitations of ISCD
In addition, another constraint is evident in terms of the proposed ISCD and the corresponding loop detection step.
This descriptor, as well as the family of egocentric or SC descriptors, are specifically designed to work with LiDAR scans from a sensor with a horizontal \ac{FoV} of 360 deg. 
The reason behind this is mainly to be able to create rotational invariant 1D descriptors. 
Therefore, the pipeline is not directly compatible with data from sensors with a reduced \ac{FoV}, such as solid-state LiDARs and depth cameras.

\begin{figure}[!htb]
	\centering
	\subfloat[]{\label{fig:reflectionsA}{
	\includegraphics[width=0.35\textwidth]{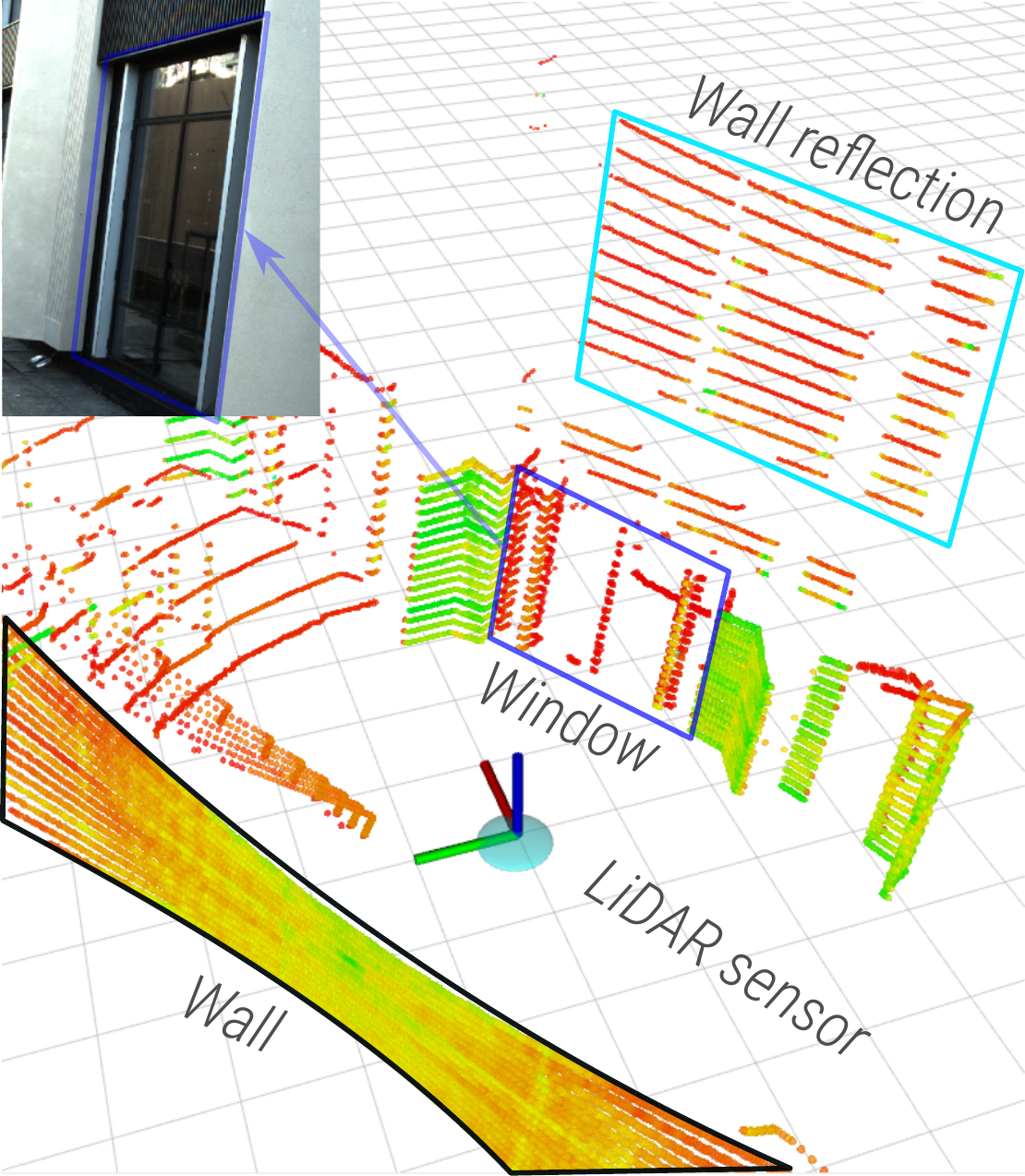}}}\hfill
	\subfloat[]{\label{fig:reflectionsB}{
	\includegraphics[width=0.63\textwidth]{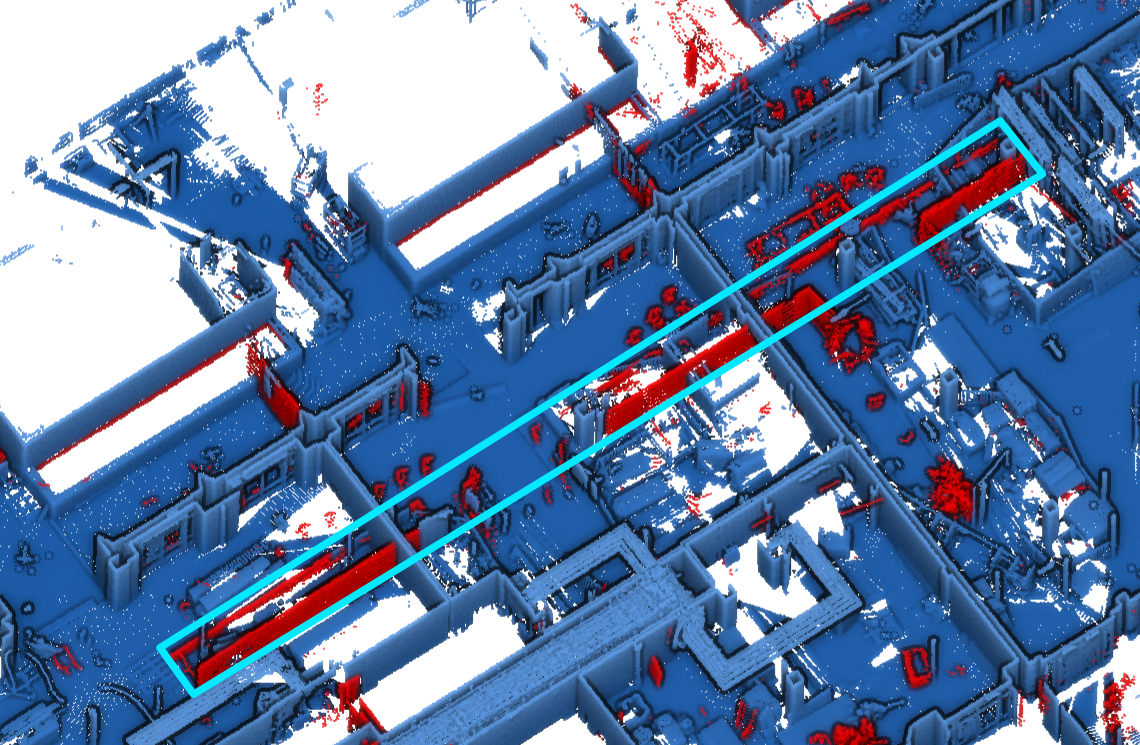}}}
    \caption{Limitation regarding wrongly detected changes following map alignment. In (a), the presence of reflective surfaces, such as windows, can lead to the generation of fictitious walls due to the LiDAR sensor's inability to filter out reflected measurements. This is illustrated in the top left, where an image of the actual window causing reflections is depicted, while at the bottom, the LiDAR measurement captures the wall, window, and the reflected fictitious wall. In (b), these reflections result in inaccurately detected changes in sequences 3, 4, and 5.}
\end{figure}

% Step 3 -> Change detection
In \textit{Step 3}, a primary limitation of our current pipeline emanate from reflections in window elements. As depicted in Figures \ref{fig:reflectionsA} and \ref{fig:reflectionsB}, these windows induce reflections in the LiDAR measurements, leading to the reconstruction of fictitious reflected elements within the detected changes. Specifically, the presence of reflected walls is noticeable in sequences 3, 4, and 5 (see last column of Figure \ref{fig:results_visual}. Notably, this issue is absent in sequence 2, attributed to the absence of window installations and the delimitation of the scan trajectory within the boundaries of walls with windows. 
To address this issue, one potential approach involves utilizing camera measurements to selectively filter out LiDAR measurements acquired in regions with window presence. Alternatively, a manual and more labor-intensive approach would involve physically occluding windows prior to scanning.

% 8: Conclusions & Future work 
\section{Conclusions}
\label{chap:conclusions}
% 1/3. Summary of Key Findings, 
% 2/3. Discussion of Implications, emphasizing the significance of the results (Motivation). 
% 3/3. Reflection on Limitations and Future Directions: suggestions for future research directions.

%%%%%%%%%%%%%%
% 1/3. Summary of Key Findings: 
This paper presents SLAM2REF, a modular framework to allow automatic 3D \ac{LiDAR} data alignment and change detection with a reference map, which can be a \ac{BIM model} or a point cloud. 

The framework operates independently of the sensor's initial position, eliminating the necessity for the scanning process to start within the provided map boundaries. 
Consequently, our framework enables map alignment and extension even when the reference map is outside the field of view (FoV) of the sensor or if only a portion of the map has been scanned.

Moreover, if an accurate \ac{TLS} point cloud is available, it can serve as a reference map to correct the poses of a query session and even retrieve centimeter-accurate ground truth poses.

More specifically, the following are our main contributions: 

\begin{itemize}

    \item A method to extract \ac{OGM}s out of complex \ac{BIM model}s or large-scale point clouds, which could serve to allow robot path planning and autonomous navigation in indoor GPS-denied environments.
    
    \item An efficient method to convert large-scale 3D maps into session data for fast 3D-LiDAR place recognition.

    \item A module that leverages quick place recognition and multi-session anchoring to allow the alignment and correction of drifted session acquired with \ac{SLAM} or odometry systems, given the presence of a reference map and considering motion distortion correction as in \citep{chen2023directDLIO}. 

    \item Provided that the reference map is accurate enough, the framework enables the retrieval of precise 6-\ac{DoF} poses of the entire trajectory, also enabling map extension.

    \item We introduced YawGICP, a robust implementation of GICP  tailored to effectively address registration problems primarily characterized by yaw angle variations.
 
    \item Similarly, we introduce \ac{ISC}, an innovative 2D descriptor that allows the alignment of the data with \ac{BIM model}s in scenarios with \ac{Scan-Map deviations}. 

    \item We present an extensive quantitative comparison of the steps in the pipeline, considering a state-of-the-art 3D\ac{LiDAR}-inertial odometry algorithm for pose initialization and exploiting the multiple sequences of the real-world open-access ConSLAM dataset.
    
    \item Finally, we provide a module that can detect positive and negative differences (i.e., when parts of the original reference map are no longer present in the environment) in the aligned map and create surfaces for better visualization with the \ac{BIM model} or reference point cloud. 
\end{itemize}

These functionalities collectively contribute to facilitating resilient long-term map data management, consolidating aligned relevant 3D information within a unified reference coordinate system.

% 2/3. Discussion of Implications, emphasizing the significance of the results (Motivation). 

In conclusion, SLAM2REF offers a novel solution to the challenges of lifelong mapping by integrating 3D LiDAR data and IMU measurements with a reference map, enabling automatic alignment, precise 6-DoF trajectory estimation, map extension, and change detection. 

By allowing \ac{Scan-Map deviations}, SLAM2REF offers a robust solution for automated 3D data alignment, even with as-designed BIM models that typically have significant deviations from as-built environments.

Our approach provides indirect support for the development of Digital Twins (DTs) for buildings, allowing the automatic alignment of newly acquired data with digital models. 
These models require continuous data integration to maintain its accuracy and relevance.

Practical applications are found in areas such as construction site monitoring, emergency response, disaster management, and others, where fast-updated digital 3D maps contribute to better decision-making and productivity.

Furthermore,  since our method is capable of exploiting BIM models that are semantically enhanced or point clouds as reference maps for localization, it can be used to support the development of autonomous robotic activities. 

Another advantage of SLAM2REF is that it advances SLAM research by enabling the automatic retrieval of centimeter-level accurate 6-DoF GT poses for large-scale indoor and outdoor trajectories.

% 3/3. Future Directions:
\section{Future Directions}
\label{chap:futurework}
In the future, we would like to extend the method, considering calibrated camera data, to leverage semantics from the real world, such as done by \citep{zimmerman2023}. By enriching camera measurements with semantic information, we anticipate better mitigation of the window reflection issue illustrated in Figure \ref{fig:reflectionsA}.
Additionally, an extended framework that is able to handle the alignment and correction of data with a reduced \ac{FoV} would allow it to work with data captured from solid-state LiDARs or depth cameras.

Further validation on more datasets, such as the Newer College \citep{Ramezani_2020, zhang2022multicamera}, or the Hilti \citep{hiltiChallenge2022} datasets, would give more insights into the robustness of the method.

Moreover, extending the efficiency and robustness of the method towards a real-time pipeline represents a promising direction for various tasks, including collaborative robot mapping and localization \citep{cramariuc2022maplab, Lajoie:2024:Swarm_SLAM}, such as to be able to solve the kidnapping robot problem in indoor environments with our proposed \ac{ISC} descriptor.
An essential aspect of achieving more robust alignment involves leveraging deep-learning-based place recognition algorithms, which are anticipated to become progressively reliable for indoor scenarios with sufficient training data in the future.

% Acknowledgement -------------------------------------------------------
\section{Acknowledgement}
The presented research was conducted in the frame of the project ``Intelligent Toolkit for Reconnaissance and assessmEnt in Perilous Incidents'' (INTREPID) funded by the EU's research and innovation funding programme Horizon 2020 under Grant agreement ID: 883345, as well as by the TUM Georg Nemetschek Institute in the frame of the AI4TWINNING project.

% Appendices 
\appendix
\section{Appendix}\label{appendix}

The table below lists the mathematical variables used in this manuscript, along with their corresponding descriptions.

\begin{longtable}{m{3.5cm}|m{9.5cm}}
\caption{Explanation of Variables} \label{tab:variables} \\
\hline
\textbf{Variable} & \textbf{Description} \\
\hline
\endfirsthead
\hline
\textbf{Variable} & \textbf{Description} \\
\hline
\endhead
\hline
\endfoot
\hline
\endlastfoot
$p(X|z)$ & Posterior density of the states $X$ given the measurements $Z$. \\
\hline
$F=(\mathcal{U}, \mathcal{V}, \mathcal{E})$ & Factor graph comprising nodes ($x_{i}$ or $\phi_{i}$) connected by edges $e_{ij}$. \\
\hline
$\mathcal{E}$ & Set of factor edges $e_{ij}$. \\
\hline
$\mathcal{U}$ & Set of factor nodes $\phi_{i}$. \\
\hline
$\mathcal{V}$ & Set of variable nodes $x_{i}$. \\
\hline
$X_{i}$ & Group of variables $x_{i}$ connected to a factor $\phi_{i}$. \\
\hline
$\phi(X)$ & Global function factorized as $\phi(X) = \prod_{i} \phi_{i}(X_{i})$. \\
\hline
$z_{i}$ & Measurements (observed data point or variable). \\
\hline
$h_{i}(x_i,l_i)$ & Mean measurement function of $x_i$ and $l_i$. Represents the expected value of $z_{i}$ given $x_{i}$ and $l_{i}$. \\
\hline
$\Sigma_{i}$ & Covariance matrix associated with $z_{i}$, representing zero-mean Gaussian noise. \\
\hline
$p(z_{i}|x_{i},l_i)$ & Conditional density on the measurement $z_{i}$. \\
\hline
$\mathcal{N}(z_{i};h_{i}(x_i,l_i),\Sigma_{i})$ & Multivariate normal distribution for the variable $z_{i}$ with mean $h_{i}(x_i,l_i)$ and covariance matrix $\Sigma_{i}$. \\
\hline
$\left\| h_{i}(x_i, l_i) - z_{i} \right\|_{\Sigma_{i}}^{2}$ & Mahalanobis distance between $z_{i}$ and its mean $h_{i}(x_i, l_i)$. \\
\hline
$\mathrm{SE}(3)$ & Special Euclidean group. \\
\hline
$\eta$ & Normally distributed zero-mean measurement noise with covariance $\Sigma_c$. \\
\hline
$\mathcal{S_R}$ & Synthetic reference session (created in Step 1, Section \ref{step:1}). \\
\hline
$\mathcal{S_Q}$ & Real-world motion-undistorted query session (created in Step 2.1, Section \ref{substep:2_1}). \\
\hline
$\mathbf{x}_{R}$ & Set of poses of the reference session. \\
\hline
$\mathbf{x}_{Q}$ & Set of poses of the query session. \\
\hline
$f(\cdot)$ & Odometry model function. \\
\hline
$\mathbf{u}_i^s$ & Constraints between consecutive poses $\mathbf{x}_{i}$ and $\mathbf{x}_{i+1}$. \\
\hline
$M_s$ & Number of poses in the session $\mathcal{S}$. \\
\hline
$N_e$ & Number of encounters between sessions. \\
\hline
$\mathbf{p}_s$ & Prior factor. \\
\hline
$\Delta_Q^*$ & Anchor node which facilitates the global alignment of the query session to the reference map. \\
\hline
$\mathcal{G}$ & Pose-graph map containing coordinates of pose nodes, odometry edges, and optionally recognized intra-session loop edges with uncertainty matrices. \\
\hline
$\left(\mathcal{P}_i, d_i\right)$ & Pairs of 3D LiDAR scans $\mathcal{P}_i$ with their corresponding global descriptors $d_i$ of the $i^{th}$ keyframe. \\
\hline
$n$ & Total number of equidistantly sampled keyframes. \\
\hline
$N_c$ & Amount of Top descriptors candidates selected from the reference session after the comparison of the rotational invariant descriptors. \\
\hline
$\Sigma_c$ & Covariance matrix of the detected loops or encounters incorporated into the factor graph problem as factors between sessions with anchoring. \\
\hline
$\mathbf{x}_Q^*$ & Optimized 6-DoF poses of each scan of the query session. \\
\hline
$\mathbf{c}$ & Loop closure detections, also called encounters denoting correspondences between the sessions. \\
\hline
$\mathbf{\nu}_Q$ & Confidence level list providing the reliability of each pose after scan registration. \\
\hline
$h(\cdot)$ & Original measurement model. \\
\hline
$h'(\cdot)$ & Modified measurement model that incorporates anchor nodes. \\
\hline
$\mathbf{x}_{R, j}$ & 6-DoF (in $\mathrm{SE}(3)$) Pose $j$ in the reference session. \\
\hline
$\mathbf{x}_{Q, j}$ &  6-DoF (in $\mathrm{SE}(3)$) Pose $j$ in the query session. \\
\hline
$\Delta_{R}$ & Anchor node for the reference session (also in (in $\mathrm{SE}(3)$). \\
\hline
$\Delta_{Q}$ & Anchor node for the query session (in $\mathrm{SE}(3)$). \\
\hline
$\mathbf{c}_j$ & Difference in the global frame between poses $\mathbf{x}_{R}$ and $\mathbf{x}_{Q}$ (pose in $\mathrm{SE}(3)$). \\
\hline
$\oplus$ & $\mathrm{SE}(3)$ pose composition operator. \\
\hline
$\ominus$ & $\mathrm{SE}(3)$ pose difference operator. \\
\hline
$\phi(\mathbf{x}_{R, i}, \mathbf{x}_{Q, j}, \Delta_{R}, \Delta_Q)$ & Factor between sessions with anchoring, used in pose graph optimization. \\
\hline
$\Sigma_P$ & Covariance assigned to the anchor node of the reference session, set to be insignificantly small. \\
\hline
$\Sigma_L$ & Covariance assigned to the anchor node of the query session, set to be significantly large. \\
\hline
${ }^Q \mathbf{x}_Q^*$ & Optimized poses of the query session in the local coordinate system. \\
\hline
${ }^W \mathbf{x}_Q^*$ & Optimized poses of the query session transformed to the global coordinate system of the reference map. \\
\hline
$W$ & Global coordinate system, same as the coordinate system of the reference session. \\
\hline
 $F_i$ & Fitness score distance threshold. The fitness score is the percentage of source inliers after point cloud registration, considering a maximum Point-to-Point (P2P) distance threshold. \\
\hline
\end{longtable}

% Bibliography
\bibliography{references}


\begin{thebibliography}{90}
\ifx \bisbn   \undefined \def \bisbn  #1{ISBN #1}\fi
\ifx \binits  \undefined \def \binits#1{#1}\fi
\ifx \bauthor  \undefined \def \bauthor#1{#1}\fi
\ifx \batitle  \undefined \def \batitle#1{#1}\fi
\ifx \bjtitle  \undefined \def \bjtitle#1{#1}\fi
\ifx \bvolume  \undefined \def \bvolume#1{\textbf{#1}}\fi
\ifx \byear  \undefined \def \byear#1{#1}\fi
\ifx \bissue  \undefined \def \bissue#1{#1}\fi
\ifx \bfpage  \undefined \def \bfpage#1{#1}\fi
\ifx \blpage  \undefined \def \blpage #1{#1}\fi
\ifx \burl  \undefined \def \burl#1{\textsf{#1}}\fi
\ifx \doiurl  \undefined \def \doiurl#1{\url{https://doi.org/#1}}\fi
\ifx \betal  \undefined \def \betal{\textit{et al.}}\fi
\ifx \binstitute  \undefined \def \binstitute#1{#1}\fi
\ifx \binstitutionaled  \undefined \def \binstitutionaled#1{#1}\fi
\ifx \bctitle  \undefined \def \bctitle#1{#1}\fi
\ifx \beditor  \undefined \def \beditor#1{#1}\fi
\ifx \bpublisher  \undefined \def \bpublisher#1{#1}\fi
\ifx \bbtitle  \undefined \def \bbtitle#1{#1}\fi
\ifx \bedition  \undefined \def \bedition#1{#1}\fi
\ifx \bseriesno  \undefined \def \bseriesno#1{#1}\fi
\ifx \blocation  \undefined \def \blocation#1{#1}\fi
\ifx \bsertitle  \undefined \def \bsertitle#1{#1}\fi
\ifx \bsnm \undefined \def \bsnm#1{#1}\fi
\ifx \bsuffix \undefined \def \bsuffix#1{#1}\fi
\ifx \bparticle \undefined \def \bparticle#1{#1}\fi
\ifx \barticle \undefined \def \barticle#1{#1}\fi
\bibcommenthead
\ifx \bconfdate \undefined \def \bconfdate #1{#1}\fi
\ifx \botherref \undefined \def \botherref #1{#1}\fi
\ifx \url \undefined \def \url#1{\textsf{#1}}\fi
\ifx \bchapter \undefined \def \bchapter#1{#1}\fi
\ifx \bbook \undefined \def \bbook#1{#1}\fi
\ifx \bcomment \undefined \def \bcomment#1{#1}\fi
\ifx \oauthor \undefined \def \oauthor#1{#1}\fi
\ifx \citeauthoryear \undefined \def \citeauthoryear#1{#1}\fi
\ifx \endbibitem  \undefined \def \endbibitem {}\fi
\ifx \bconflocation  \undefined \def \bconflocation#1{#1}\fi
\ifx \arxivurl  \undefined \def \arxivurl#1{\textsf{#1}}\fi
\csname PreBibitemsHook\endcsname

%%% 1
\bibitem[\protect\citeauthoryear{Alliez et~al.}{2020}]{alliez2020real}
\begin{bchapter}
\bauthor{\bsnm{Alliez}, \binits{P.}},
\bauthor{\bsnm{Bonardi}, \binits{F.}},
\bauthor{\bsnm{Bouchafa}, \binits{S.}},
\bauthor{\bsnm{Didier}, \binits{J.-Y.}},
\bauthor{\bsnm{Hadj-Abdelkader}, \binits{H.}},
\bauthor{\bsnm{Mu{\~n}oz}, \binits{F.I.}},
\bauthor{\bsnm{Kachurka}, \binits{V.}},
\bauthor{\bsnm{Rault}, \binits{B.}},
\bauthor{\bsnm{Robin}, \binits{M.}},
\bauthor{\bsnm{Roussel}, \binits{D.}}:
\bctitle{Real-time multi-slam system for agent localization and 3d mapping in dynamic scenarios}.
In: \bbtitle{2020 IEEE/RSJ International Conference on Intelligent Robots and Systems (IROS)},
pp. \bfpage{4894}--\blpage{4900}
(\byear{2020}).
\bcomment{IEEE}
\end{bchapter}
\endbibitem

%%% 2
\bibitem[\protect\citeauthoryear{Borrmann et~al.}{2024}]{bormann2024}
\begin{bchapter}
\bauthor{\bsnm{Borrmann}, \binits{A.}},
\bauthor{\bsnm{Biswanath}, \binits{M.}},
\bauthor{\bsnm{Braun}, \binits{A.}},
\bauthor{\bsnm{Chen}, \binits{Z.}},
\bauthor{\bsnm{Cremers}, \binits{D.}},
\bauthor{\bsnm{Heeramaglore}, \binits{M.}},
\bauthor{\bsnm{Hoegner}, \binits{L.}},
\bauthor{\bsnm{Mehranfar}, \binits{M.}},
\bauthor{\bsnm{Kolbe}, \binits{T.H.}},
\bauthor{\bsnm{Petzold}, \binits{F.}},
\bauthor{\bsnm{Rueda}, \binits{A.}},
\bauthor{\bsnm{Solonets}, \binits{S.}},
\bauthor{\bsnm{Zhu}, \binits{X.X.}}:
\bctitle{Artificial intelligence for the automated creation of multi-scale digital twins of the built world---ai4twinning}.
In: \beditor{\bsnm{Kolbe}, \binits{T.H.}},
\beditor{\bsnm{Donaubauer}, \binits{A.}},
\beditor{\bsnm{Beil}, \binits{C.}} (eds.)
\bbtitle{Recent Advances in 3D Geoinformation Science},
pp. \bfpage{233}--\blpage{247}.
\bpublisher{Springer},
\blocation{Cham}
(\byear{2024})
\end{bchapter}
\endbibitem

%%% 3
\bibitem[\protect\citeauthoryear{Boniardi et~al.}{2017}]{Boniardi.2017}
\begin{bchapter}
\bauthor{\bsnm{Boniardi}, \binits{F.}},
\bauthor{\bsnm{Caselitz}, \binits{T.}},
\bauthor{\bsnm{Kummerle}, \binits{R.}},
\bauthor{\bsnm{Burgard}, \binits{W.}}:
\bctitle{Robust lidar-based localization in architectural floor plans}.
In: \bbtitle{2017 IEEE/RSJ International Conference on Intelligent Robots and Systems (IROS)},
pp. \bfpage{3318}--\blpage{3324}.
\bpublisher{IEEE},
\blocation{Vancouver, BC, Canada}
(\byear{2017}).
\doiurl{10.1109/IROS.2017.8206168}
\end{bchapter}
\endbibitem

%%% 4
\bibitem[\protect\citeauthoryear{Boniardi et~al.}{2019}]{Boniardi.2019}
\begin{barticle}
\bauthor{\bsnm{Boniardi}, \binits{F.}},
\bauthor{\bsnm{Caselitz}, \binits{T.}},
\bauthor{\bsnm{K{\"u}mmerle}, \binits{R.}},
\bauthor{\bsnm{Burgard}, \binits{W.}}:
\batitle{A pose graph-based localization system for long-term navigation in cad floor plans}.
\bjtitle{Robotics and Autonomous Systems}
\bvolume{112},
\bfpage{84}--\blpage{97}
(\byear{2019})
\doiurl{10.1016/j.robot.2018.11.003}
\end{barticle}
\endbibitem

%%% 5
\bibitem[\protect\citeauthoryear{Blanco{-}Claraco}{2021}]{jose_tutorial2021}
\begin{botherref}
\oauthor{\bsnm{Blanco{-}Claraco}, \binits{J.L.}}:
A tutorial on $\mathbf{SE(3)}$ transformation parameterizations and on-manifold optimization.
CoRR
\textbf{abs/2103.15980}
(2021)
{\href{https://arxiv.org/abs/2103.15980}{{2103.15980}}}
\end{botherref}
\endbibitem

%%% 6
\bibitem[\protect\citeauthoryear{Besl and McKay}{1992}]{p2p_ICP:1992}
\begin{barticle}
\bauthor{\bsnm{Besl}, \binits{P.J.}},
\bauthor{\bsnm{McKay}, \binits{N.D.}}:
\batitle{A method for registration of 3-d shapes}.
\bjtitle{IEEE Transactions on Pattern Analysis and Machine Intelligence}
\bvolume{14}(\bissue{2}),
\bfpage{239}--\blpage{256}
(\byear{1992})
\doiurl{10.1109/34.121791}
\end{barticle}
\endbibitem

%%% 7
\bibitem[\protect\citeauthoryear{Blum et~al.}{2021}]{Blum.2021}
\begin{botherref}
\oauthor{\bsnm{Blum}, \binits{H.}},
\oauthor{\bsnm{Milano}, \binits{F.}},
\oauthor{\bsnm{Zurbr{\"{u}}gg}, \binits{R.}},
\oauthor{\bsnm{Siegwart}, \binits{R.}},
\oauthor{\bsnm{Cadena}, \binits{C.}},
\oauthor{\bsnm{Gawel}, \binits{A.}}:
Self-improving semantic perception on a construction robot.
CoRR
\textbf{abs/2105.01595}
(2021)
{\href{https://arxiv.org/abs/2105.01595}{{2105.01595}}}
\end{botherref}
\endbibitem

%%% 8
\bibitem[\protect\citeauthoryear{Blanco and Rai}{2014}]{blanco2014nanoflann}
\begin{botherref}
\oauthor{\bsnm{Blanco}, \binits{J.L.}},
\oauthor{\bsnm{Rai}, \binits{P.K.}}:
{nanoflann: a C++ header-only fork of FLANN, a library for Nearest Neighbor (NN) wih KD-trees}
(2014)
\end{botherref}
\endbibitem

%%% 9
\bibitem[\protect\citeauthoryear{Blum et~al.}{2020}]{Blum.2020}
\begin{botherref}
\oauthor{\bsnm{Blum}, \binits{H.}},
\oauthor{\bsnm{Stiefel}, \binits{J.}},
\oauthor{\bsnm{Cadena}, \binits{C.}},
\oauthor{\bsnm{Siegwart}, \binits{R.}},
\oauthor{\bsnm{Gawel}, \binits{A.}}:
Precise robot localization in architectural 3d plans.
arXiv preprint arXiv:2006.05137
(2020)
\end{botherref}
\endbibitem

%%% 10
\bibitem[\protect\citeauthoryear{Botín-Sanabria et~al.}{2022}]{Botin-Sanabria:2022:digitalTwin}
\begin{botherref}
\oauthor{\bsnm{Botín-Sanabria}, \binits{D.M.}},
\oauthor{\bsnm{Mihaita}, \binits{A.-S.}},
\oauthor{\bsnm{Peimbert-García}, \binits{R.E.}},
\oauthor{\bsnm{Ramírez-Moreno}, \binits{M.A.}},
\oauthor{\bsnm{Ramírez-Mendoza}, \binits{R.A.}},
\oauthor{\bsnm{Lozoya-Santos}, \binits{J.d.J.}}:
Digital twin technology challenges and applications: A comprehensive review.
Remote Sensing
\textbf{14}(6)
(2022)
\doiurl{10.3390/rs14061335}
\end{botherref}
\endbibitem

%%% 11
\bibitem[\protect\citeauthoryear{Bai et~al.}{2022}]{fasterLio2022}
\begin{barticle}
\bauthor{\bsnm{Bai}, \binits{C.}},
\bauthor{\bsnm{Xiao}, \binits{T.}},
\bauthor{\bsnm{Chen}, \binits{Y.}},
\bauthor{\bsnm{Wang}, \binits{H.}},
\bauthor{\bsnm{Zhang}, \binits{F.}},
\bauthor{\bsnm{Gao}, \binits{X.}}:
\batitle{Faster-lio: Lightweight tightly coupled lidar-inertial odometry using parallel sparse incremental voxels}.
\bjtitle{IEEE Robotics and Automation Letters}
\bvolume{7}(\bissue{2}),
\bfpage{4861}--\blpage{4868}
(\byear{2022})
\doiurl{10.1109/LRA.2022.3152830}
\end{barticle}
\endbibitem

%%% 12
\bibitem[\protect\citeauthoryear{Cramariuc et~al.}{2022}]{cramariuc2022maplab}
\begin{botherref}
\oauthor{\bsnm{Cramariuc}, \binits{A.}},
\oauthor{\bsnm{Bernreiter}, \binits{L.}},
\oauthor{\bsnm{Tschopp}, \binits{F.}},
\oauthor{\bsnm{Fehr}, \binits{M.}},
\oauthor{\bsnm{Reijgwart}, \binits{V.}},
\oauthor{\bsnm{Nieto}, \binits{J.}},
\oauthor{\bsnm{Siegwart}, \binits{R.}},
\oauthor{\bsnm{Cadena}, \binits{C.}}:
maplab 2.0--a modular and multi-modal mapping framework.
IEEE Robotics and Automation Letters
(2022)
\end{botherref}
\endbibitem

%%% 13
\bibitem[\protect\citeauthoryear{Caballero and Merino}{2021}]{caballero2021dll}
\begin{bchapter}
\bauthor{\bsnm{Caballero}, \binits{F.}},
\bauthor{\bsnm{Merino}, \binits{L.}}:
\bctitle{Dll: Direct {LiDAR} localization. a map-based localization approach for aerial robots}.
In: \bbtitle{2021 IEEE/RSJ International Conference on Intelligent Robots and Systems (IROS)},
pp. \bfpage{5491}--\blpage{5498}
(\byear{2021}).
\bcomment{IEEE}
\end{bchapter}
\endbibitem

%%% 14
\bibitem[\protect\citeauthoryear{Chen et~al.}{2023}]{chen2023directDLIO}
\begin{bchapter}
\bauthor{\bsnm{Chen}, \binits{K.}},
\bauthor{\bsnm{Nemiroff}, \binits{R.}},
\bauthor{\bsnm{Lopez}, \binits{B.T.}}:
\bctitle{Direct lidar-inertial odometry: Lightweight lio with continuous-time motion correction}.
In: \bbtitle{2023 IEEE International Conference on Robotics and Automation (ICRA)},
pp. \bfpage{3983}--\blpage{3989}
(\byear{2023}).
\bcomment{IEEE}
\end{bchapter}
\endbibitem

%%% 15
\bibitem[\protect\citeauthoryear{Chen et~al.}{2024}]{iglio2024}
\begin{botherref}
\oauthor{\bsnm{Chen}, \binits{Z.}},
\oauthor{\bsnm{Xu}, \binits{Y.}},
\oauthor{\bsnm{Yuan}, \binits{S.}},
\oauthor{\bsnm{Xie}, \binits{L.}}:
ig-lio: An incremental gicp-based tightly-coupled lidar-inertial odometry.
IEEE Robotics and Automation Letters,
1--8
(2024)
\doiurl{10.1109/LRA.2024.3349915}
\end{botherref}
\endbibitem

%%% 16
\bibitem[\protect\citeauthoryear{Dugstad et~al.}{2022}]{Dugstad:2022:SAS-Indoor-PP}
\begin{bchapter}
\bauthor{\bsnm{Dugstad}, \binits{A.}},
\bauthor{\bsnm{Dubey}, \binits{R.K.}},
\bauthor{\bsnm{Abualdenien}, \binits{J.}},
\bauthor{\bsnm{Borrmann}, \binits{A.}}:
\bctitle{{BIM}-based disaster response: Facilitating indoor path planning for various agents}.
In: \bbtitle{Proc. of European Conference on Product and Process Modeling 2022},
pp. \bfpage{265}--\blpage{289}
(\byear{2022})
\end{bchapter}
\endbibitem

%%% 17
\bibitem[\protect\citeauthoryear{Dellaert}{2021}]{dellaert2021factor}
\begin{barticle}
\bauthor{\bsnm{Dellaert}, \binits{F.}}:
\batitle{Factor graphs: Exploiting structure in robotics}.
\bjtitle{Annual Review of Control, Robotics, and Autonomous Systems}
\bvolume{4},
\bfpage{141}--\blpage{166}
(\byear{2021})
\end{barticle}
\endbibitem

%%% 18
\bibitem[\protect\citeauthoryear{Dellaert et~al.}{2017}]{dellaert2017factor}
\begin{barticle}
\bauthor{\bsnm{Dellaert}, \binits{F.}},
\bauthor{\bsnm{Kaess}, \binits{M.}}, \betal:
\batitle{Factor graphs for robot perception}.
\bjtitle{Foundations and Trends in Robotics}
\bvolume{6}(\bissue{1-2}),
\bfpage{1}--\blpage{139}
(\byear{2017})
\end{barticle}
\endbibitem

%%% 19
\bibitem[\protect\citeauthoryear{de~Teruel et~al.}{2017}]{LOPEZDETERUEL2017_wifi}
\begin{barticle}
\bauthor{\bsnm{Lopez-de-Teruel}, \binits{P.E.}},
\bauthor{\bsnm{Garcia}, \binits{F.J.}},
\bauthor{\bsnm{Canovas}, \binits{O.}},
\bauthor{\bsnm{Gonzalez}, \binits{R.}},
\bauthor{\bsnm{Carrasco}, \binits{J.A.}}:
\batitle{Human behavior monitoring using a passive indoor positioning system: a case study in a sme}.
\bjtitle{Procedia Computer Science}
\bvolume{110},
\bfpage{182}--\blpage{189}
(\byear{2017})
\doiurl{10.1016/j.procs.2017.06.076} .
\bcomment{14th International Conference on Mobile Systems and Pervasive Computing (MobiSPC 2017) / 12th International Conference on Future Networks and Communications (FNC 2017) / Affiliated Workshops}
\end{barticle}
\endbibitem

%%% 20
\bibitem[\protect\citeauthoryear{Ercan et~al.}{2020}]{ErcanJenny.2020}
\begin{bchapter}
\bauthor{\bsnm{Ercan}, \binits{S.}},
\bauthor{\bsnm{Blum}, \binits{H.}},
\bauthor{\bsnm{Gawel}, \binits{A.}},
\bauthor{\bsnm{Siegwart}, \binits{R.}},
\bauthor{\bsnm{Gramazio}, \binits{F.}},
\bauthor{\bsnm{Kohler}, \binits{M.}}:
\bctitle{Online synchronization of building model for on-site mobile robotic construction}.
In: \bbtitle{37th International Symposium on Automation and Robotics in Construction (ISARC 2020)(virtual)},
pp. \bfpage{1508}--\blpage{1514}
(\byear{2020}).
\bcomment{International Association for Automation and Robotics in Construction}
\end{bchapter}
\endbibitem

%%% 21
\bibitem[\protect\citeauthoryear{Forster et~al.}{2016}]{forster2016manifold}
\begin{barticle}
\bauthor{\bsnm{Forster}, \binits{C.}},
\bauthor{\bsnm{Carlone}, \binits{L.}},
\bauthor{\bsnm{Dellaert}, \binits{F.}},
\bauthor{\bsnm{Scaramuzza}, \binits{D.}}:
\batitle{On-manifold preintegration for real-time visual--inertial odometry}.
\bjtitle{IEEE Transactions on Robotics}
\bvolume{33}(\bissue{1}),
\bfpage{1}--\blpage{21}
(\byear{2016})
\end{barticle}
\endbibitem

%%% 22
\bibitem[\protect\citeauthoryear{Follini et~al.}{2020}]{Follini.2020}
\begin{barticle}
\bauthor{\bsnm{Follini}, \binits{C.}},
\bauthor{\bsnm{Magnago}, \binits{V.}},
\bauthor{\bsnm{Freitag}, \binits{K.}},
\bauthor{\bsnm{Terzer}, \binits{M.}},
\bauthor{\bsnm{Marcher}, \binits{C.}},
\bauthor{\bsnm{Riedl}, \binits{M.}},
\bauthor{\bsnm{Giusti}, \binits{A.}},
\bauthor{\bsnm{Matt}, \binits{D.T.}}:
\batitle{{BIM}-integrated collaborative robotics for application in building construction and maintenance}.
\bjtitle{Robotics}
\bvolume{10}(\bissue{1}),
\bfpage{2}
(\byear{2020})
\doiurl{10.3390/robotics10010002}
\end{barticle}
\endbibitem

%%% 23
\bibitem[\protect\citeauthoryear{Gawel et~al.}{2019}]{gawel2019fully}
\begin{bchapter}
\bauthor{\bsnm{Gawel}, \binits{A.}},
\bauthor{\bsnm{Blum}, \binits{H.}},
\bauthor{\bsnm{Pankert}, \binits{J.}},
\bauthor{\bsnm{Kr{\"a}mer}, \binits{K.}},
\bauthor{\bsnm{Bartolomei}, \binits{L.}},
\bauthor{\bsnm{Ercan}, \binits{S.}},
\bauthor{\bsnm{Farshidian}, \binits{F.}},
\bauthor{\bsnm{Chli}, \binits{M.}},
\bauthor{\bsnm{Gramazio}, \binits{F.}},
\bauthor{\bsnm{Siegwart}, \binits{R.}}, \betal:
\bctitle{A fully-integrated sensing and control system for high-accuracy mobile robotic building construction}.
In: \bbtitle{2019 IEEE/RSJ International Conference on Intelligent Robots and Systems (IROS)},
pp. \bfpage{2300}--\blpage{2307}
(\byear{2019}).
\bcomment{IEEE}
\end{bchapter}
\endbibitem

%%% 24
\bibitem[\protect\citeauthoryear{Gschwandtner et~al.}{2011}]{blensor2011}
\begin{bchapter}
\bauthor{\bsnm{Gschwandtner}, \binits{M.}},
\bauthor{\bsnm{Kwitt}, \binits{R.}},
\bauthor{\bsnm{Uhl}, \binits{A.}},
\bauthor{\bsnm{Pree}, \binits{W.}}:
\bctitle{Blensor: Blender sensor simulation toolbox}.
In: \beditor{\bsnm{Bebis}, \binits{G.}},
\beditor{\bsnm{Boyle}, \binits{R.}},
\beditor{\bsnm{Parvin}, \binits{B.}},
\beditor{\bsnm{Koracin}, \binits{D.}},
\beditor{\bsnm{Wang}, \binits{S.}},
\beditor{\bsnm{Kyungnam}, \binits{K.}},
\beditor{\bsnm{Benes}, \binits{B.}},
\beditor{\bsnm{Moreland}, \binits{K.}},
\beditor{\bsnm{Borst}, \binits{C.}},
\beditor{\bsnm{DiVerdi}, \binits{S.}},
\beditor{\bsnm{Yi-Jen}, \binits{C.}},
\beditor{\bsnm{Ming}, \binits{J.}} (eds.)
\bbtitle{Advances in Visual Computing},
pp. \bfpage{199}--\blpage{208}.
\bpublisher{Springer},
\blocation{Berlin, Heidelberg}
(\byear{2011})
\end{bchapter}
\endbibitem

%%% 25
\bibitem[\protect\citeauthoryear{Geiger et~al.}{2013}]{kitti_raw_Geiger2013IJRR}
\begin{botherref}
\oauthor{\bsnm{Geiger}, \binits{A.}},
\oauthor{\bsnm{Lenz}, \binits{P.}},
\oauthor{\bsnm{Stiller}, \binits{C.}},
\oauthor{\bsnm{Urtasun}, \binits{R.}}:
Vision meets robotics: The kitti dataset.
International Journal of Robotics Research (IJRR)
(2013)
\end{botherref}
\endbibitem

%%% 26
\bibitem[\protect\citeauthoryear{Grupp}{2017}]{grupp2017evo}
\begin{botherref}
\oauthor{\bsnm{Grupp}, \binits{M.}}:
evo: Python package for the evaluation of odometry and {SLAM}.
\url{https://github.com/MichaelGrupp/evo}
(2017)
\end{botherref}
\endbibitem

%%% 27
\bibitem[\protect\citeauthoryear{Gschwandtner}{2013}]{BlenSor_PhdThesis}
\begin{botherref}
\oauthor{\bsnm{Gschwandtner}, \binits{M.}}:
Support framework for obstacle detection on autonomous trains.
PhD thesis,
Department of Computer Sciences, University of Salzburg
(2013)
\end{botherref}
\endbibitem

%%% 28
\bibitem[\protect\citeauthoryear{Hendrikx et~al.}{2022}]{hendrikx2022local}
\begin{botherref}
\oauthor{\bsnm{Hendrikx}, \binits{R.}},
\oauthor{\bsnm{Bruyninckx}, \binits{H.}},
\oauthor{\bsnm{Elfring}, \binits{J.}},
\oauthor{\bsnm{Van De~Molengraft}, \binits{M.}}:
Local-to-global hypotheses for robust robot localization.
Frontiers in Robotics and AI,
171
(2022)
\end{botherref}
\endbibitem

%%% 29
\bibitem[\protect\citeauthoryear{Hess et~al.}{2016}]{Hess.2016b}
\begin{bchapter}
\bauthor{\bsnm{Hess}, \binits{W.}},
\bauthor{\bsnm{Kohler}, \binits{D.}},
\bauthor{\bsnm{Rapp}, \binits{H.}},
\bauthor{\bsnm{Andor}, \binits{D.}}:
\bctitle{Real-time loop closure in 2d {LiDAR} {SLAM}}.
In: \bbtitle{2016 IEEE International Conference on Robotics and Automation (ICRA)},
pp. \bfpage{1271}--\blpage{1278}.
\bpublisher{IEEE},
\blocation{Stockholm, Sweden}
(\byear{2016}).
\doiurl{10.1109/ICRA.2016.7487258}
\end{bchapter}
\endbibitem

%%% 30
\bibitem[\protect\citeauthoryear{Hendrikx et~al.}{2021}]{Hendrikx.2021}
\begin{bchapter}
\bauthor{\bsnm{Hendrikx}, \binits{R.W.M.}},
\bauthor{\bsnm{Pauwels}, \binits{P.}},
\bauthor{\bsnm{Torta}, \binits{E.}},
\bauthor{\bsnm{Bruyninckx}, \binits{H.J.P.}},
\bauthor{\bsnm{{van de Molengraft}}, \binits{M.J.G.}}:
\bctitle{Connecting semantic building information models and robotics: An application to 2d lidar-based localization}.
In: \bbtitle{2021 IEEE International Conference on Robotics and Automation (ICRA)},
pp. \bfpage{11654}--\blpage{11660}.
\bpublisher{IEEE},
\blocation{Xi'an, China}
(\byear{2021}).
\doiurl{10.1109/ICRA48506.2021.9561129}
\end{bchapter}
\endbibitem

%%% 31
\bibitem[\protect\citeauthoryear{Hornung et~al.}{2013}]{octomap}
\begin{barticle}
\bauthor{\bsnm{Hornung}, \binits{A.}},
\bauthor{\bsnm{Wurm}, \binits{K.M.}},
\bauthor{\bsnm{Bennewitz}, \binits{M.}},
\bauthor{\bsnm{Stachniss}, \binits{C.}},
\bauthor{\bsnm{Burgard}, \binits{W.}}:
\batitle{Octomap: An efficient probabilistic 3d mapping framework based on octrees}.
\bjtitle{Autonomous robots}
\bvolume{34},
\bfpage{189}--\blpage{206}
(\byear{2013})
\end{barticle}
\endbibitem

%%% 32
\bibitem[\protect\citeauthoryear{He et~al.}{2021}]{Guo2021MappingAndLocalization}
\begin{bchapter}
\bauthor{\bsnm{He}, \binits{G.}},
\bauthor{\bsnm{Zhang}, \binits{F.}},
\bauthor{\bsnm{Li}, \binits{X.}},
\bauthor{\bsnm{Shang}, \binits{W.}}:
\bctitle{Robust mapping and localization in offline 3d point cloud maps}.
In: \bbtitle{2021 6th IEEE International Conference on Advanced Robotics and Mechatronics (ICARM)},
pp. \bfpage{765}--\blpage{770}
(\byear{2021}).
\doiurl{10.1109/ICARM52023.2021.9536181}
\end{bchapter}
\endbibitem

%%% 33
\bibitem[\protect\citeauthoryear{{IfcOpenShell Contributors}}{2023a}]{IfcConvert}
\begin{botherref}
\oauthor{\bsnm{{IfcOpenShell Contributors}}}:
IfcConvert: An application for converting {IFC} geometry into several file formats.
Software.
URL: \url{https://ifcopenshell.sourceforge.net/ifcconvert.html}
(2023).
\url{https://ifcopenshell.sourceforge.net/ifcconvert.html}
\end{botherref}
\endbibitem

%%% 34
\bibitem[\protect\citeauthoryear{{IfcOpenShell Contributors}}{2023b}]{Ifcconvert_documentation}
\begin{botherref}
\oauthor{\bsnm{{IfcOpenShell Contributors}}}:
Ifcconvert documentation.
Online.
URL: \url{https://blenderbim.org/docs-python/ifcconvert/usage.html}
(2023).
\url{https://blenderbim.org/docs-python/ifcconvert/usage.html}
\end{botherref}
\endbibitem

%%% 35
\bibitem[\protect\citeauthoryear{Jurić et~al.}{2021}]{juric2021Comparison}
\begin{bchapter}
\bauthor{\bsnm{Jurić}, \binits{A.}},
\bauthor{\bsnm{Kendeš}, \binits{F.}},
\bauthor{\bsnm{Marković}, \binits{I.}},
\bauthor{\bsnm{Petrović}, \binits{I.}}:
\bctitle{A comparison of graph optimization approaches for pose estimation in {SLAM}}.
In: \bbtitle{2021 44th International Convention on Information, Communication and Electronic Technology (MIPRO)},
pp. \bfpage{1113}--\blpage{1118}
(\byear{2021}).
\doiurl{10.23919/MIPRO52101.2021.9596721}
\end{bchapter}
\endbibitem

%%% 36
\bibitem[\protect\citeauthoryear{Karimi et~al.}{2021}]{Karimi.21.04.2021}
\begin{botherref}
\oauthor{\bsnm{Karimi}, \binits{S.}},
\oauthor{\bsnm{Braga}, \binits{R.G.}},
\oauthor{\bsnm{Iordanova}, \binits{I.}},
\oauthor{\bsnm{St-Onge}, \binits{D.}}:
Semantic Navigation Using Building Information on Construction Sites
(2021).
\url{http://arxiv.org/pdf/2104.10296v1}
\end{botherref}
\endbibitem

%%% 37
\bibitem[\protect\citeauthoryear{Kim et~al.}{2021}]{kim2021scan}
\begin{barticle}
\bauthor{\bsnm{Kim}, \binits{G.}},
\bauthor{\bsnm{Choi}, \binits{S.}},
\bauthor{\bsnm{Kim}, \binits{A.}}:
\batitle{Scan context++: Structural place recognition robust to rotation and lateral variations in urban environments}.
\bjtitle{IEEE Transactions on Robotics}
\bvolume{38}(\bissue{3}),
\bfpage{1856}--\blpage{1874}
(\byear{2021})
\end{barticle}
\endbibitem

%%% 38
\bibitem[\protect\citeauthoryear{Koenig and Howard}{2004}]{Gazebo_koenig2004design}
\begin{bchapter}
\bauthor{\bsnm{Koenig}, \binits{N.}},
\bauthor{\bsnm{Howard}, \binits{A.}}:
\bctitle{Design and use paradigms for gazebo, an open-source multi-robot simulator}.
In: \bbtitle{2004 IEEE/RSJ International Conference on Intelligent Robots and Systems (IROS)(IEEE Cat. No. 04CH37566)},
vol. \bseriesno{3},
pp. \bfpage{2149}--\blpage{2154}
(\byear{2004}).
\bcomment{IEEE}
\end{bchapter}
\endbibitem

%%% 39
\bibitem[\protect\citeauthoryear{Karimi et~al.}{2020}]{Karimi.2020}
\begin{botherref}
\oauthor{\bsnm{Karimi}, \binits{S.}},
\oauthor{\bsnm{Iordanova}, \binits{I.}},
\oauthor{\bsnm{St-Onge}, \binits{D.}}:
An ontology-based approach to data exchanges for robot navigation on construction sites.
Journal of Information Technology in Construction
(2020)
\end{botherref}
\endbibitem

%%% 40
\bibitem[\protect\citeauthoryear{Kim and Kim}{2018}]{kim2018scan}
\begin{bchapter}
\bauthor{\bsnm{Kim}, \binits{G.}},
\bauthor{\bsnm{Kim}, \binits{A.}}:
\bctitle{Scan context: Egocentric spatial descriptor for place recognition within 3d point cloud map}.
In: \bbtitle{2018 IEEE/RSJ International Conference on Intelligent Robots and Systems (IROS)},
pp. \bfpage{4802}--\blpage{4809}
(\byear{2018}).
\bcomment{IEEE}
\end{bchapter}
\endbibitem

%%% 41
\bibitem[\protect\citeauthoryear{Kim and Kim}{2022}]{kim2022ltMapper}
\begin{bchapter}
\bauthor{\bsnm{Kim}, \binits{G.}},
\bauthor{\bsnm{Kim}, \binits{A.}}:
\bctitle{Lt-mapper: A modular framework for lidar-based lifelong mapping}.
In: \bbtitle{2022 International Conference on Robotics and Automation (ICRA)},
pp. \bfpage{7995}--\blpage{8002}
(\byear{2022}).
\bcomment{IEEE}
\end{bchapter}
\endbibitem

%%% 42
\bibitem[\protect\citeauthoryear{Kim et~al.}{2010}]{kim2010multiple}
\begin{bchapter}
\bauthor{\bsnm{Kim}, \binits{B.}},
\bauthor{\bsnm{Kaess}, \binits{M.}},
\bauthor{\bsnm{Fletcher}, \binits{L.}},
\bauthor{\bsnm{Leonard}, \binits{J.}},
\bauthor{\bsnm{Bachrach}, \binits{A.}},
\bauthor{\bsnm{Roy}, \binits{N.}},
\bauthor{\bsnm{Teller}, \binits{S.}}:
\bctitle{Multiple relative pose graphs for robust cooperative mapping}.
In: \bbtitle{2010 IEEE International Conference on Robotics and Automation},
pp. \bfpage{3185}--\blpage{3192}
(\byear{2010}).
\bcomment{IEEE}
\end{bchapter}
\endbibitem

%%% 43
\bibitem[\protect\citeauthoryear{Koide et~al.}{2022}]{Koide2022_tags}
\begin{bchapter}
\bauthor{\bsnm{Koide}, \binits{K.}},
\bauthor{\bsnm{Oishi}, \binits{S.}},
\bauthor{\bsnm{Yokozuka}, \binits{M.}},
\bauthor{\bsnm{Banno}, \binits{A.}}:
\bctitle{Scalable fiducial tag localization on a 3d prior map via graph-theoretic global tag-map registration}.
In: \bbtitle{2022 IEEE/RSJ International Conference on Intelligent Robots and Systems (IROS)},
pp. \bfpage{5347}--\blpage{5353}
(\byear{2022}).
\doiurl{10.1109/IROS47612.2022.9981079}
\end{bchapter}
\endbibitem

%%% 44
\bibitem[\protect\citeauthoryear{Kim and Peavy}{2022}]{Kim.2022}
\begin{barticle}
\bauthor{\bsnm{Kim}, \binits{K.}},
\bauthor{\bsnm{Peavy}, \binits{M.}}:
\batitle{{BIM}-based semantic building world modeling for robot task planning and execution in built environments}.
\bjtitle{Automation in Construction}
\bvolume{138},
\bfpage{104247}
(\byear{2022})
\doiurl{10.1016/j.autcon.2022.104247}
\end{barticle}
\endbibitem

%%% 45
\bibitem[\protect\citeauthoryear{Kim et~al.}{2021}]{Kim.2021}
\begin{barticle}
\bauthor{\bsnm{Kim}, \binits{S.}},
\bauthor{\bsnm{Peavy}, \binits{M.}},
\bauthor{\bsnm{Huang}, \binits{P.-C.}},
\bauthor{\bsnm{Kim}, \binits{K.}}:
\batitle{Development of {BIM}-integrated construction robot task planning and simulation system}.
\bjtitle{Automation in Construction}
\bvolume{127},
\bfpage{103720}
(\byear{2021})
\doiurl{10.1016/j.autcon.2021.103720}
\end{barticle}
\endbibitem

%%% 46
\bibitem[\protect\citeauthoryear{Krijnen}{2015}]{krijnen2015ifcopenshell}
\begin{botherref}
\oauthor{\bsnm{Krijnen}, \binits{T.}}:
IfcOpenShell
(2015).
\url{https://github.com/IfcOpenShell/IfcOpenShell}
\end{botherref}
\endbibitem

%%% 47
\bibitem[\protect\citeauthoryear{Kayhani et~al.}{2023}]{Kayhani2023_AprilTags}
\begin{botherref}
\oauthor{\bsnm{Kayhani}, \binits{N.}},
\oauthor{\bsnm{Schoellig}, \binits{A.}},
\oauthor{\bsnm{McCabe}, \binits{B.}}:
Perception-aware tag placement planning for robust localization of uavs in indoor construction environments.
Journal of Computing in Civil Engineering
\textbf{37}(2)
(2023)
\doiurl{10.1061/JCCEE5.CPENG-5068} .
Cited by: 1; All Open Access, Green Open Access
\end{botherref}
\endbibitem

%%% 48
\bibitem[\protect\citeauthoryear{Kim et~al.}{2022}]{kim2022ScLidar}
\begin{bchapter}
\bauthor{\bsnm{Kim}, \binits{G.}},
\bauthor{\bsnm{Yun}, \binits{S.}},
\bauthor{\bsnm{Kim}, \binits{J.}},
\bauthor{\bsnm{Kim}, \binits{A.}}:
\bctitle{Sc-lidar-slam: A front-end agnostic versatile lidar {SLAM} system}.
In: \bbtitle{2022 International Conference on Electronics, Information, and Communication (ICEIC)},
pp. \bfpage{1}--\blpage{6}
(\byear{2022}).
\doiurl{10.1109/ICEIC54506.2022.9748644}
\end{bchapter}
\endbibitem

%%% 49
\bibitem[\protect\citeauthoryear{Kayhani et~al.}{2022}]{KAYHANI2022_AprilTags}
\begin{barticle}
\bauthor{\bsnm{Kayhani}, \binits{N.}},
\bauthor{\bsnm{Zhao}, \binits{W.}},
\bauthor{\bsnm{McCabe}, \binits{B.}},
\bauthor{\bsnm{Schoellig}, \binits{A.P.}}:
\batitle{Tag-based visual-inertial localization of unmanned aerial vehicles in indoor construction environments using an on-manifold extended kalman filter}.
\bjtitle{Automation in Construction}
\bvolume{135},
\bfpage{104112}
(\byear{2022})
\doiurl{10.1016/j.autcon.2021.104112}
\end{barticle}
\endbibitem

%%% 50
\bibitem[\protect\citeauthoryear{Lajoie and Beltrame}{2024}]{Lajoie:2024:Swarm_SLAM}
\begin{barticle}
\bauthor{\bsnm{Lajoie}, \binits{P.-Y.}},
\bauthor{\bsnm{Beltrame}, \binits{G.}}:
\batitle{Swarm-slam: Sparse decentralized collaborative simultaneous localization and mapping framework for multi-robot systems}.
\bjtitle{IEEE Robotics and Automation Letters}
\bvolume{9}(\bissue{1}),
\bfpage{475}--\blpage{482}
(\byear{2024})
\doiurl{10.1109/lra.2023.3333742}
\end{barticle}
\endbibitem

%%% 51
\bibitem[\protect\citeauthoryear{Liu et~al.}{2021}]{liu2021simultaneous}
\begin{botherref}
\oauthor{\bsnm{Liu}, \binits{Y.}},
\oauthor{\bsnm{Fu}, \binits{Y.}},
\oauthor{\bsnm{Chen}, \binits{F.}},
\oauthor{\bsnm{Goossens}, \binits{B.}},
\oauthor{\bsnm{Tao}, \binits{W.}},
\oauthor{\bsnm{Zhao}, \binits{H.}}:
Simultaneous localization and mapping related datasets: A comprehensive survey.
arXiv preprint arXiv:2102.04036
(2021)
\end{botherref}
\endbibitem

%%% 52
\bibitem[\protect\citeauthoryear{Lee et~al.}{1994}]{lee1994building}
\begin{barticle}
\bauthor{\bsnm{Lee}, \binits{T.-C.}},
\bauthor{\bsnm{Kashyap}, \binits{R.L.}},
\bauthor{\bsnm{Chu}, \binits{C.-N.}}:
\batitle{Building skeleton models via 3-d medial surface axis thinning algorithms}.
\bjtitle{CVGIP: Graphical Models and Image Processing}
\bvolume{56}(\bissue{6}),
\bfpage{462}--\blpage{478}
(\byear{1994})
\end{barticle}
\endbibitem

%%% 53
\bibitem[\protect\citeauthoryear{Li et~al.}{2021}]{li2021ssc}
\begin{bchapter}
\bauthor{\bsnm{Li}, \binits{L.}},
\bauthor{\bsnm{Kong}, \binits{X.}},
\bauthor{\bsnm{Zhao}, \binits{X.}},
\bauthor{\bsnm{Huang}, \binits{T.}},
\bauthor{\bsnm{Li}, \binits{W.}},
\bauthor{\bsnm{Wen}, \binits{F.}},
\bauthor{\bsnm{Zhang}, \binits{H.}},
\bauthor{\bsnm{Liu}, \binits{Y.}}:
\bctitle{Ssc: Semantic scan context for large-scale place recognition}.
In: \bbtitle{2021 IEEE/RSJ International Conference on Intelligent Robots and Systems (IROS)},
pp. \bfpage{2092}--\blpage{2099}
(\byear{2021}).
\bcomment{IEEE}
\end{bchapter}
\endbibitem

%%% 54
\bibitem[\protect\citeauthoryear{Lv et~al.}{2020}]{lv2020targetless}
\begin{bchapter}
\bauthor{\bsnm{Lv}, \binits{J.}},
\bauthor{\bsnm{Xu}, \binits{J.}},
\bauthor{\bsnm{Hu}, \binits{K.}},
\bauthor{\bsnm{Liu}, \binits{Y.}},
\bauthor{\bsnm{Zuo}, \binits{X.}}:
\bctitle{Targetless Calibration of Lidar-imu System Based on Continuous-time Batch Estimation}.
In: \bbtitle{2020 IEEE/RSJ International Conference on Intelligent Robots and Systems (IROS)},
pp. \bfpage{9968}--\blpage{9975}
(\byear{2020}).
\bcomment{IEEE}
\end{bchapter}
\endbibitem

%%% 55
\bibitem[\protect\citeauthoryear{Lv et~al.}{2022}]{lv2022}
\begin{barticle}
\bauthor{\bsnm{Lv}, \binits{J.}},
\bauthor{\bsnm{Zuo}, \binits{X.}},
\bauthor{\bsnm{Hu}, \binits{K.}},
\bauthor{\bsnm{Xu}, \binits{J.}},
\bauthor{\bsnm{Huang}, \binits{G.}},
\bauthor{\bsnm{Liu}, \binits{Y.}}:
\batitle{{OA-LICalib}: Observability-aware intrinsic and extrinsic calibration of lidar-imu systems}.
\bjtitle{IEEE Transactions on Robotics}
\bvolume{38}(\bissue{6}),
\bfpage{3734}--\blpage{3753}
(\byear{2022})
\end{barticle}
\endbibitem

%%% 56
\bibitem[\protect\citeauthoryear{Macenski and Jambrecic}{2021}]{Macenski.2021}
\begin{barticle}
\bauthor{\bsnm{Macenski}, \binits{S.}},
\bauthor{\bsnm{Jambrecic}, \binits{I.}}:
\batitle{Slam toolbox: {SLAM} for the dynamic world}.
\bjtitle{Journal of Open Source Software}
\bvolume{6}(\bissue{61}),
\bfpage{2783}
(\byear{2021})
\doiurl{10.21105/joss.02783}
\end{barticle}
\endbibitem

%%% 57
\bibitem[\protect\citeauthoryear{McDonald et~al.}{2013}]{McDonald2013_1144}
\begin{barticle}
\bauthor{\bsnm{McDonald}, \binits{J.}},
\bauthor{\bsnm{Kaess}, \binits{M.}},
\bauthor{\bsnm{Cadena}, \binits{C.}},
\bauthor{\bsnm{Neira}, \binits{J.}},
\bauthor{\bsnm{Leonard}, \binits{J.J.}}:
\batitle{Real-time 6-dof multi-session visual {SLAM} over large-scale environments}.
\bjtitle{Robotics and Autonomous Systems}
\bvolume{61}(\bissue{10}),
\bfpage{1144}--\blpage{1158}
(\byear{2013})
\doiurl{10.1016/j.robot.2012.08.008} .
\bcomment{Selected Papers from the 5th European Conference on Mobile Robots (ECMR 2011)}
\end{barticle}
\endbibitem

%%% 58
\bibitem[\protect\citeauthoryear{Mylonas et~al.}{2021}]{Mylonas:2021:digitalTwin}
\begin{barticle}
\bauthor{\bsnm{Mylonas}, \binits{G.}},
\bauthor{\bsnm{Kalogeras}, \binits{A.}},
\bauthor{\bsnm{Kalogeras}, \binits{G.}},
\bauthor{\bsnm{Anagnostopoulos}, \binits{C.}},
\bauthor{\bsnm{Alexakos}, \binits{C.}},
\bauthor{\bsnm{Muñoz}, \binits{L.}}:
\batitle{Digital twins from smart manufacturing to smart cities: A survey}.
\bjtitle{IEEE Access}
\bvolume{9},
\bfpage{143222}--\blpage{143249}
(\byear{2021})
\doiurl{10.1109/ACCESS.2021.3120843}
\end{barticle}
\endbibitem

%%% 59
\bibitem[\protect\citeauthoryear{Macenski et~al.}{2023}]{Macenski_2023}
\begin{barticle}
\bauthor{\bsnm{Macenski}, \binits{S.}},
\bauthor{\bsnm{Moore}, \binits{T.}},
\bauthor{\bsnm{Lu}, \binits{D.V.}},
\bauthor{\bsnm{Merzlyakov}, \binits{A.}},
\bauthor{\bsnm{Ferguson}, \binits{M.}}:
\batitle{From the desks of {ROS} maintainers: A survey of modern \& capable mobile robotics algorithms in the robot operating system 2}.
\bjtitle{Robotics and Autonomous Systems}
\bvolume{168},
\bfpage{104493}
(\byear{2023})
\doiurl{10.1016/j.robot.2023.104493}
\end{barticle}
\endbibitem

%%% 60
\bibitem[\protect\citeauthoryear{Moura et~al.}{2021}]{Moura2021}
\begin{botherref}
\oauthor{\bsnm{Moura}, \binits{M.S.}},
\oauthor{\bsnm{Rizzo}, \binits{C.}},
\oauthor{\bsnm{Serrano}, \binits{D.}}:
{{BIM}-based Localization and Mapping for Mobile Robots in Construction}.
2021 IEEE International Conference on Autonomous Robot Systems and Competitions, ICARSC 2021,
12--18
(2021)
\doiurl{10.1109/ICARSC52212.2021.9429779}
\end{botherref}
\endbibitem

%%% 61
\bibitem[\protect\citeauthoryear{Macenski et~al.}{2023}]{macenski2023regulated}
\begin{botherref}
\oauthor{\bsnm{Macenski}, \binits{S.}},
\oauthor{\bsnm{Singh}, \binits{S.}},
\oauthor{\bsnm{Mart{\'\i}n}, \binits{F.}},
\oauthor{\bsnm{Gin{\'e}s}, \binits{J.}}:
Regulated pure pursuit for robot path tracking.
Autonomous Robots,
1--10
(2023)
\end{botherref}
\endbibitem

%%% 62
\bibitem[\protect\citeauthoryear{NavVis et~al.}{2022}]{NavVis2022}
\begin{botherref}
\oauthor{\bsnm{NavVis}},
\oauthor{\bsnm{News}, \binits{L.}},
\oauthor{\bsnm{Magazine}, \binits{L.}},
\oauthor{\bsnm{American~Surveyor}},
\oauthor{\bsnm{GoGeomatics}},
\oauthor{\bsnm{International}, \binits{G.}},
\oauthor{\bsnm{Week}, \binits{G.}},
\oauthor{\bsnm{BIMplus}},
\oauthor{\bsnm{Source}, \binits{S.}},
\oauthor{\bsnm{GeoConnexion}}:
{State of Mobile Mapping Survey 2022}.
NavVis
(2022)
\end{botherref}
\endbibitem

%%% 63
\bibitem[\protect\citeauthoryear{Ozog et~al.}{2016}]{ozog2016long}
\begin{barticle}
\bauthor{\bsnm{Ozog}, \binits{P.}},
\bauthor{\bsnm{Carlevaris-Bianco}, \binits{N.}},
\bauthor{\bsnm{Kim}, \binits{A.}},
\bauthor{\bsnm{Eustice}, \binits{R.M.}}:
\batitle{Long-term mapping techniques for ship hull inspection and surveillance using an autonomous underwater vehicle}.
\bjtitle{Journal of Field Robotics}
\bvolume{33}(\bissue{3}),
\bfpage{265}--\blpage{289}
(\byear{2016})
\end{barticle}
\endbibitem

%%% 64
\bibitem[\protect\citeauthoryear{Oelsch et~al.}{2021}]{Oelsch2021}
\begin{barticle}
\bauthor{\bsnm{Oelsch}, \binits{M.}},
\bauthor{\bsnm{Karimi}, \binits{M.}},
\bauthor{\bsnm{Steinbach}, \binits{E.}}:
\batitle{{R-LOAM: Improving LiDAR Odometry and Mapping with Point-to-Mesh Features of a Known 3D Reference Object}}.
\bjtitle{IEEE Robotics and Automation Letters}
\bvolume{6}(\bissue{2}),
\bfpage{2068}--\blpage{2075}
(\byear{2021})
\doiurl{10.1109/LRA.2021.3060413}
\end{barticle}
\endbibitem

%%% 65
\bibitem[\protect\citeauthoryear{Oelsch et~al.}{2022}]{Oelsch.2022}
\begin{botherref}
\oauthor{\bsnm{Oelsch}, \binits{M.}},
\oauthor{\bsnm{Karimi}, \binits{M.}},
\oauthor{\bsnm{Steinbach}, \binits{E.}}:
Ro-loam: 3d reference object-based trajectory and map optimization in lidar odometry and mapping.
IEEE Robotics and Automation Letters,
1--1
(2022)
\doiurl{10.1109/LRA.2022.3177846}
\end{botherref}
\endbibitem

%%% 66
\bibitem[\protect\citeauthoryear{Prieto et~al.}{2020}]{Prieto.2020}
\begin{bchapter}
\bauthor{\bsnm{Prieto}, \binits{S.A.}},
\bauthor{\bsnm{{Garcia de Soto}}, \binits{B.}},
\bauthor{\bsnm{Adan}, \binits{A.}}:
\bctitle{A methodology to monitor construction progress using autonomous robots}.
In: \beditor{\bsnm{Osumi}, \binits{H.}} (ed.)
\bbtitle{Proceedings of the 37th International Symposium on Automation and Robotics in Construction (ISARC)}.
\bsertitle{Proceedings of the International Symposium on Automation and Robotics in Construction (IAARC)},
pp. \bfpage{265}--\blpage{289}.
\bpublisher{{International Association for Automation and Robotics in Construction (IAARC)}},
\blocation{Tokyo, Japan}
(\byear{2020}).
\doiurl{10.22260/ISARC2020/0210}
\end{bchapter}
\endbibitem

%%% 67
\bibitem[\protect\citeauthoryear{Perez-Grau et~al.}{2017}]{amcl3d:2017}
\begin{botherref}
\oauthor{\bsnm{Perez-Grau}, \binits{F.J.}},
\oauthor{\bsnm{Caballero}, \binits{F.}},
\oauthor{\bsnm{Viguria}, \binits{A.}},
\oauthor{\bsnm{Ollero}, \binits{A.}}:
Multi-sensor three-dimensional monte carlo localization for long-term aerial robot navigation.
International Journal of Advanced Robotic Systems
\textbf{14}(5)
(2017)
\doiurl{10.1177/1729881417732757}
\end{botherref}
\endbibitem

%%% 68
\bibitem[\protect\citeauthoryear{Ramezani et~al.}{2020}]{Ramezani_2020}
\begin{bchapter}
\bauthor{\bsnm{Ramezani}, \binits{M.}},
\bauthor{\bsnm{Wang}, \binits{Y.}},
\bauthor{\bsnm{Camurri}, \binits{M.}},
\bauthor{\bsnm{Wisth}, \binits{D.}},
\bauthor{\bsnm{Mattamala}, \binits{M.}},
\bauthor{\bsnm{Fallon}, \binits{M.}}:
\bctitle{The newer college dataset: Handheld lidar, inertial and vision with ground truth}.
In: \bbtitle{2020 IEEE/RSJ International Conference on Intelligent Robots and Systems (IROS)}.
\bpublisher{IEEE},
\blocation{Las Vegas, USA}
(\byear{2020}).
\doiurl{10.1109/iros45743.2020.9340849} .
\burl{http://dx.doi.org/10.1109/IROS45743.2020.9340849}
\end{bchapter}
\endbibitem

%%% 69
\bibitem[\protect\citeauthoryear{Shaheer et~al.}{2022}]{shaheer2022robot}
\begin{botherref}
\oauthor{\bsnm{Shaheer}, \binits{M.}},
\oauthor{\bsnm{Bavle}, \binits{H.}},
\oauthor{\bsnm{Sanchez-Lopez}, \binits{J.L.}},
\oauthor{\bsnm{Voos}, \binits{H.}}:
Robot localization using situational graphs and building architectural plans.
arXiv preprint arXiv:2209.11575
(2022)
\end{botherref}
\endbibitem

%%% 70
\bibitem[\protect\citeauthoryear{Sturm et~al.}{2012}]{Sturm2012ABF}
\begin{botherref}
\oauthor{\bsnm{Sturm}, \binits{J.}},
\oauthor{\bsnm{Engelhard}, \binits{N.}},
\oauthor{\bsnm{Endres}, \binits{F.}},
\oauthor{\bsnm{Burgard}, \binits{W.}},
\oauthor{\bsnm{Cremers}, \binits{D.}}:
A benchmark for the evaluation of rgb-d {SLAM} systems.
2012 IEEE/RSJ International Conference on Intelligent Robots and Systems,
573--580
(2012)
\end{botherref}
\endbibitem

%%% 71
\bibitem[\protect\citeauthoryear{Shan et~al.}{2020}]{lio_sam_shan2020lio}
\begin{bchapter}
\bauthor{\bsnm{Shan}, \binits{T.}},
\bauthor{\bsnm{Englot}, \binits{B.}},
\bauthor{\bsnm{Meyers}, \binits{D.}},
\bauthor{\bsnm{Wang}, \binits{W.}},
\bauthor{\bsnm{Ratti}, \binits{C.}},
\bauthor{\bsnm{Rus}, \binits{D.}}:
\bctitle{Lio-sam: Tightly-coupled lidar inertial odometry via smoothing and mapping}.
In: \bbtitle{2020 IEEE/RSJ International Conference on Intelligent Robots and Systems (IROS)},
pp. \bfpage{5135}--\blpage{5142}
(\byear{2020}).
\bcomment{IEEE}
\end{bchapter}
\endbibitem

%%% 72
\bibitem[\protect\citeauthoryear{Shaheer et~al.}{2023}]{shaheer2023graphbased}
\begin{botherref}
\oauthor{\bsnm{Shaheer}, \binits{M.}},
\oauthor{\bsnm{Millan-Romera}, \binits{J.A.}},
\oauthor{\bsnm{Bavle}, \binits{H.}},
\oauthor{\bsnm{Sanchez-Lopez}, \binits{J.L.}},
\oauthor{\bsnm{Civera}, \binits{J.}},
\oauthor{\bsnm{Voos}, \binits{H.}}:
Graph-based Global Robot Localization Informing Situational Graphs with Architectural Graphs
(2023)
\end{botherref}
\endbibitem

%%% 73
\bibitem[\protect\citeauthoryear{Smith et~al.}{1990}]{smith1990estimating}
\begin{bbook}
\bauthor{\bsnm{Smith}, \binits{R.}},
\bauthor{\bsnm{Self}, \binits{M.}},
\bauthor{\bsnm{Cheeseman}, \binits{P.}}:
In: \beditor{\bsnm{Cox}, \binits{I.J.}},
\beditor{\bsnm{Wilfong}, \binits{G.T.}} (eds.)
\bbtitle{Estimating Uncertain Spatial Relationships in Robotics},
pp. \bfpage{167}--\blpage{193}.
\bpublisher{Springer},
\blocation{New York, NY}
(\byear{1990}).
\doiurl{10.1007/978-1-4613-8997-2_14} .
\burl{https://doi.org/10.1007/978-1-4613-8997-2_14}
\end{bbook}
\endbibitem

%%% 74
\bibitem[\protect\citeauthoryear{Trzeciak et~al.}{2023a}]{trzeciak2023conslamExtension}
\begin{barticle}
\bauthor{\bsnm{Trzeciak}, \binits{M.}},
\bauthor{\bsnm{Pluta}, \binits{K.}},
\bauthor{\bsnm{Fathy}, \binits{Y.}},
\bauthor{\bsnm{Alcalde}, \binits{L.}},
\bauthor{\bsnm{Chee}, \binits{S.}},
\bauthor{\bsnm{Bromley}, \binits{A.}},
\bauthor{\bsnm{Brilakis}, \binits{I.}},
\bauthor{\bsnm{Alliez}, \binits{P.}}:
\batitle{Conslam: Construction data set for {SLAM}}.
\bjtitle{Journal of Computing in Civil Engineering}
\bvolume{37}(\bissue{3}),
\bfpage{04023009}
(\byear{2023})
\end{barticle}
\endbibitem

%%% 75
\bibitem[\protect\citeauthoryear{Trzeciak et~al.}{2023b}]{trzeciak2023conslam}
\begin{bchapter}
\bauthor{\bsnm{Trzeciak}, \binits{M.}},
\bauthor{\bsnm{Pluta}, \binits{K.}},
\bauthor{\bsnm{Fathy}, \binits{Y.}},
\bauthor{\bsnm{Alcalde}, \binits{L.}},
\bauthor{\bsnm{Chee}, \binits{S.}},
\bauthor{\bsnm{Bromley}, \binits{A.}},
\bauthor{\bsnm{Brilakis}, \binits{I.}},
\bauthor{\bsnm{Alliez}, \binits{P.}}:
\bctitle{Conslam: Periodically collected real-world construction dataset for {SLAM} and progress monitoring}.
In: \bbtitle{Computer Vision--ECCV 2022 Workshops: Tel Aviv, Israel, October 23--27, 2022, Proceedings, Part VII},
pp. \bfpage{317}--\blpage{331}
(\byear{2023}).
\bcomment{Springer}
\end{bchapter}
\endbibitem

%%% 76
\bibitem[\protect\citeauthoryear{Umeyama}{1991}]{umeyama}
\begin{barticle}
\bauthor{\bsnm{Umeyama}, \binits{S.}}:
\batitle{Least-squares estimation of transformation parameters between two point patterns}.
\bjtitle{IEEE Transactions on Pattern Analysis and Machine Intelligence}
\bvolume{13}(\bissue{4}),
\bfpage{376}--\blpage{380}
(\byear{1991})
\doiurl{10.1109/34.88573}
\end{barticle}
\endbibitem

%%% 77
\bibitem[\protect\citeauthoryear{Vizzo et~al.}{2023}]{vizzo2023kiss}
\begin{barticle}
\bauthor{\bsnm{Vizzo}, \binits{I.}},
\bauthor{\bsnm{Guadagnino}, \binits{T.}},
\bauthor{\bsnm{Mersch}, \binits{B.}},
\bauthor{\bsnm{Wiesmann}, \binits{L.}},
\bauthor{\bsnm{Behley}, \binits{J.}},
\bauthor{\bsnm{Stachniss}, \binits{C.}}:
\batitle{Kiss-icp: In defense of point-to-point icp--simple, accurate, and robust registration if done the right way}.
\bjtitle{IEEE Robotics and Automation Letters}
\bvolume{8}(\bissue{2}),
\bfpage{1029}--\blpage{1036}
(\byear{2023})
\end{barticle}
\endbibitem

%%% 78
\bibitem[\protect\citeauthoryear{Vega~Torres et~al.}{2022}]{vega:2022:2DLidarLocalization}
\begin{bchapter}
\bauthor{\bsnm{Vega~Torres}, \binits{M.A.}},
\bauthor{\bsnm{Braun}, \binits{A.}},
\bauthor{\bsnm{Borrmann}, \binits{A.}}:
\bctitle{Occupancy grid map to pose graph-based map: Robust {BIM}-based 2d- lidar localization for lifelong indoor navigation in changing and dynamic environments}.
In: \beditor{\bsnm{Eilif~Hjelseth}, \binits{S.F.S.}},
\beditor{\bsnm{Scherer}, \binits{R.}} (eds.)
\bbtitle{eWork and eBusiness in Architecture, Engineering and Construction: ECPPM 2022},
pp. \bfpage{265}--\blpage{289}.
\bpublisher{CRC Press},
\blocation{Trondheim, Norway}
(\byear{2022}).
\burl{https://publications.cms.bgu.tum.de/2022_ECPPM_Vega.pdf}
\end{bchapter}
\endbibitem

%%% 79
\bibitem[\protect\citeauthoryear{Vega~Torres et~al.}{2023}]{vega:2023:BIM_SLAM}
\begin{bchapter}
\bauthor{\bsnm{Vega~Torres}, \binits{M.A.}},
\bauthor{\bsnm{Braun}, \binits{A.}},
\bauthor{\bsnm{Borrmann}, \binits{A.}}:
\bctitle{{BIM}-{SLAM}: Integrating {BIM} models in multi-session {SLAM} for lifelong mapping using {3D LiDAR}}.
In: \bbtitle{Proceedings of the 40th International Symposium on Automation and Robotics in Construction (ISARC 2023)}.
\bpublisher{{International Association for Automation and Robotics in Construction (IAARC)}},
\blocation{Chennai, India}
(\byear{2023}).
\burl{https://publications.cms.bgu.tum.de/vega_2023_BIM_SLAM_ISARC.pdf}
\end{bchapter}
\endbibitem

%%% 80
\bibitem[\protect\citeauthoryear{Vega~Torres et~al.}{2021}]{Vega:2022:ObjectDetection}
\begin{barticle}
\bauthor{\bsnm{Vega~Torres}, \binits{M.A.}},
\bauthor{\bsnm{Braun}, \binits{A.}},
\bauthor{\bsnm{Noichl}, \binits{F.}},
\bauthor{\bsnm{Borrmann}, \binits{A.}},
\bauthor{\bsnm{Bauer}, \binits{H.}},
\bauthor{\bsnm{Wohlfeld}, \binits{D.}}:
\batitle{Recognition of temporary vertical objects in large point clouds of construction sites}.
\bjtitle{Proceedings of the Institution of Civil Engineers - Smart Infrastructure and Construction}
\bvolume{174}(\bissue{4}),
\bfpage{134}--\blpage{149}
(\byear{2021})
\doiurl{10.1680/jsmic.21.00033}
\end{barticle}
\endbibitem

%%% 81
\bibitem[\protect\citeauthoryear{Wang et~al.}{2020}]{wang2020intensity}
\begin{bchapter}
\bauthor{\bsnm{Wang}, \binits{H.}},
\bauthor{\bsnm{Wang}, \binits{C.}},
\bauthor{\bsnm{Xie}, \binits{L.}}:
\bctitle{Intensity scan context: Coding intensity and geometry relations for loop closure detection}.
In: \bbtitle{2020 IEEE International Conference on Robotics and Automation (ICRA)},
pp. \bfpage{2095}--\blpage{2101}
(\byear{2020}).
\bcomment{IEEE}
\end{bchapter}
\endbibitem

%%% 82
\bibitem[\protect\citeauthoryear{Xu et~al.}{2022}]{FAST-LIO2_Xu.2022}
\begin{botherref}
\oauthor{\bsnm{Xu}, \binits{W.}},
\oauthor{\bsnm{Cai}, \binits{Y.}},
\oauthor{\bsnm{He}, \binits{D.}},
\oauthor{\bsnm{Lin}, \binits{J.}},
\oauthor{\bsnm{Zhang}, \binits{F.}}:
{FAST-LIO2}: Fast direct lidar-inertial odometry.
IEEE Transactions on Robotics,
1--21
(2022)
\doiurl{10.1109/TRO.2022.3141876}
\end{botherref}
\endbibitem

%%% 83
\bibitem[\protect\citeauthoryear{Yang}{2018}]{Yang.2018}
\begin{botherref}
\oauthor{\bsnm{Yang}, \binits{H.}}:
GitHub - PgmMapcreator: Create pgm map from Gazebo World File for ROS Localization.
(2018).
\url{https://github.com/hyfan1116/pgmmapcreator}
\end{botherref}
\endbibitem

%%% 84
\bibitem[\protect\citeauthoryear{Yin et~al.}{2023}]{YIN_2023_104641}
\begin{barticle}
\bauthor{\bsnm{Yin}, \binits{H.}},
\bauthor{\bsnm{Lin}, \binits{Z.}},
\bauthor{\bsnm{Yeoh}, \binits{J.K.W.}}:
\batitle{Semantic localization on {BIM}-generated maps using a 3d lidar sensor}.
\bjtitle{Automation in Construction}
\bvolume{146},
\bfpage{104641}
(\byear{2023})
\doiurl{10.1016/j.aut con.2022.104641}
\end{barticle}
\endbibitem

%%% 85
\bibitem[\protect\citeauthoryear{Zhang et~al.}{2022}]{zhang2022multicamera}
\begin{botherref}
\oauthor{\bsnm{Zhang}, \binits{L.}},
\oauthor{\bsnm{Camurri}, \binits{M.}},
\oauthor{\bsnm{Wisth}, \binits{D.}},
\oauthor{\bsnm{Fallon}, \binits{M.}}:
Multi-Camera LiDAR Inertial Extension to the Newer College Dataset
(2022)
\end{botherref}
\endbibitem

%%% 86
\bibitem[\protect\citeauthoryear{Zimmerman et~al.}{2023}]{zimmerman2023}
\begin{barticle}
\bauthor{\bsnm{Zimmerman}, \binits{N.}},
\bauthor{\bsnm{Guadagnino}, \binits{T.}},
\bauthor{\bsnm{Chen}, \binits{X.}},
\bauthor{\bsnm{Behley}, \binits{J.}},
\bauthor{\bsnm{Stachniss}, \binits{C.}}:
\batitle{Long-term localization using semantic cues in floor plan maps}.
\bjtitle{IEEE Robotics and Automation Letters}
\bvolume{8}(\bissue{1}),
\bfpage{176}--\blpage{183}
(\byear{2023})
\doiurl{10.1109/LRA.2022.3223556}
\end{barticle}
\endbibitem

%%% 87
\bibitem[\protect\citeauthoryear{Zhang et~al.}{2023}]{hiltiChallenge2022}
\begin{barticle}
\bauthor{\bsnm{Zhang}, \binits{L.}},
\bauthor{\bsnm{Helmberger}, \binits{M.}},
\bauthor{\bsnm{Fu}, \binits{L.F.T.}},
\bauthor{\bsnm{Wisth}, \binits{D.}},
\bauthor{\bsnm{Camurri}, \binits{M.}},
\bauthor{\bsnm{Scaramuzza}, \binits{D.}},
\bauthor{\bsnm{Fallon}, \binits{M.}}:
\batitle{Hilti-oxford dataset: A millimeter-accurate benchmark for simultaneous localization and mapping}.
\bjtitle{IEEE Robotics and Automation Letters}
\bvolume{8}(\bissue{1}),
\bfpage{408}--\blpage{415}
(\byear{2023})
\doiurl{10.1109/LRA.2022.3226077}
\end{barticle}
\endbibitem

%%% 88
\bibitem[\protect\citeauthoryear{Zelinsky et~al.}{1993}]{Zelinsky.1993}
\begin{bchapter}
\bauthor{\bsnm{Zelinsky}, \binits{A.}},
\bauthor{\bsnm{Jarvis}, \binits{R.A.}},
\bauthor{\bsnm{Byrne}, \binits{J.}},
\bauthor{\bsnm{Yuta}, \binits{S.}}, \betal:
\bctitle{Planning paths of complete coverage of an unstructured environment by a mobile robot}.
In: \beditor{\bsnm{Zelinsky}, \binits{A.}},
\beditor{\bsnm{Jarvis}, \binits{R.A.}},
\beditor{\bsnm{Byrne}, \binits{J.C.}},
\beditor{\bsnm{Yuta}, \binits{S.}} (eds.)
\bbtitle{Proceedings of International Conference on Advanced Robotics}.
\bsertitle{Proceedings of international conference on advanced robotics},
vol. \bseriesno{13},
pp. \bfpage{533}--\blpage{538}.
\bpublisher{Citeseer},
\blocation{Tsukuba, Japan}
(\byear{1993}).
\bcomment{Citeseer}.
\burl{http://pinkwink.kr/attachment/cfile3.uf@1354654A4E8945BD13FE77.pdf}
\end{bchapter}
\endbibitem

%%% 89
\bibitem[\protect\citeauthoryear{Zhang and Singh}{2014}]{Zhang.07122014}
\begin{bchapter}
\bauthor{\bsnm{Zhang}, \binits{J.}},
\bauthor{\bsnm{Singh}, \binits{S.}}:
\bctitle{{LOAM}: Lidar odometry and mapping in real-time.}
In: \bbtitle{Robotics: Science and Systems},
vol. \bseriesno{2}.
\bconflocation{Berkeley, CA},
pp. \bfpage{1}--\blpage{9}
(\byear{2014}).
\doiurl{10.15607/RSS.2014.X.007}
\end{bchapter}
\endbibitem

%%% 90
\bibitem[\protect\citeauthoryear{Zheng and Zhu}{2023}]{zheng2023traj}
\begin{botherref}
\oauthor{\bsnm{Zheng}, \binits{X.}},
\oauthor{\bsnm{Zhu}, \binits{J.}}:
Traj-lo: In defense of lidar-only odometry using an effective continuous-time trajectory.
arXiv preprint arXiv:2309.13842
(2023)
\end{botherref}
\endbibitem

\end{thebibliography}

\end{document}